\documentclass[iicol,sn-basic]{sn-jnl}
\usepackage{natbib}
\usepackage{booktabs} 
\usepackage{balance}
\usepackage{amsmath}
\usepackage{epsfig}
\usepackage{graphicx}
\usepackage{caption}
\usepackage{subfigure}
\usepackage{float}
\usepackage{subeqnarray}
\usepackage{cases}
\usepackage{amssymb}
\usepackage{amsmath}
\usepackage{amsthm}
\usepackage{bm}
\usepackage{url}
\usepackage{hhline}
\usepackage{setspace}

\usepackage{multirow}
\usepackage{makecell}
\usepackage{threeparttable}
\usepackage{color}
\usepackage{makecell}

\theoremstyle{definition}

\theoremstyle{remark}

\usepackage{color}
\usepackage{bm}
\usepackage{amsmath}
\usepackage{stfloats}  
\usepackage{subfigure}
\usepackage{float}
\usepackage{caption}
\usepackage{multirow}
\jyear{2021}%

\begin{document}

\title[Article Title]{CoCoNet: Coupled Contrastive Learning Network with Multi-level Feature Ensemble for Multi-modality Image Fusion}

\author{\fnm{Jinyuan} \sur{Liu}\textsuperscript{2,}\textsuperscript{3}\footnotemark[2]}\email{atlantis918@hotmail.com}
\author{\fnm{Runjia} \sur{Lin}\textsuperscript{1,}\textsuperscript{3}\footnotemark[2]}\email{linrunja@gmail.com}
\author[1,3]{\fnm{Guanyao} \sur{Wu}}\email{rollingplainko@gamil.com}
\author[2,3,4]{\fnm{Risheng} \sur{Liu}}\email{rsliu@dlut.edu.cn}
\author[2,3]{\fnm{Zhongxuan} \sur{Luo}}\email{zxluo@dlut.edu.cn}
\author*[2,3]{\fnm{Xin} \sur{Fan}}\email{xin.fan@dlut.edu.cn}

\affil[1]{\orgdiv{ School of Software}, \orgname{Dalian University of Technology}, \orgaddress{ \city{Dalian 116024}, \country{China}}}
\affil[2]{\orgdiv{ DUT-RU International School of Information Science $\&$ Engineering}, \orgname{Dalian University of Technology}, \orgaddress{ \city{Dalian 116620},  \country{China}}}
\affil[3]{\orgdiv{Key Laboratory for Ubiquitous Network and Service Software of Liaoning Province}, \orgaddress{ \city{Dalian 116620}, \country{China}}}
\affil[4]{ \orgdiv{Peng Cheng Laboratory}, \orgaddress{ \city{Shenzhen}, \postcode{518000},   \country{China}}}

\abstract{Infrared and visible image fusion targets to provide an informative image by combining complementary information from different sensors. Existing learning-based fusion approaches attempt to construct various loss functions to preserve complementary features, while neglecting to discover the inter-relationship between the two modalities, leading to redundant or even invalid information on the fusion results. Moreover, most methods focus on strengthening the network with an increase in depth while neglecting the importance of feature transmission, causing vital information degeneration. To alleviate these issues, we propose a coupled contrastive learning network, dubbed CoCoNet, to realize infrared and visible image fusion in an end-to-end manner. Concretely, to simultaneously retain typical features from both modalities and to avoid artifacts emerging on the fused result, we develop a coupled contrastive constraint in our loss function. In a fused image, its foreground target /  background detail part is pulled close to the infrared / visible source and pushed far away from the visible / infrared source in the representation space. We further exploit image characteristics to provide data-sensitive weights, allowing our loss function to build a more reliable relationship with source images. A multi-level attention module is established to learn rich hierarchical feature representation and to comprehensively transfer features in the fusion process. We also apply the proposed CoCoNet on medical image fusion of different types, e.g., magnetic resonance image, positron emission tomography image, and single photon emission computed tomography image. Extensive experiments demonstrate that our method achieves state-of-the-art (SOTA) performance under both subjective and objective evaluation, especially in preserving prominent targets and recovering vital textural details. 
}


\maketitle

\section{Introduction}

Multi-sensor images can acquire complementary and comprehensive information from the same scene for better visual understanding and scene perception, which breaks through the limitations of single sensor imaging~\cite{li2018fusion}. By combining important information from different sensors, a composite image is generated for follow-up image processing or decision-making. In particular, infrared and visible image fusion~(IVIF) is an indispensable branch in computer vision community. The gnerated fused results have been widely used for the subsequent applications, including object detection~\cite{wong2017nitroaromatic,wang2023interactively,liu2023bi}, pedestrian re-identification~\cite{duan2017pedestrian}, semantic segmentation~\cite{pu2018graphnet,liu2023multi,liu2023paif}, and military monitoring.

Visible sensors imaging by reflecting lights to provide high spatial resolution background details. However, the targets cannot be seen clearly caused by poor lighting or camouflage conditions. In contrast, infrared sensors image by discriminative thermal radiation emitted from objects, which is immune to challenging conditions and work all day and night. Therefore, it is worthwhile to fuse the infrared and visible image into a single image that simultaneously retain the vital information for both sides.

In the past few years, a large number of approaches for realizing IVIF have been proposed. According to their corresponding adopted theories, these methods can be divided into five categories, including multi-scale transform based methods~\cite{li2013image}, sparse representation based methods~\cite{zhang2013dictionary,zhang2018sparse}, subspace decomposition based methods~\cite{Lu2014}, hybrid-based methods~\cite{ma2017infrared,liu2015general}, optimization model based methods~\cite{ma2016infrared,zhao2020bayesian} and others. Some of these methods are dedicated to designing various feature transforms to learn better feature representation~\cite{li2013image,yan2015infrared}. The others are attempt to discover appropriate fusion rules~\cite{ma2017infrared,zhang2018sparse}. However, these methods rely on hand-craft ways and typically have to be time consuming.

Recently, researchers introduced convolutional neural networks~(CNNs) to the field of IVIF, with state-of-the-art performance~\cite{li2018densefuse,li2018infrared,ma2019fusiongan,xu2019learning}. Generally, the deep learning-based methods can be divided into three categories,~\emph{i.e.,}~auto-encoder-based methods~\cite{li2018densefuse,MFEIF2021,zhao2020didfuse}, end-to-end CNN-based methods~\cite{rfn2021,U2Fusion2020,PMGI} and generative adversarial~\cite{ma2019fusiongan,GANMcC}
network-based methods. 

\begin{figure}
	\centering
	\setlength{\tabcolsep}{1pt}
	\begin{tabular}{c}
		\includegraphics[width=0.48\textwidth]{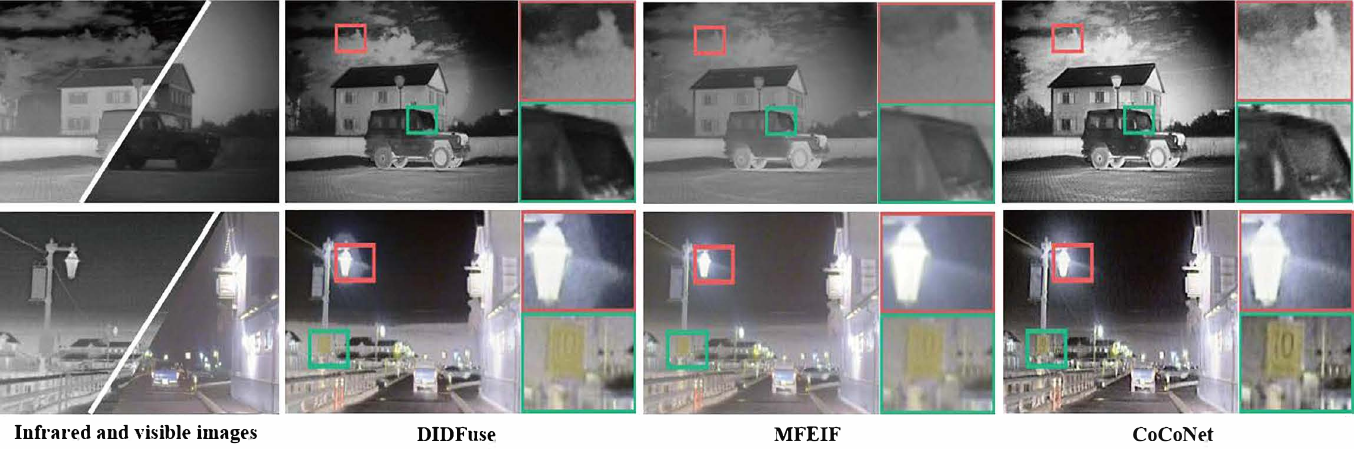}
		\\
	\end{tabular}
	\caption{Visual illustration that highlights the existing issues in infrared and visible image fusion. Observe that the regions marked by yellow arrows signify redundant information, while those indicated by blue arrows denote information degradation. For example, DIDFuse is prone to retaining redundant information, as evidenced by the front windshield of the Jeep and the surrounding halo of light. On the other hand, MFEIF can result in the loss of essential information during its fusion process, as exemplified by the reduced visibility of clouds in the sky and the obscured indicator on the pole.}
	\label{fig:first}
\end{figure}

These existing learning-based methods have achieved advanced performance, but several unresolved issues necessitate further attention~(shown in Figure~\ref{fig:first}). First,  utilizing CNNs in the IVIF is challenging due to the lack of labeled fused images for supervision. Existing approaches attempt to solve it by designing various loss functions to penalize differences between input and fused images, causing massive redundant information emerged to the fused result~\cite{zhao2020didfuse}.
Second, to fuse corresponding features from both sides, existing approaches rely on adjusting trade-off parameters in their loss functions. This causes imbalanced fusion performance and is labor-intensive~\cite{li2018densefuse,PMGI}.
Third, existing learning-based methods introduce skip connections to reduce gradient disappearance and feature degradation during the fusion process. However, the fused results still suffer from vital information loss~\cite{MFEIF2021}.

In this paper, to address the issues mentioned above, we propose a coupled contrastive learning network with multi-level feature ensemble for fusing the infrared and visible images in an end-to-end manner, termed CoCoNet.  First, we develop a coupled contrastive learning scheme to guide the model to distinguish significant complementary features, \emph{i.e.,} distinct targets and textural details. This makes the model capable of extracting and fusing only the desirable features from each modality. Second, a measuring mechanism is applied to compute the proportional importance of source images for generating data-driven weights. Afterward, these generated weights are applied in our loss function to replace the manually-craft trade-off parameters. Under this design, the model can generate a fused image with adaptation to specific source images. In addition, a multi-level attention module is designed to learn rich hierarchical feature representation and ensure these features have been fully utilized. Experiments demonstrate that CoCoNet can be generalized for fusing the different types of medical images, \emph{e.g.,} magnetic resonance image~(MRI) and single photon emission computed tomography~(SPECT) image, aiming to retain anatomical information from MRI image and functional information from SPECT image simultaneously.
Our contributions are three-fold:

\begin{itemize}
	
	\item In light of the major cornerstone of IVIF, preserving complementary information while simultaneously eliminating redundancy across two modalities, we introduce coupled contrastive constraints to achieve this goal and seamlessly integrate it into the loss function.   
	
	\item We raise a data-driven mechanism to calculate the information retention degrees to boost intensity and detail consistency between the source images and fused results. This approach mitigates the need for labor-intensive manual parameterization within the loss function and facilitates adaptation to source image characteristics.  	
	
	\item By designing a multi-level attention module~(MAM), our network is able to learn rich hierarchical features representation and effectively avoid feature degeneration during the fusion process.
\end{itemize}
Extensive qualitative and quantitative experiments on multiple datasets demonstrate the superiority of our method, outperforming nine state-of-the-art IVIF methods by a large margin. Besides, CoCoNet is able to be extended to medical images and achieves superior performance.

\section{Related Works}
{	In this section, we briefly overview the traditional-based fusion methods and deep learning-based ones. Furthermore, the utilization of attention mechanism and contrastive learning in deep learning are also given in following.
	\subsection{Multi-Modality Image Fusion Methods}
	\subsubsection{Infrared and Visible Image Fusion}
	\noindent\textbf{Traditional Fusion Methods}
	
	In the past few decades, extensive traditional infrared and visible image fusion have been proposed and applied well. Generally, according to their corresponding adopted theories, all of these traditional-based methods can be divided into six representative categories, \emph{i.e.,} multi-scale transform~(MST)-based methods~\cite{li2013image,ma2017infrared}, spare representation~(SR)-based methods~\cite{Cui2015Detail,zhang2018sparse}, saliency-based methods~\cite{ma2017infrared}, sub-space-based methods, model-based methods~\cite{zhao2018multisensor,bilevel,zhao2020bayesian}, hybrid models and other methods~\cite{gangapure2017superpixel}.  
	
	MST are widely used in the field of IVIF, achieving outstanding fusion performance. These MST-methods target to design various transformation tools, \emph{e.g.,} wavelet transform~\cite{petrovic2004gradient,lewis2007pixel}, non-subsampled contourlet transform~\cite{bhatnagar2013directive}, contourlet transform~\cite{da2006nonsubsampled}, edge-preserving filter based transform~\cite{ma2017infrared}, and Retinex theory-based transform, to extract features at different scales. Then these transformed features are merged by specific fusion rules. Finally, the fused results are provided by reverse their adopted transforms. \cite{li2013image} applied a guided filter in dealing IVIF task, which provides visual-pleasant fused results with less noise interference. To preserve abundant details on the fused results, \cite{meng2017image} introduced an IVIF method was based on the NSCT and object region detection.
	
	Different from MST-based fusion methods with prefixed basis functions, SR-based methods~\cite{yin2017novel,kim2016joint,zhang2018sparse} targets to construct an over-complete dictionary from high-quality natural images. The learned dictionary can sparsely represent the infrared and visible images, thus potentially enhancing the representation of the final fused results. For instance,  \cite{kim2016joint} proposed a method that was based on patch clustering, which achieves appealing fusion performance and removes the redundancy of the learned dictionary.
	
	Saliency targets to calculate the significant pixel than neighbours, which attracts visual attention under a bottom-up manner. To this end, researchers have adopted saliency methods to the IVIF task. \cite{ma2017infrared} designed a rolling guided filter to decompose the source images into the base and detail layer. Then they used the visual saliency map and weighted least square optimization to merge the base layer and detail layer, respectively. 
	
	The core idea of subspace-based methods is to project high-dimensional source images into low-dimensional subspaces, which is easy to capture the intrinsic structures. Principal Component Analysis
	(PCA)~\cite{PCA}, Independent component analysis~(ICA)~\cite{ICA}, and Intensity-Hue-Saturation~(IHS)
	~\cite{IHS}~are in this category. \cite{Bavirisetti2017Multi} used fourth-order partial differential equations to decompose image, and then merge the decomposed detail information by PCA. Thus, the abundant can be transferred to the fusion results.
	
	Model-based methods also shed new light on IVIF~\cite{ma2016infrared,bilevel}. Based on total variation, \cite{ma2016infrared} first introduced a method for IVIF, which kept the intensity information of infrared images and retained the detailed information of visible images simultaneously. More recently, \cite{bilevel} proposed a bilevel optimization-based method to solve the IVIF and medical image fusion. In addition, the data-driven weight is employed in the model to replace the hand-craft parameters and further boost the fusion performance. 
	
	The aforementioned IVIF methods all have two sides, and it is worthwhile to combine their advantages to improve the fusion performance. To this end, \cite{liu2015general} introduced a unified fusion framework by combining MST and improved SR; the MST is employed to decompose the source images, and the SR is utilized to obtain fusion coefficients. 
	
	Although these traditional-based methods play their roles to the IVIF task with achieving meaningful performance. However, the hand-craft feature extractors and manually designed fusion rules make these traditional-based methods more and more complex, resulting in time-consuming and limited fusion performance for various scenes.  
	
	\noindent\textbf{Deep Learning-Based Fusion Methods}
	
	Deep learning technique has achieve significant advances in tremendous  tasks~\cite{li2018densefuse,bilevel,ma2019fusiongan,MFEIF2021,U2Fusion2020,rfn2021,GANMcC,zhao2020didfuse,wang2022unsupervised,jiang2022bilevel,jiang2022towards,liu2022twin,liu2021retinex,ma2022toward,10231109,liu2022learning,ma2023bilevel,ma2022low}, due to its strong non-linear fitting ability from massive data. Early IVIF methods only employed deep learning for feature extraction or generating weight maps. For instance, \cite{bilevel} adopted two pre-trained CNN to generate two weight maps for merging the base and detail layer, respectively. However, the whole process is still under a traditional optimization model, which limits the fusion performance. 
	
	Recently, a part of learning-based methods~\cite{li2018densefuse,MFEIF2021,zhao2020didfuse,rfn2021,zhao2023cddfuse} that utilizing auto-encoder architecture have been proposed. The pre-trained auto-encoder are employed to realize feature extraction and feature reconstruction, in which the fusion rules are fulfilled by manually-designed. \cite{li2018densefuse} first introduce an auto-encoder network for IVIF. By integrating a dense block in the encoder part, the feature can be extracted comprehensively. Then they used addition and $l1$-norm rule in the fusion layer to generate fused results. Consider that vital information often degenerates from the network, \cite{MFEIF2021} employed different reception dilated convolutions to extract feature from a multi-scale prospective, and then merged these extract features by the edge attention mechanism. More recently, \cite{zhao2020didfuse} proposed an auto-encoder based fusion network, in which the encoder decomposes an image into background and detail feature maps with low-/high-frequency information, respectively. Then the fused result is generated by via the decoder part. 
	
	Apart from that, extensive generative adversarial network (GAN)-based fusion methods~\cite{ma2019fusiongan,ddcgan,GANMcC} have been proposed, due to its powerful unsupervised distribution estimation ability. For the first time, ~\cite{ma2019fusiongan} established an adversarial game between the visible image and fused result to enhance the textural details. However, they only used the information from the visible image, thus losing contrast or contour of the target on the fused result. To ameliorate this issue, they later introduce a dual discriminators GAN~\cite{ddcgan}, in which both infrared and visible images are participate in the network, thus significantly boost the fusion performance. As more attempts, ~\cite{rfn2021} introduced an end-to-end GAN model that integrates multi-classification constraints. ~\cite{liu2022target} designed a fusion network with one generator and dual discriminators. By introducing a salient mask in their discriminative process, it can preserves structural information of targets from the infrared and textural details from the visible.
	
	Furthermore, a growing number of researchers concentrate on designing the general image fusion network~\cite{IFCNNeltit,PMGI,U2Fusion2020,lei2023galfusion,li2022learning,liu2022unified,li2023learning,liu2022attention}. ~\cite{IFCNNeltit} introduce a unified fusion network for realizing various image fusion tasks with high efficiency. The network only requires training on one type of fusion dataset, and adjust the fusion rule to face the other types of fusion tasks. ~\cite{zhang2021sdnet}introduced the idea of squeeze and decomposition into the field of image fusion, combined gradient and intensity information to construct a general loss function, and proposed a general fusion network. To realize multiple fusion tasks into a single model, \cite{U2Fusion2020} came up with a novel fusion network, in which overcomes the storage and computation issues or catastrophic forgetting in the training phase.
	
	Recently, transformer~\cite{vaswani2017attention} has received extensive attention since it was proposed in the field of natural language processing. Later, \cite{dosovitskiy2020image} proposed the Vision Transformer(ViT) for image classification. This successful examples in other areas of computer vision have inspired the extensive development of transformer based methods in the field of image fusion. 
	\cite{vs2021image} took the lead in proposing an image fusion transformer model that can simultaneously use local information and long-range information, which makes up for the lack of the ability of CNN model to extract global context information. The transformer more effectively fuses complementary information of different modalities. \cite{Ma2022SwinFusion} proposed a general fusion method, it can retain pixels of source modalities with maximum intensity, which actually intends to retain foreground targets from the thermal image and background textures from the visible image, since they are exactly regions with higher pixel intensity in each modality. Besides, recent advances in diffusion model also provide fresh perspectives for IVIF~\cite{zhao2023ddfm}.
	
	\subsubsection{Medical Image Fusion}
	Similar to IVIF, the existing traditional methods of MIF can be roughly classified into two types: multi-scale transform based and sparse representation based medical image fusion. 
	
	MST is also a commonly used mean in the field of medical image fusion. Compared with MST in IVIF, their processing flow are similar, but different in details~\cite{li2023gesenet}. In the medical field, the common multi-scale transformation methods usually use different wavelet to transform the domain. \cite{Yang2008Multimodality} used the contourlet domain for medical image fusion, proposed a contrast measurement method based on the characteristics of the contour wave to select the part suitable for the human visual system, and further improved the quality of the fused image by combining various fusion rules.
	
	In the field of sparse representation on medical images, \cite{liu2015simultaneous}proposed an adaptive sparse representation model, which discards the redundant dictionary to learn a compact sub dictionary. The source image block adaptively selects features from the sub dictionary to achieve the effect of diminishing computing costs and effectively reducing artifacts. \cite{liu2019medical}integrated morphological principal component analysis and convolutional sparse representation into a unified optimization framework, and realized stable visualization effect.
	
	With the extensive application of deep learning in other image fusion fields~\cite{liu2022attention}, some general fusion frameworks~\cite{zhang2021sdnet,Ma2022SwinFusion,U2Fusion2020} also integrate medical image fusion as their branch tasks. \cite{xu2021emfusion} proposed an unsupervised enhanced medical image fusion network to retain both surface and deep-level constraints information. 

\begin{figure}
	\centering
	\setlength{\tabcolsep}{1pt}
	\begin{tabular}{cccccc}
		\includegraphics[width=0.074\textwidth]{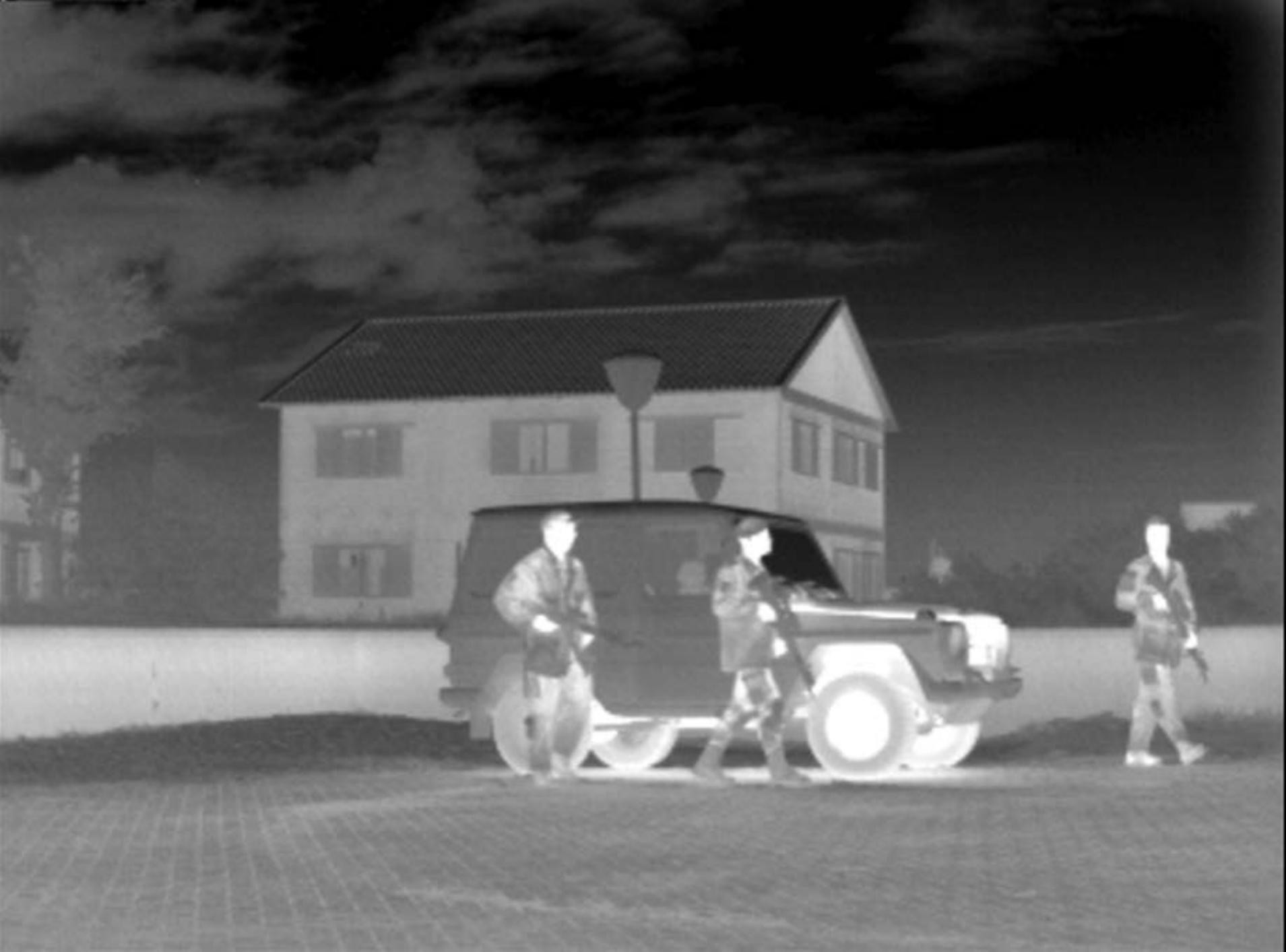}
		&\includegraphics[width=0.074\textwidth]{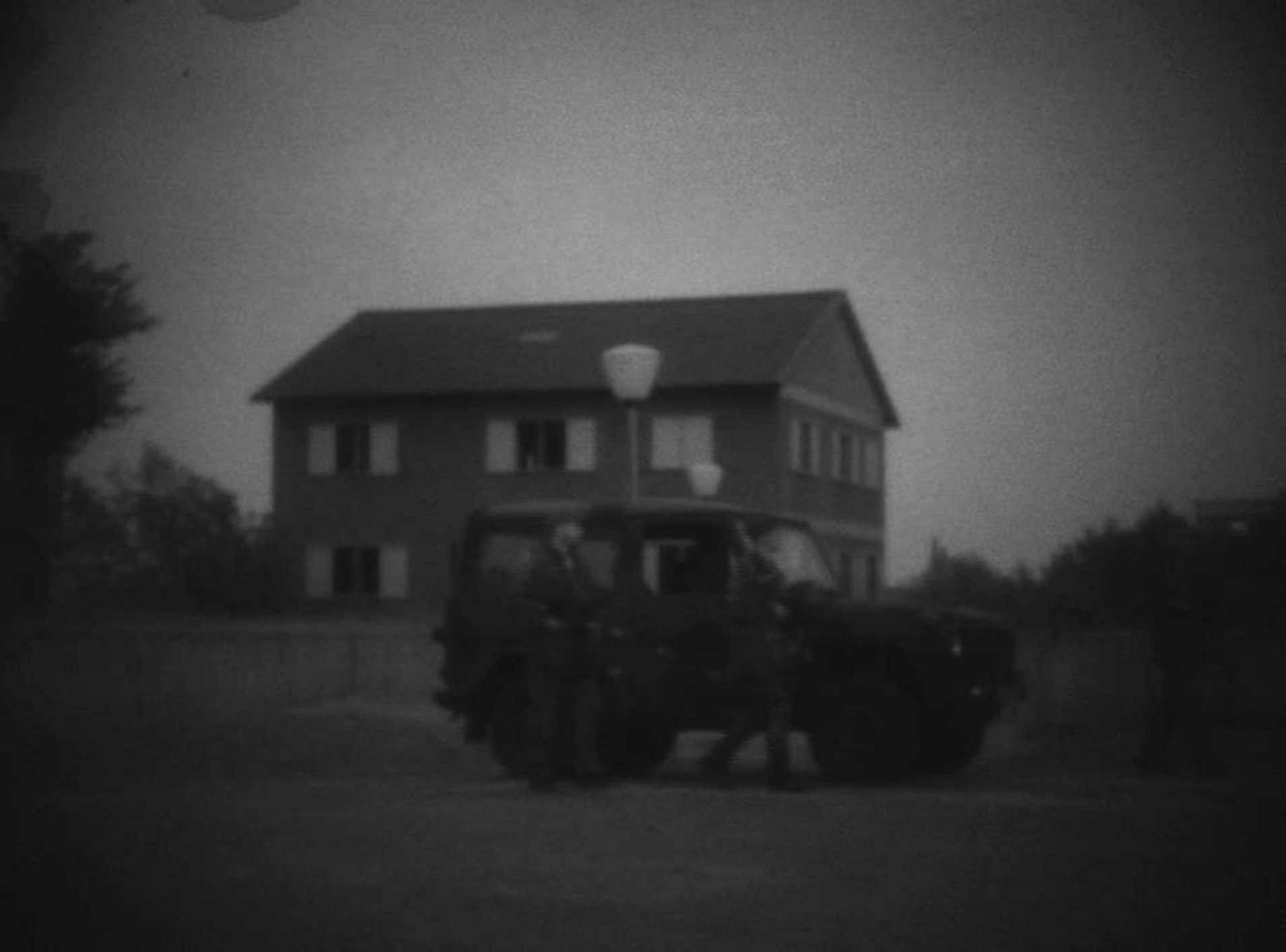}
		&\includegraphics[width=0.074\textwidth]{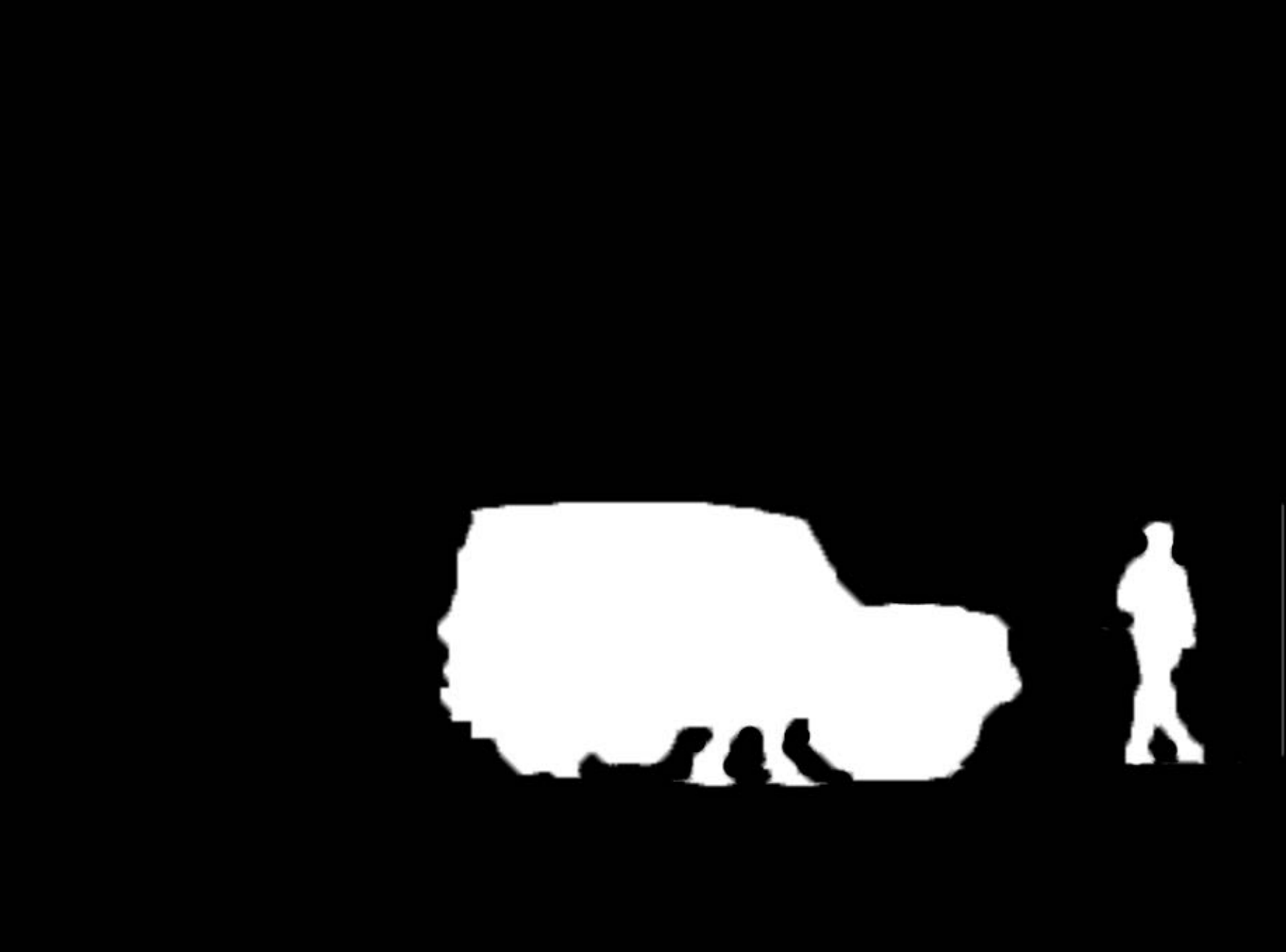}
		&\includegraphics[width=0.074\textwidth]{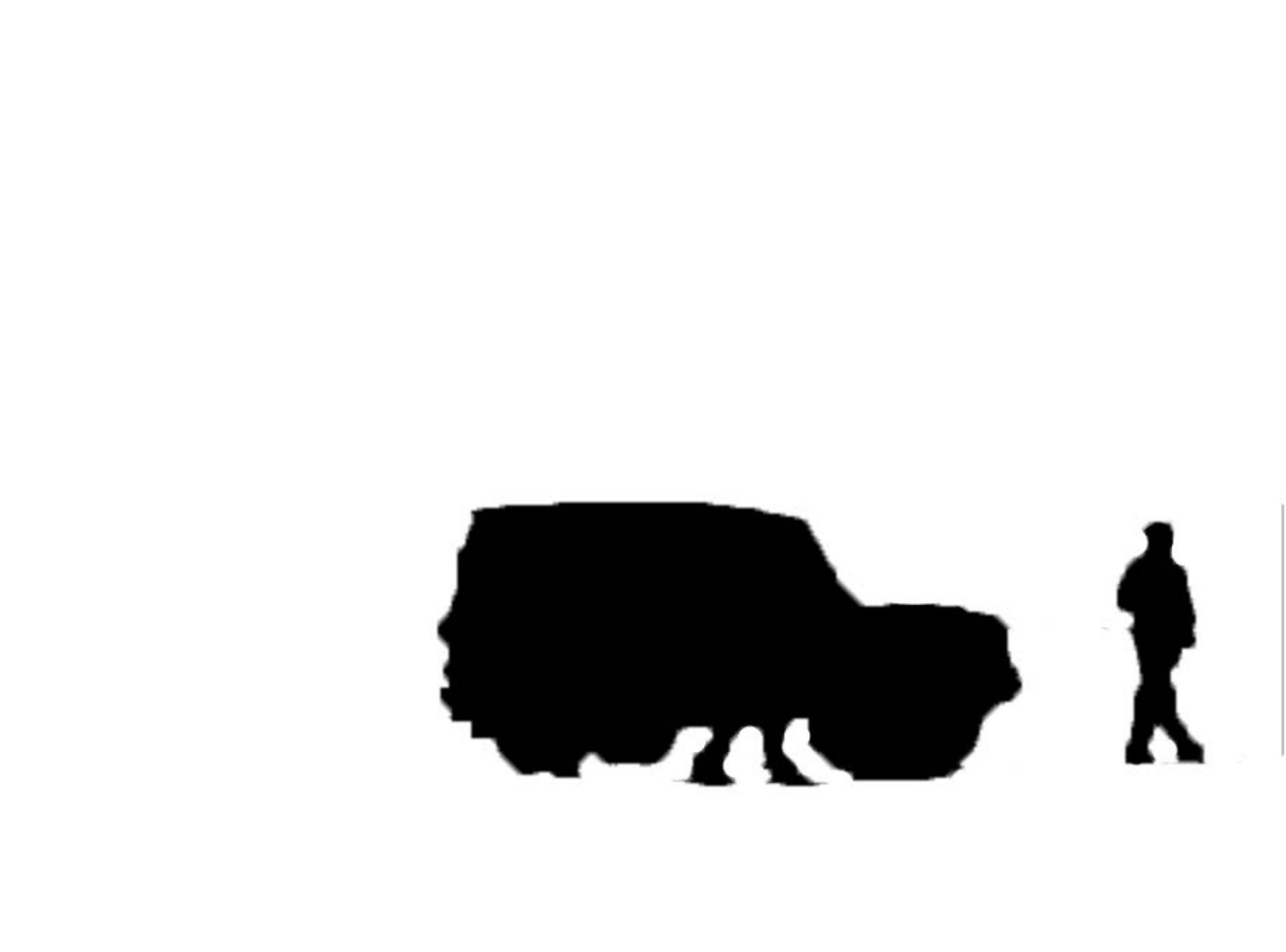}
		&\includegraphics[width=0.074\textwidth]{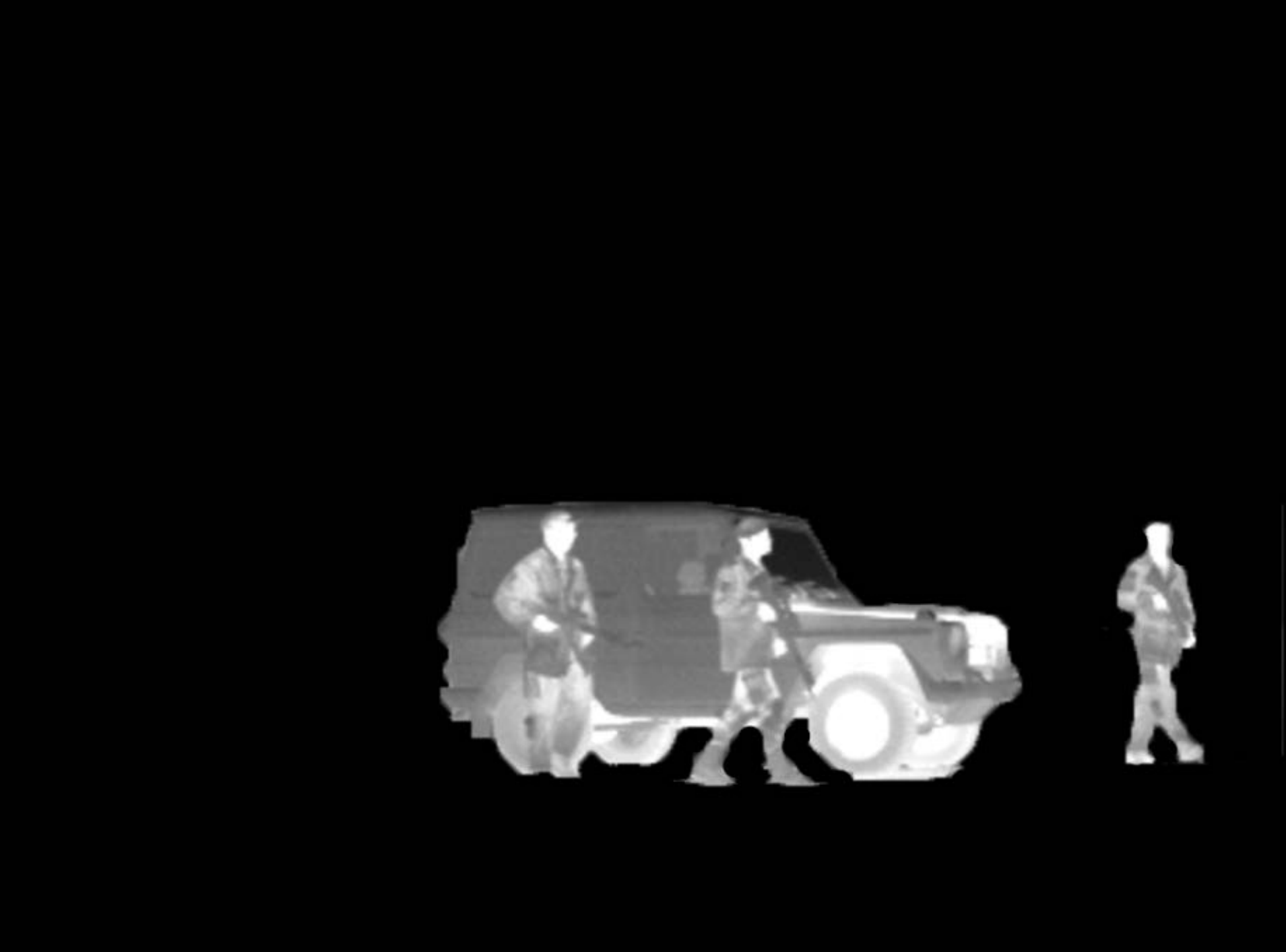}
		&\includegraphics[width=0.074\textwidth]{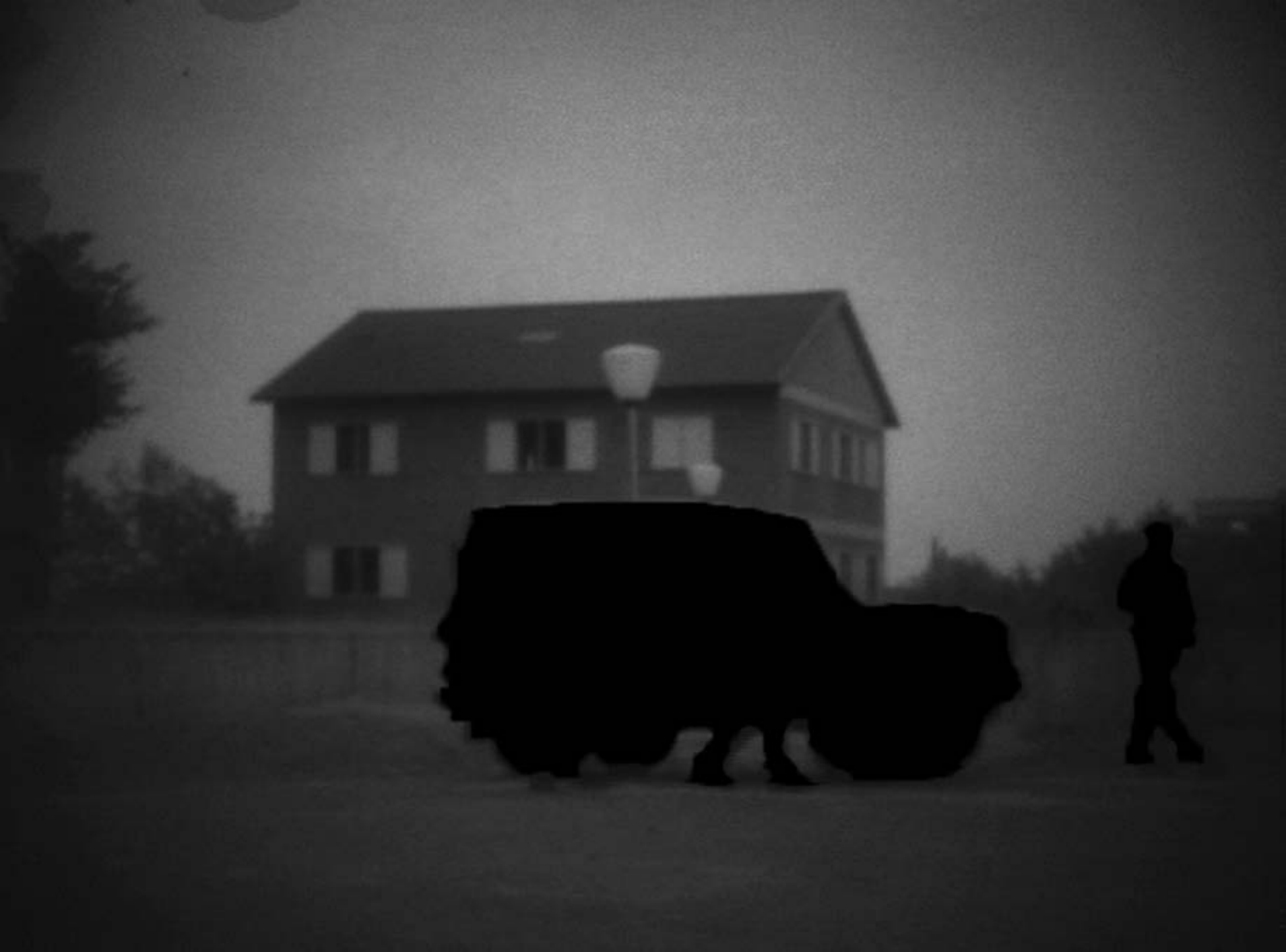}
		\\
		\includegraphics[width=0.074\textwidth]{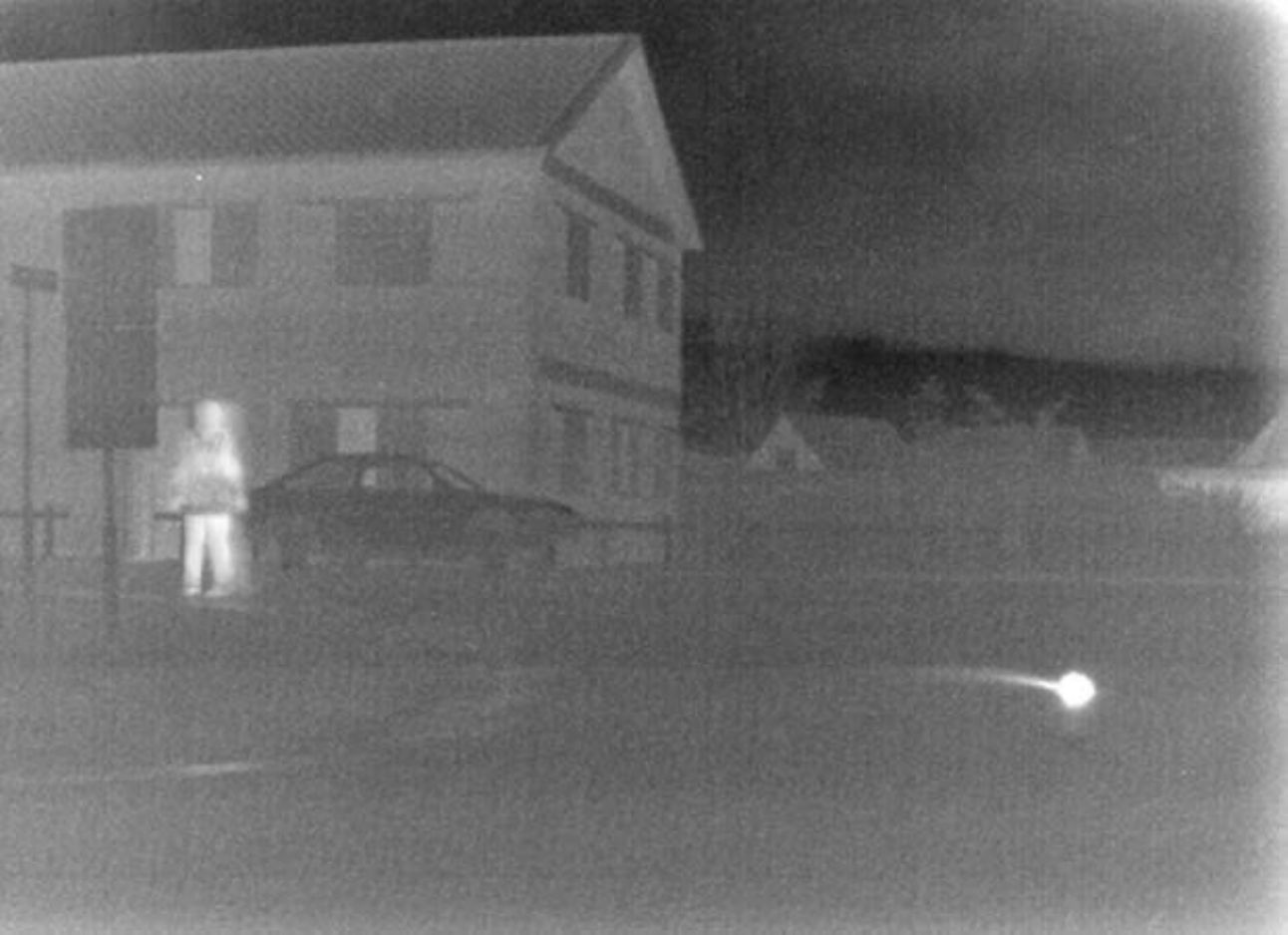}
		&\includegraphics[width=0.074\textwidth]{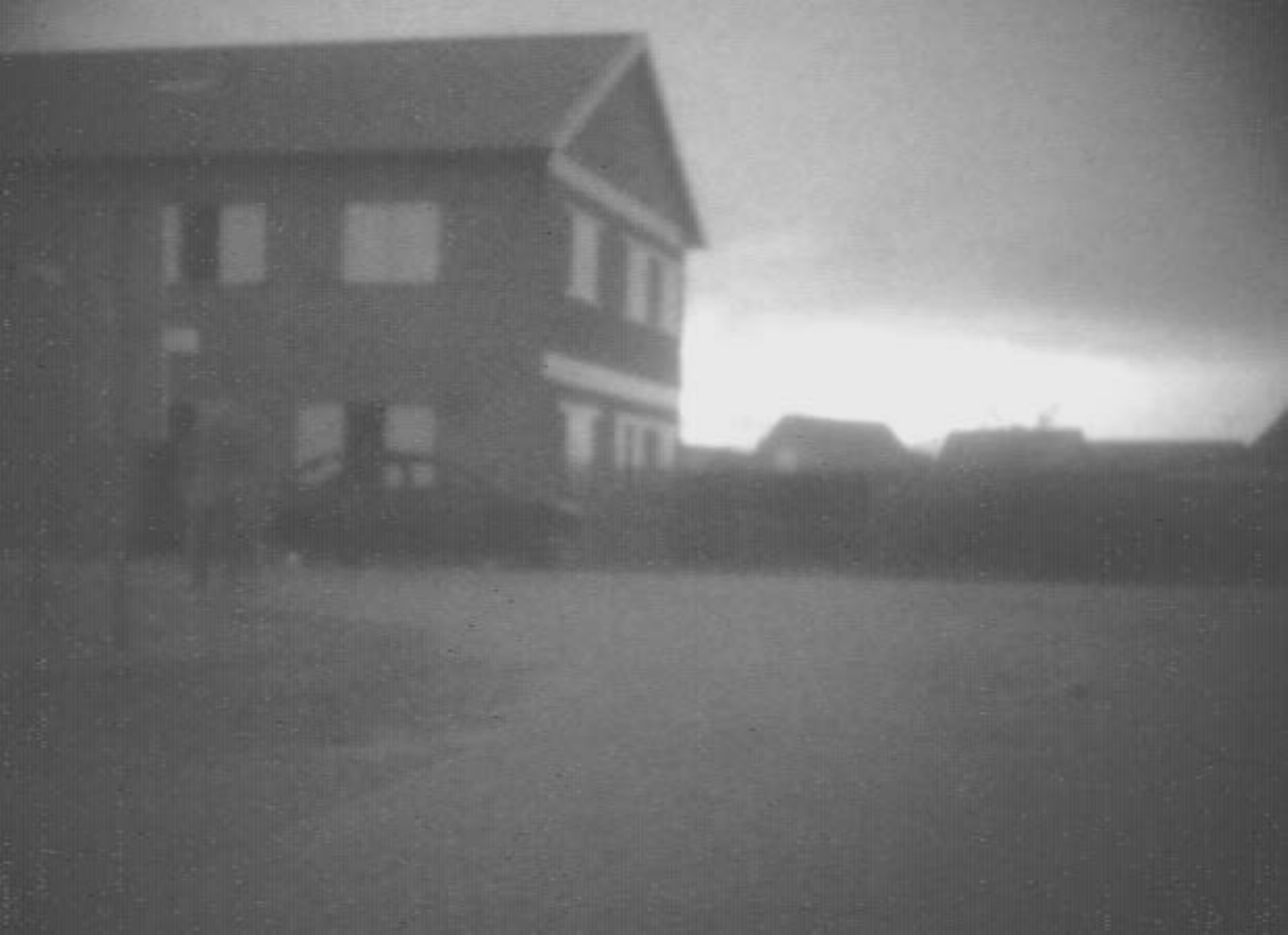}
		&\includegraphics[width=0.074\textwidth]{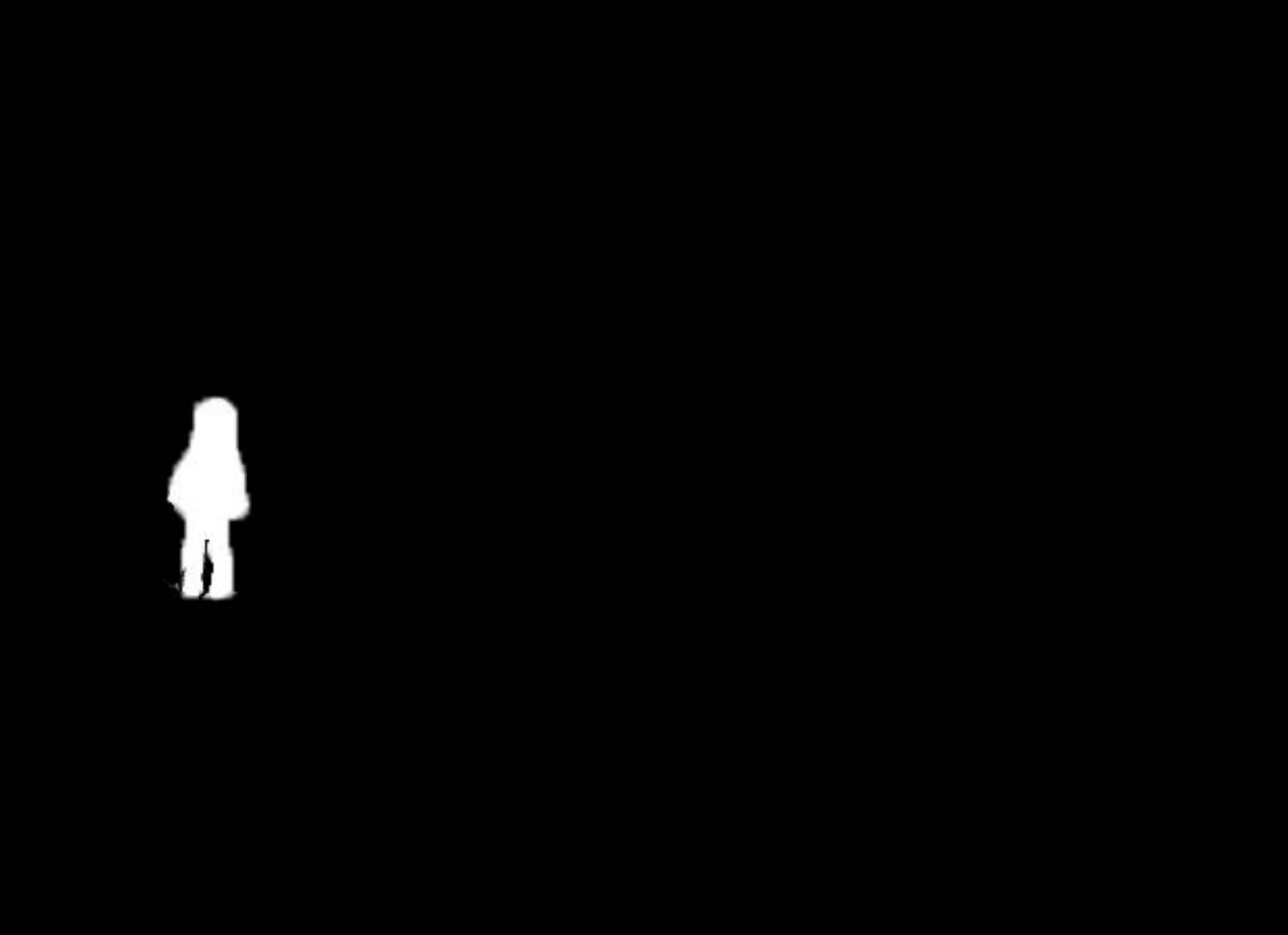}
		&\includegraphics[width=0.074\textwidth]{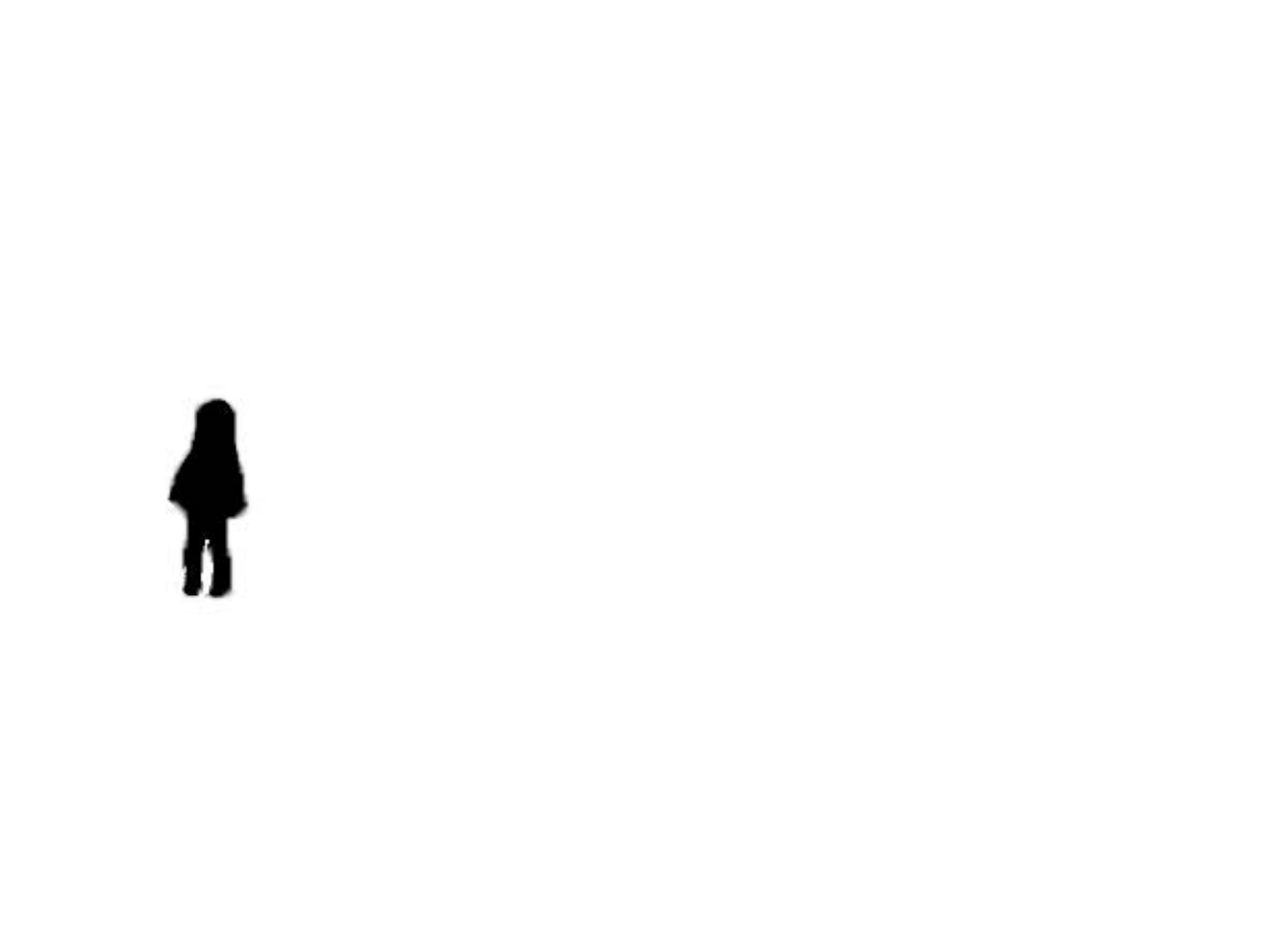}
		&\includegraphics[width=0.074\textwidth]{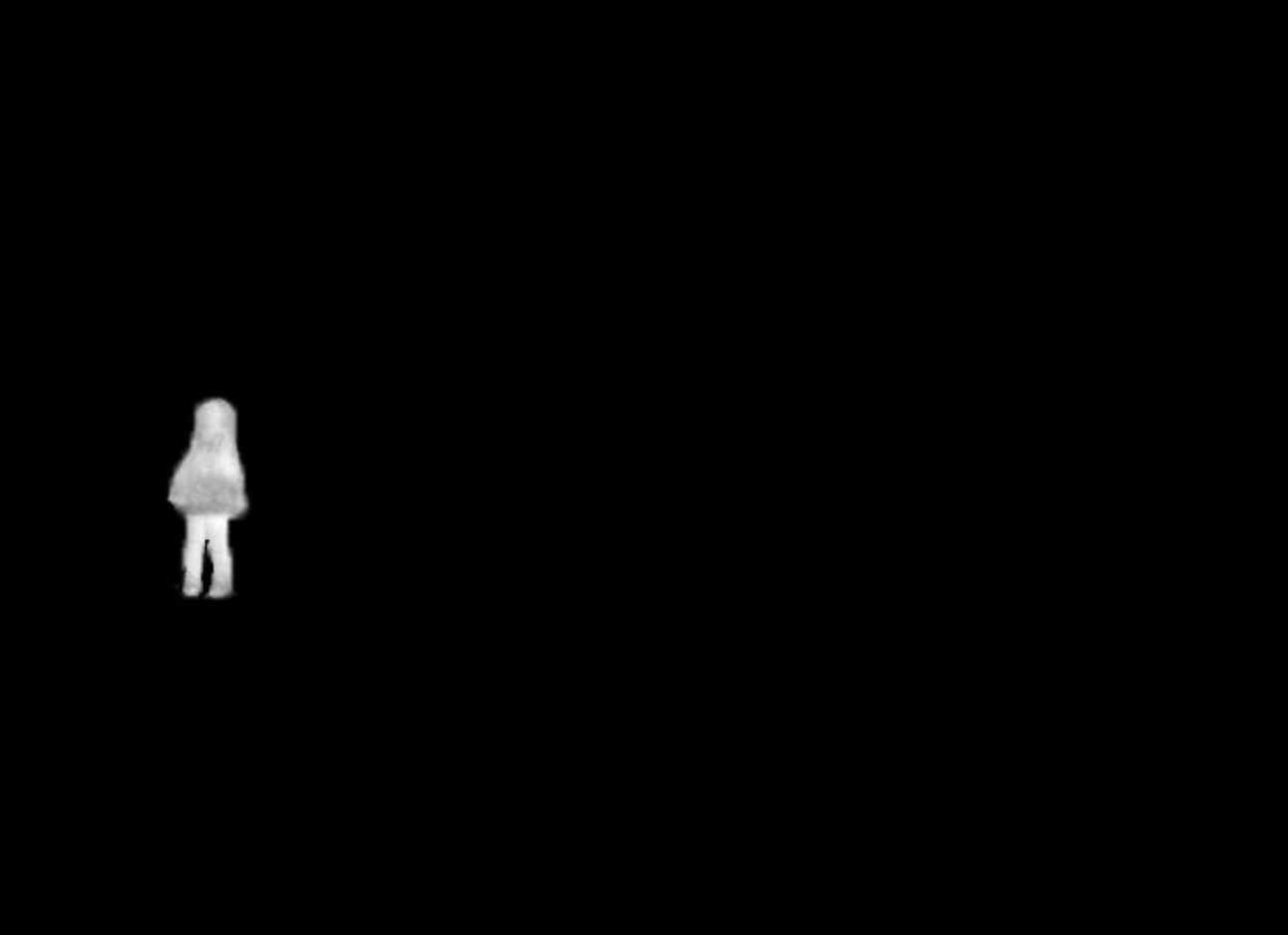}
		&\includegraphics[width=0.074\textwidth]{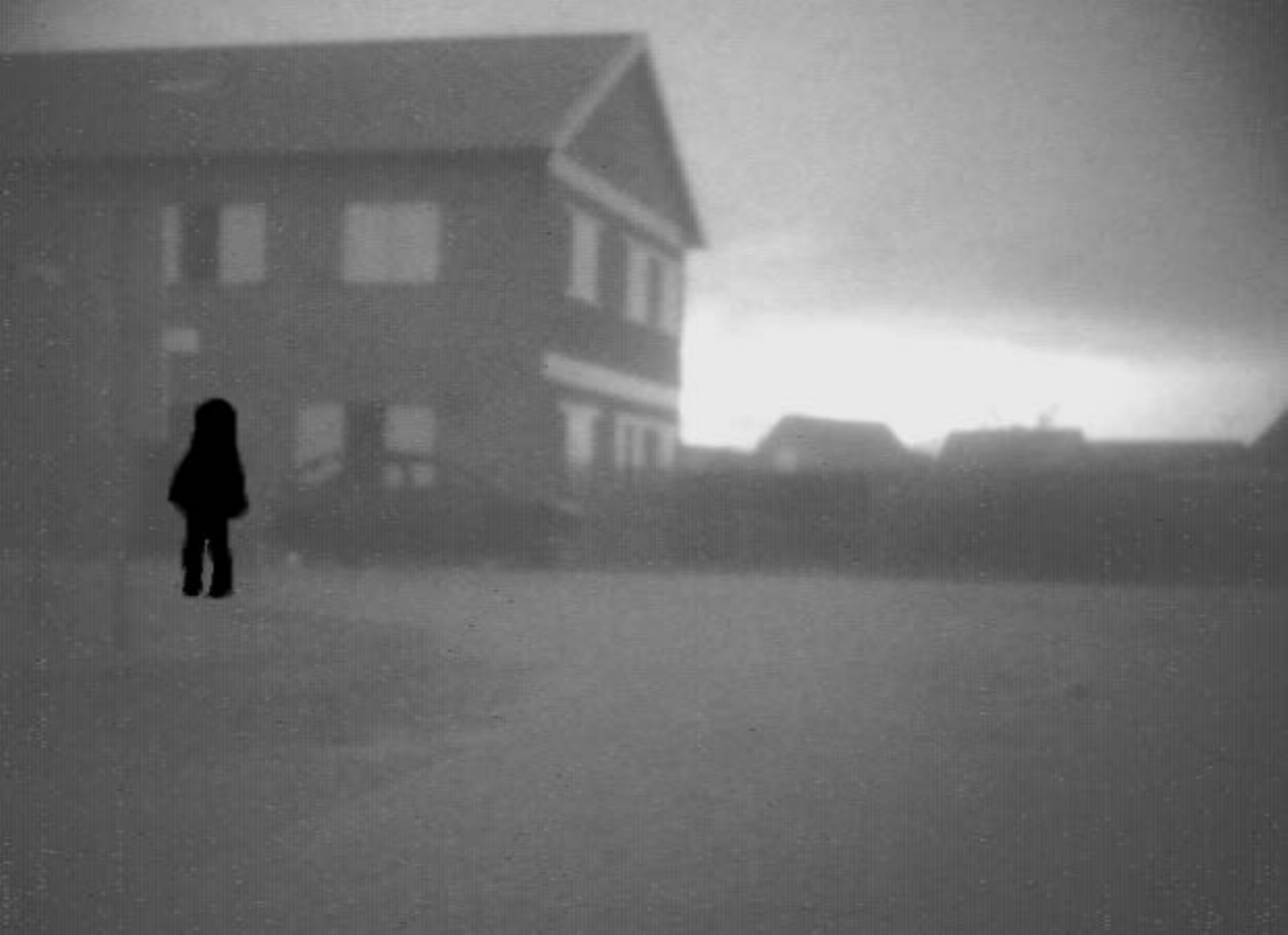}
		\\
		$\rm{I}_{R}$&$\rm{I}_{V}$&$\mathcal{M}$&$\mathcal{\overline{M}}$&\small $\rm{I}_{R}\odot\mathcal{M}$&\small $\rm{I}_{V}\odot\mathcal{\overline{M}}$
	\end{tabular}
	\caption{Typical examples of salient mask $\mathcal{M}$ in TNO dataset.}
	\label{fig:M}
\end{figure}

\begin{figure*}
	\centering
	\setlength{\tabcolsep}{1pt} 
	
	\includegraphics[width=0.98\textwidth,height=0.22\textheight]{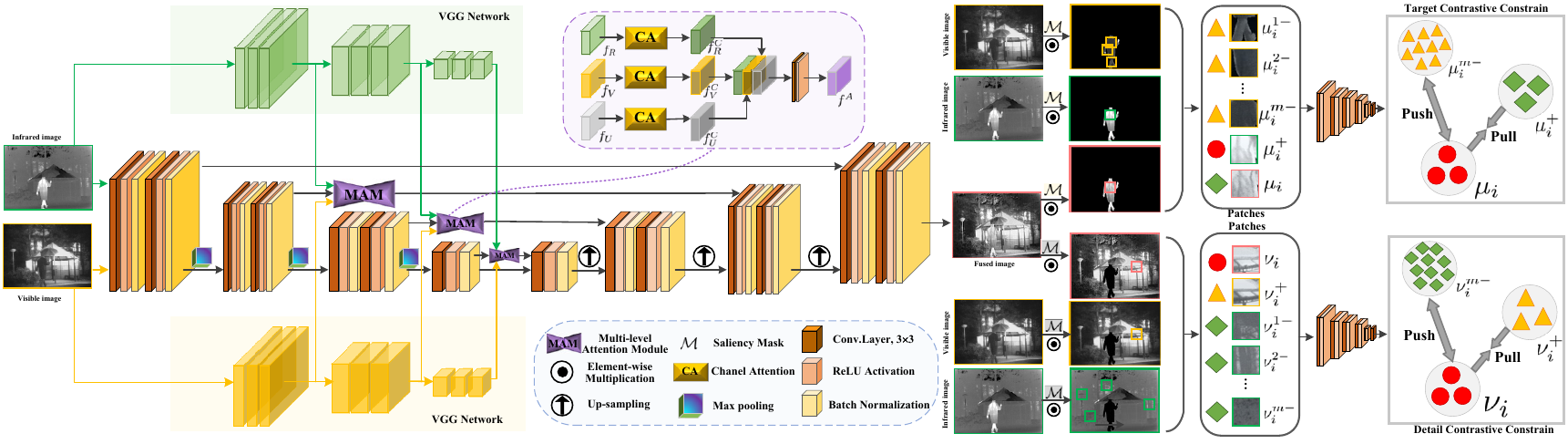}
	
	\caption{Overall architecture of CoCoNet.}
	\label{fig:workflow} 
\end{figure*}

	\subsection{Attention Mechanism in Deep learning}
	The attention mechanism drives from Natural Language Processing~(NLP)~\cite{attentionNLP} and has been successfully applied in CNN-based computer vision tasks~\cite{attentionSS,attentioni2017need}, \emph{e.g.,} saliency object detection~\cite{attention2021bilateral}, semantic segmentation~\cite{attentionSS,attentionSS2}, image enhancement~\cite{attentionlowlight} and image restoration~\cite{attentionrestoration}. The attention mechanism is explained by the human biological visual system that allow human to capture ROI (region of interest) information and ignore other unimportant ones more easier.
	
	To explore salient information in both foreground and background regions, \cite{attention2021bilateral} introduced a Bilateral Attention Network (BiANet) for the RGB-D saliency object detection task, in which complementary attention mechanism can coordinate refine the uncertain details between fore-/background regions. \cite{liu2021halder} proposed a hierarchical attention-guided module for multi-exposure image fusion, which allows the network can capture the most important information in extremely exposure regions.
	\subsection{Contrastive Learning}
	Contrastive Learning has obtained comprehensive attention in the self-supervised learning area~\cite{clmoco,clhigh-levev}. Compared with other techniques that employ a fixed target, contrastive learning aims to maximize mutual information by employing positive and negative samples. More precisely, the learned model needs to pull the anchor close to positive samples while pushing it far away from negative ones. Previous work applies contrastive learning to both high-level and low-level vision tasks~\cite{clmoco,cldehaze,cldetection}, \emph{e.g.,} object detection~\cite{cldetection}, image dehazing~\cite{cldehaze}, image super-resolution~\cite{clsuper} and multi-exposure image fusion~\cite{liu2023holoco}, achieving SOTA performance. In this study,  we show that contrastive learning can be employed to learn representations of salient targets and textural details for effective fusion.
}
\section{The Proposed Method}
In this section, we first describe the motivation of CoCoNet, then introduce the designed loss function, i.e.,  coupled contrastive constraint and self-adaptive learning loss. Afterwards, detailed network architecture and multi-level feature ensemble module are presented. Finally, we describe details for extending CoCoNet to medical image fusion.
\subsection{Motivation}

We consider the goal of IVIF as preserving complementary information while removing redundancy from two modalities~\cite{Ma2022SwinFusion,liu2022target}. However, no supervision signals exist as explicit guidance for the IVIF task. To address this issue, existing work designs only structure or pixel-level terms in the loss function, which does not ensure that the model is optimized by valid features, e.g., blurry textures in most infrared images or dark targets in the visible images should not be the valid supervision signal. Consequently, targets/details of the fused result often contain unpleasant artifacts. In this study, we argue that there is intrinsic feature guidance in the source image pair, i.e., salient thermal targets in the infrared and abundant textural details in the visible. With manual prior involved, we design two loss terms based on contrastive learning to impose explicit constraints for prominent targets and vivid textures. Besides, most fusion methods use skip connections to avoid information loss in the fusion process. However, these direct skip connections may also introduce unfiltered information, bringing noises to the fused image. Moreover, hand-crafted trade-off hyper-parameters in the loss function are usually difficult to adjust, posing latent threats to the model's flexibility to specific data.

Therefore, we introduce a coupled contrastive learning network to alleviate these issues. By elaborating the constructed coupled contrastive constraints with self-adaptive weights in the loss function, we are capable of fusing the most significant information and automatically determining their individual weights in the loss function. A multi-level attention module is also incorporated to learn comprehensive feature representation. 
\subsection{The Proposed CoCoNet}
\subsubsection{Coupled Contrastive Learning}
Inspired by previous work on contrastive learning, we propose a coupled contrastive regularization for IVIF, based on two pairs of constraints, \emph{i.e.}, target constraint and detail constraint. Different from supervised learning with explicit guidance of reference images, there are no clear indications of positive and negative samples for infrared and visible image fusion. Hence, the core of applying contrastive learning lies in determining how to construct positive and negative pairs. In this paper, we argue that the desirable features served as positive and negative samples are included in source images. Concretely, for an infrared image, its foreground salient thermal target is of more interest compared to the rest parts. Similarly, in a visible image, the background vivid textural details are of higher demand compared to its dark foreground part. We utilize this prior to construct contrastive pairs so that our model can learn to distinguish salient targets with high pixel intensities and background textual details. Finally, our model becomes aware of typical features from source images, achieving complementary fusion. 

We target to achieve two objectives based on two groups of constraints for a fused result: to improve the saliency of foreground objects, the corresponding targets from the infrared image are utilized as positive samples while the corresponding regions from the visible image are utilized as negative samples. Meanwhile, we set the visible image as positive while the infrared image as negative samples when we need to preserve clear background details in the fused result.

In order to introduce artificial prior to maximize the above objectives, we manually generated masks for corresponding image pairs based on typical infrared images captured in TNO dataset. As shown in
Figure~\ref{fig:M}, let $\mathcal{M}$ denote the salient mask of foreground and~$\mathcal{\overline{M}}$ represents the salient mask of background~( $\mathcal{\overline{M}}=1-\mathcal{M}$). This explicit guidance forces the model to distinguish saliency and textural details, and be able to extract and fuse them from both visible and thermal sensors.

To this end, the positive and negative samples for improving foreground saliency, termed as target constraint, should be ${\rm I}_{R}\odot \mathcal{M}$ and  ${\rm I}_{V} \odot \mathcal{M}$. For the latent feature space, we select the commonly used VGG-19~\cite{simonyan2014very}, denoted as $G$, with pre-trained weights.  We formulate the loss function of this goal as follows:

\begin{equation}
	\mathcal{L}_{ir} = \sum_{i=1}^{N}w_{i}\frac{\parallel{\mu_i}-{\mu_i}^{+}\parallel_{1}}{\sum_{m}^{M}\parallel{\mu_i}-{\mu}_i^{m-}\parallel_{1}},
	\label{eq5}
\end{equation} 
where $N$ and $M$ are the number of VGG layers and negative samples for each positive sample, respectively. $\mu_i$ denotes the foreground feature of the fused image, which is defined as $G_i({\rm I}_{F}\odot\mathcal{M})$. $\mu_i^{+}$ and $\mu_i^{m-}$ are the positive and negative samples, formulated as $\mu_i^{+}=G_i({\rm I}_{R}\odot\mathcal{M})$, $\mu_i^{m-}=G_i({\rm I}^{m}_{V}\odot\mathcal{M})$, respectively. $m$ means the $m$th negative sample. $\parallel \cdot \parallel_{1}$ denotes the $\ell 1$ norm.

Likewise, for the background part, we hope to retain more vivid details from visible images, treating background of the infrared image as negative samples, while the visible image background as positive samples. Therefore, the object function for detail constraint can be given as:
\begin{equation}
	\mathcal{L}_{vis} = \sum_{i=1}^{N}w_{i}\frac{\parallel{\nu_i}-{\nu_i}^{+}\parallel_{1}}{\sum_{m}^{M}\parallel{\nu_i}-{\nu_i}^{m-}\parallel_{1}},
	\label{eq5}
\end{equation} 
where $\nu_i$ denotes the background feature of the fused image, which is defined as $G_i({\rm I}_{F}\odot\mathcal{{\overline{M}}})$. $\nu_i^{+}$ and $\nu_i^{m-}$ are the positive and negative samples, formulated as $\nu_i^{+}=G_i({\rm I}^{m}_{V}\odot\mathcal{{\overline{M}}})$, $\nu_i^{m-}=G_i({\rm I}_{R}\odot\mathcal{{\overline{M}}})$, respectively. $m$ means the $m$th negative sample. Illustration of the contrastive learning process is provided in Figure~\ref{fig:workflow}.

\subsubsection{Self-adaptive Learning Weight}
Image fusion targets to provide an information-abundant image with
sufficient details and balanced intensities by combining favorable features of source images. For the infrared and visible fusion task, we learn to minimize the similarity of source images and the fused image. The loss function is mainly composed of two parts, \emph{i.e.}, the structure similarity loss and the intensity similarity loss, which can be define as:
\begin{eqnarray}
\mathcal{L_P} = \alpha \mathcal{L_S}+\mathcal{L_N},
\end{eqnarray}
where $\alpha$ is a tuning parameter, $\mathcal{L_S}$ and $\mathcal{L_N}$ denote structure similarity loss and intensity similarity loss. $\mathcal{L_S}$ is measured by the structure similarity index measure (SSIM)~\cite{wang2004image}, which is widely used to indicate the difference of images based on similarities of contrast, light, and structure. It is given as follows:
\begin{eqnarray}
\mathcal{L_S} = \sigma^a(1-\mathcal{S}(\mathbf{I}_V,\mathbf{I}_F))+\sigma^b(1-\mathcal{S}(\mathbf{I}_R,\mathbf{I}_F)),
\label{eq:sig}
\end{eqnarray}
where $\mathcal{S}(\cdot)$ denotes SSIM value. 

$\mathcal{L_N}$ is adopted to strengthen the constraints on the differences of intensity distributions, formulated as:
\begin{eqnarray}
\mathcal{L_N} = \gamma^a\parallel \mathbf{I}_V-\mathbf{I}_F\parallel_2+\gamma^b\parallel \mathbf{I}_R-\mathbf{I}_F\parallel_2,
\label{eq:gamma}
\end{eqnarray}
where $\parallel\cdot\parallel_2$ is Mean Square Error (MSE).

In Equation~\ref{eq:sig} and Equation~\ref{eq:gamma}, $\sigma$ and $\gamma$ are two pairs of proportional weights that balance the proportion of the visible image and infrared image. $\sigma$ and $\gamma$ consists of~$\{\sigma^a, \sigma^b\}$ and $\{\gamma^a, \gamma^b\}$, respectively.
	They are empirically set to fixed values in existing methods~\cite{PMGI}. However, a fixed manner is insufficient to fully exploit the source image features. Therefore, we design a self-adaptive loss to consider the data characteristics by optimizing the image-specific weights $\sigma$ and $\gamma$.
	
	For one thing, we expect the fused image to retain significant textures (\emph{e.g.,} structural information). Average Gradient (AG) is applied to optimize the weight parameter $\sigma$ of SSIM loss. The equation of AG is given as:
	\begin{eqnarray}
	{\rm AG} =\mathcal{G}(\mathbf{I}_F)= \frac {1}{HW}(\parallel \nabla_h \mathbf{I}_F \parallel_1 + \parallel \nabla_v \mathbf{I}_F \parallel_1),
	\label{AG}
	\end{eqnarray}
	where the $\nabla_h \mathbf{I}_F$ and $\nabla_v \mathbf{I}_F$ represent the first-order differential of the fused image from horizontal and vertical direction, respectively. $H$ and $W$ are the height and width. $\parallel \cdot \parallel_{1}$ denotes $\ell 1$ norm. Since AG reflects the basic intensity change of an image, it is considered to well match the goal of SSIM loss, i.e., to constrain the fused image from structure similarity. Thus, $\sigma$ can be determined by the following equation:
	\begin{eqnarray}
	{\rm \sigma}^{a}, {\rm \sigma}^{b} = \frac {e^{\mathcal{G}(\mathbf{I}_V)}}{e^{\mathcal{G}(\mathbf{I}_V)}+e^{\mathcal{G}(\mathbf{I}_R)}},\frac {e^{\mathcal{G}(\mathbf{I}_R)}}{e^{\mathcal{G}(\mathbf{I}_V)}+e^{\mathcal{G}(\mathbf{I}_R)}},
	\label{sigma}
	\end{eqnarray}
	For another, to fuse images with high contrast, image Entropy (EN) is employed to update weight parameter $\gamma$ of the MSE loss. EN is formulated as follows:
	\begin{equation}
	{\rm EN}=\mathcal{E}(\mathbf{I}_F) = -\sum_{x=0}^{L-1}p_{x}{\rm{log_{2}}}p_{x},
	\label{EN}
	\end{equation}
	where $L$ denotes the grey level of a given image, $p_{x}$ is the probability that a pixel lies in the corresponding grey level. As Equation~\ref{EN} suggests, EN measures the amount of information of an image and is computed in pixel level, it is closely correlated to the MSE constraint, which is also a pixel-level constraint. Hence, the modality with a higher EN (i.e., more information) should deserve a higher wight for maximizing meaningful features. Therefore, the $\gamma$ can be updated with the following rule:
	\begin{eqnarray}
	{\rm \gamma}^{a}, {\rm \gamma}^{b} = \frac {e^{\mathcal{E}(\mathbf{I}_V)}}{e^{\mathcal{E}(\mathbf{I}_V)}+e^{\mathcal{E}(\mathbf{I}_R)}},\frac {e^{\mathcal{E}(\mathbf{I}_R)}}{e^{\mathcal{E}(\mathbf{I}_V)}+e^{\mathcal{E}(\mathbf{I}_R)}},
	\label{sigma}
	\end{eqnarray}
	
	Consequently, combining all the restrictions above, we give the following loss function to guide the learning process:
	\begin{equation}
	\mathcal{L}_{total} = \mathcal{L_P} + \mathcal{L}_{ir} + \mathcal{L}_{vis},
	\end{equation}
	where $\mathcal{L_P}$ is the self-adaptive loss, $\mathcal{L}_{ir}$ and $\mathcal{L}_{vis}$ are the two pairs of contrastive loss respectively.

\subsection{Network Architecture}
As shown in Figure~\ref{fig:workflow}, each convolution block consists of two groups of 3$\times$3 convolutional layers followed by batch normalization and LeakyRuLU. The feature maps from each depth layers
can be represented as: $f_{U0}$, $f_{U1}$, $f_{U2}$, $f_{U3}$ from layers with 32, 64, 128, 256 channels respectively. For the multi-level attention module, we select two VGG-19 with pre-trained weights as our backbone. It takes the visible image and infrared image as inputs in a separate manner, as an attempt to make full use of high level features of the source image. Infrared features obtained from the backbone are denoted as $f_{R1}$, $f_{R2}$, $f_{R3}$, which come from layers of 64, 128 and 256 channels, respectively. Likewise, visible features extracted from the backbone are represented as $f_{V1}$, $f_{V2}$, $f_{V3}$ from the corresponding layers.

To employ more high-level features into the fused image, we propose a multi-level attention module~(MAM) to achieve comprehensive feature representation from the source images. Meanwhile, we expect this attention to intensify the extracted features by a global enhancement. Based on the obtained features $f_{U}$, $f_{R}$ and $f_{V}$ above, a channel attention is first performed:
\begin{equation}
	\begin{aligned}
		f_U^{C} = {\mathbf {CA}}(f_U), f_R^{C} = {\mathbf {CA}}(f_R), f_V^{C} = {\mathbf {CA}}(f_V),
	\end{aligned}
\end{equation} 
where $\mathbf {CA}$ denotes channel attention, which will be described in the following part. To fuse these features, we apply  convolution operations for each group of features:

\begin{equation}
	\begin{aligned}
		f_{1}^A = \mathbf{Conv}(\mathbf{Concate}(f_{U1}^{C}, f_{R1}^{C}, f_{V1}^{C}))&...\\
		f_{n}^A = \mathbf{Conv}(\mathbf{Concate}(f_{Un}^{C}, f_{Rn}^{C}, f_{Vn}^{C}))&,
	\end{aligned}
\end{equation}
where $\mathbf {Conv}$ represents a convolution layer with 3$\times$3 kernels, and $\mathbf {Concate}$ means concatenation.

\begin{figure}[!htb]
	\centering
	\setlength{\tabcolsep}{1pt} 
	
	\includegraphics[width=0.48\textwidth]{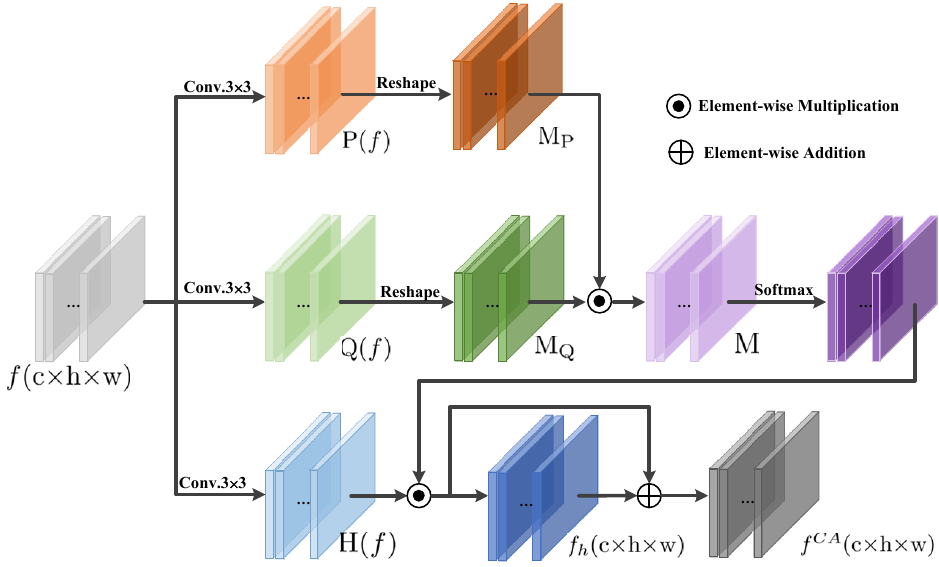}
	
	\caption{Architecture of channel attention.}
	\label{fig:CA} 
\end{figure}

The detailed architecture of channel attention is illustrated in Figure~\ref{fig:CA}. Given a feature ${f}$ of dimension ${\rm {R}}^{C\times H \times W}$, we first use a convolution layer to generate three components ${\mathbf P(f), \mathbf Q(f), \mathbf H(f)}$, and reshape ${\mathbf P(f), \mathbf Q(f)}$ from ${\rm {R}}^{C\times H \times W}$ to ${\rm {R}}^{C\times HW}$, obtaining ${\mathbf M_{P}, \mathbf M_{Q}}$. Then a matrix multiplication is applied on $\mathbf M_{P}$ and the transpose of $\mathbf M_{Q}$ followed by a softmax layer to compute the attention feature map $\mathbf {M}\in {\rm {R}}^{C\times C}$. Thereafter, a matrix multiplication is performed between the transpose of $\mathbf H(f)$ and $\mathbf M$. The result is finally reshaped and added back to the source image $\mathbf H_{f}$.

\subsection{Extension to Medical Image Fusion}
In this section, we extend our CoCoNet to fuse the medical images, \emph{e.g.,} MRI and PET image fusion, MRI and SPECT image fusion. The PET and SPECT images are regarded as pseudo-color images. We first transform them to color images, and then apply CoCoNet to fuse the MRI image and intensity component of PET and SPECT image, respectively.

\subsubsection{Medical Image Fusion Background}
With the rapid development of clinical requirement, a series of medical imaging technologies, \emph{e.g.,} X-ray, computed tomography (CT), MRI, PET and SPECT, has been introduced in the past few decades. However, each imaging technology has its advantages and limitations. For instance, X-ray is an electromagnetic wave with extremely high frequency, short wavelength, and high energy, which has strong penetrability. It has been widely used in fluoroscopy before diagnosis or surgery. Compared with the conventional X -ray photography, CT can provide detect the slight difference of the bone density with high resolution, but it has limitation in representing tissue characterization.
Apart from that, MRI can not only display the morphological structure of the organizational organs, but also display the functional conditions and biochemical information of certain organs. However, MRI lacks of the description of the soft tissue activity. In contrast, both PET and SPECT are functional imaging, which present the difference in the intensity of the human tissue activity according to the difference in the concentration of the aggregation. Evidently, each imaging modality unavoidably has its respected characteristics and inherent drawbacks. It is worthwhile to combine the advantages from different modality images and then provide an informative and complementary fused image for the clinical diagnosis.  

In recent years, a series of hybrid imaging technologies, \emph{e.g.,} CT-MRI, MRI-PET and MRI- SPECT, have been introduced in our daily life. In this paper, we take two typical medical image fusion, \emph{i.e.,} MRI-PET and MRI-SPECT as an example and apply CoCoNet to solve this issue. As mentioned above, both PET and SPECT images can provide functional and metabolic information, which are widely used for analyzing the functions or metabolic conditions of each organs. These obtained images are rich in color but lack in resolution. In contrast, MRI image can better portray soft tissue structures in organs. Usually, it has high spatial resolution. Therefore, by integrating advantages from each modality image, we can obtain complementary and comprehensive information in a single image. 

\begin{figure}
	\centering
	\setlength{\tabcolsep}{1pt}
	\begin{tabular}{ccccc}
		\includegraphics[width=0.09\textwidth]{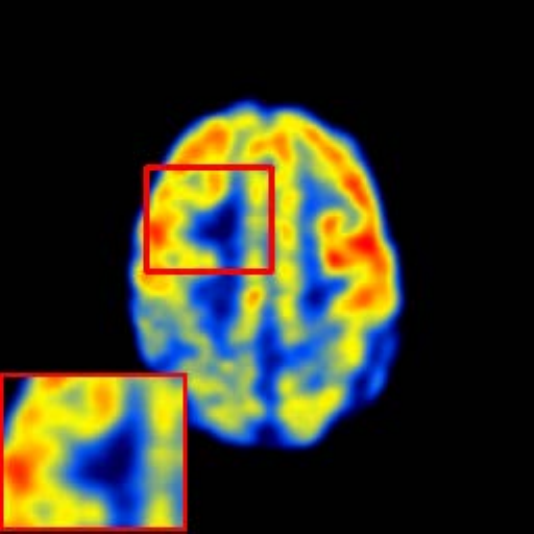}
		&\includegraphics[width=0.09\textwidth]{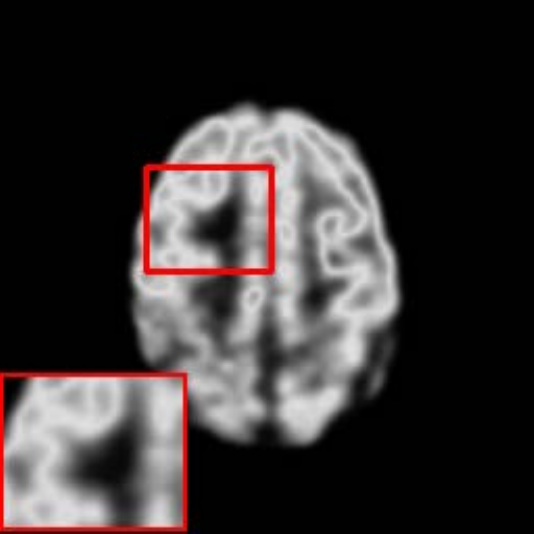}
		&\includegraphics[width=0.09\textwidth]{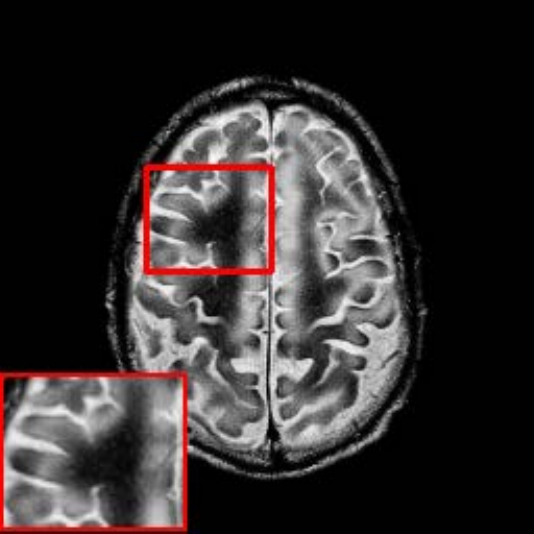}
		&\includegraphics[width=0.09\textwidth]{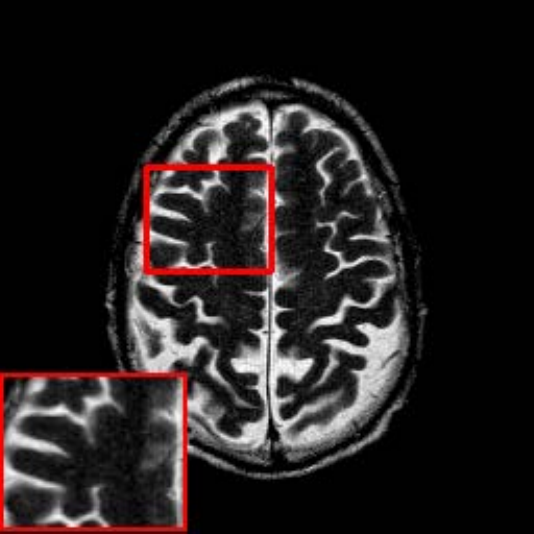}
		&\includegraphics[width=0.09\textwidth]{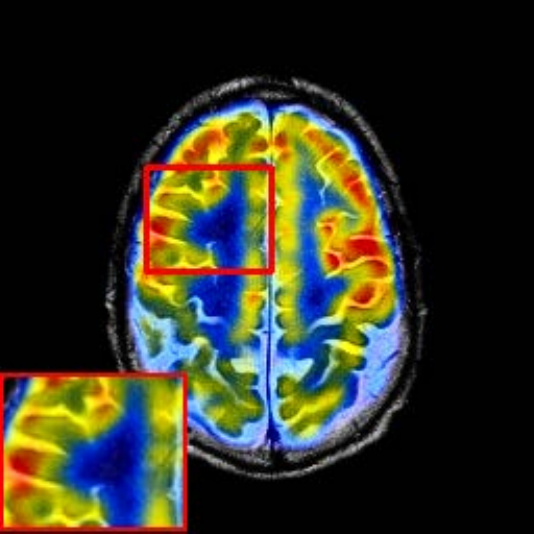}
		\\
		\includegraphics[width=0.09\textwidth]{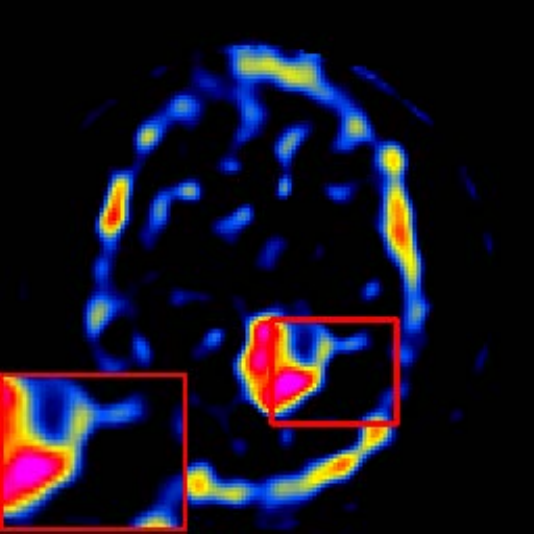}
		&\includegraphics[width=0.09\textwidth]{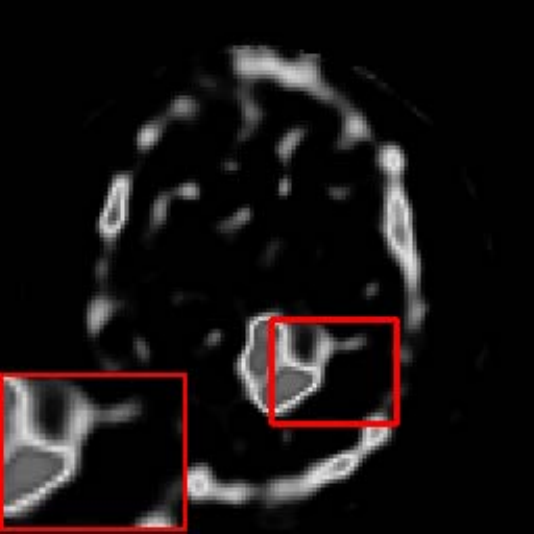}
		&\includegraphics[width=0.09\textwidth]{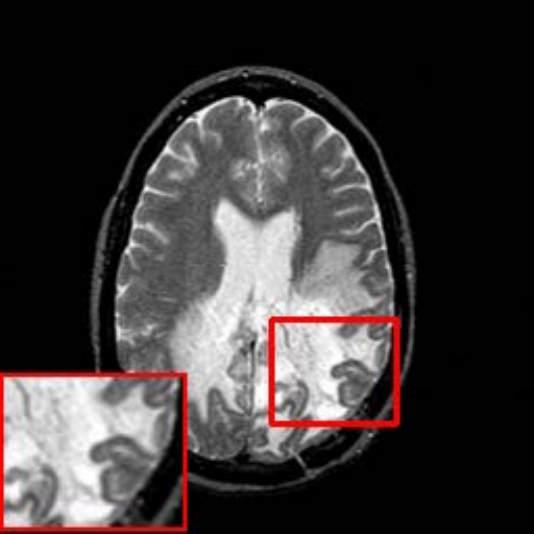}
		&\includegraphics[width=0.09\textwidth]{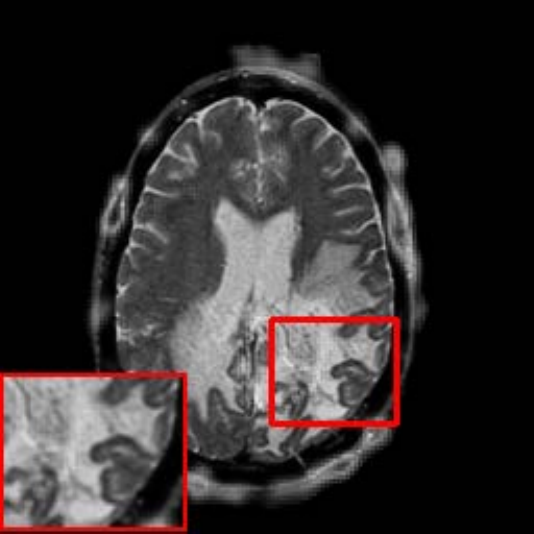}
		&\includegraphics[width=0.09\textwidth]{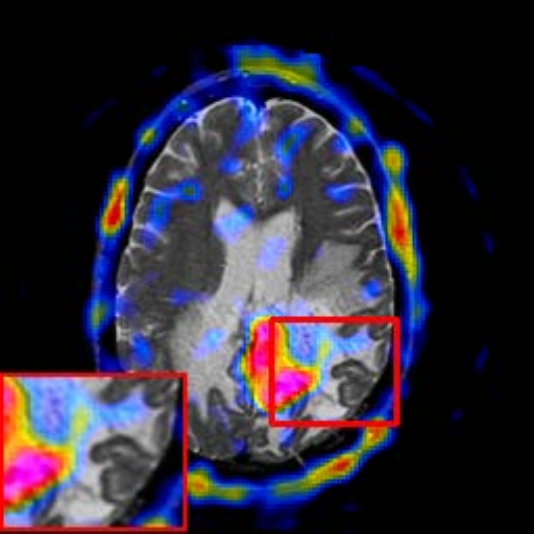}
		\\
		(a)&(b)&(c)&(d)&(e)
	\end{tabular}
	\caption{Schematic illustration of MIF tasks, where (a) pet/spect image, (b) Luminance~(Y) channel of pet/spect image, (c) mri image, (d) Luminance channel~(Y) of fused image, (e) fused image.}
	\label{fig:mif5x2}
\end{figure}

In most of cases, as shown in Figure~\ref{fig:mif5x2}(a), both PET and SPECT are shown in pseudo-color, in which the color are represented as functional information. For the fused image, the color information should in line with the PET or SPECT images. To this end, we decouple the PET and SPECT image into YCbCr color space with three channels. Then, we employ Y channel (the luminance channel shown in Figure~\ref{fig:mif5x2}(b)) to fuse with the MRI image PET and SPECT use the intensity information of Y channel to represent the feature distribution, which is similar to the infrared image, while MRI image shown in Figure~\ref{fig:mif5x2}(c) has rich texture organization details, which is similar to the visible image.
Therefore, MIF and IVIF have similar task objectives, that is, to reduce spatial detail distortion and color intensity distortion of fused images and corresponding modalities. After getting the fused image(Figure~\ref{fig:mif5x2}(d)) by Y channel and MRI images, other two channels are remained unchanged to recover the color information, as shown in Figure~\ref{fig:mif5x2}(e).

\subsubsection{CoCoNet for Medical Image Fusion}

Building on the unique characteristics of MRI sequence and functional sequences (i.e., PET and SPECT) mentioned above, we can also apply the proposed coupled contrastive learning to integrate desirable features from different medical modalities. We should first define features of interest in MRI and functional (i.e., PET/SPECT) modalities respectively. MRI sequences are rich in soft tissue structure, which provides clear indications of the brain skeleton. On the other hand, To better retain salient structure information from the MRI image, and simultaneously fuse functional information which reflects metabolic activity of organs or tissues, as well as the function and distribution of receptors. Therefore, we are able to clarify the useful features from two modalities as salient structure information from the MRI sequence, and functional indications from the functional sequence. Specifically, to combine useful features from both sides, we propose to impose a MRI-segmented mask and its reverse version on the MRI sequence and functional sequence, respectively, for a better constraint on the features we aim to extract. For a fused image, we hope its salient regions to resemble the corresponded MRI image, but less similar to the same region from its functional counterpart. In the same vein, we expect the other regions in the fused image to be closer to the functional sequence, but further from the MRI counterpart in the latent feature space. In fact, a common problem for fusing MRI and functional sequences is that the texture details in MRIs tend to get covered, thus weakened, after fusing with its functional counterpart. To partially alleviate this issue, our salient masks are first generated by segmenting the MRI image following the research of \cite{li2005mr}, denoted as $\mathcal{M}_{m}$ = $\mathcal{M}$, as shown in Figure~\ref{fig:M2}. We expect the fused image to resemble the MRI sequence under regions masked by $\mathcal{M}_{m}$, for reserving soft tissue textures. At the same time, according to contrastive learning, the functional sequence in the same region serves as negative samples, which aid to stress MRI features. This process can be described as follows:
\begin{equation}
	\mathcal{L}_{mri} = \sum_{i=1}^{N}w_{i}\frac{\parallel{\mu_i}-{\mu_i}^{+}\parallel_{1}}{\sum_{m}^{M}\parallel{\mu_i}-{\mu}_i^{m-}\parallel_{1}},
	\label{eq5}
\end{equation} 
where $N$ and $M$ are the number of VGG layers and negative samples for each positive sample, respectively. $\mu_i$ denotes the MRI structure feature of the fused image, which is defined as $G_i({\rm I}_{F}\odot\mathcal{M}_{m})$. $\mu_i^{+}$ and $\mu_i^{m-}$ are the positive and negative samples, formulated as $\mu_i^{+}=G_i({\rm I}_{MRI}\odot\mathcal{M}_{m})$, $\mu_i^{m-}=G_i({\rm I}^{m}_{Fun}\odot\mathcal{M}_{m})$, respectively. Superscript $m$ means the $m$th negative sample. $\parallel \cdot \parallel_{1}$ denotes the $\ell 1$ norm.

On the contrary, the functional sequence can provide abundant intensity information of the subject's functional activities, e.g., blood flow. To retain most favorable features from functional sequences, we first reverse the MRI-segmented mask $\mathcal{M}_{m}$ to get $\mathcal{M}_{f}$ = 1 - $\mathcal{M}_{m}$. Then $\mathcal{M}_{f}$ is imposed on a functional sequence to extract most informative features, which include metabolic information of a subject. Regions of the fused image under mask $\mathcal{M}_{f}$ should be consistent with the functional image, while less similar to the MRI image. This whole process can be modeled using a twin contrastive learning-based loss function given as follows:

\begin{figure}
	\centering
	\setlength{\tabcolsep}{1pt}
	\begin{tabular}{cccccc}
		\includegraphics[width=0.074\textwidth]{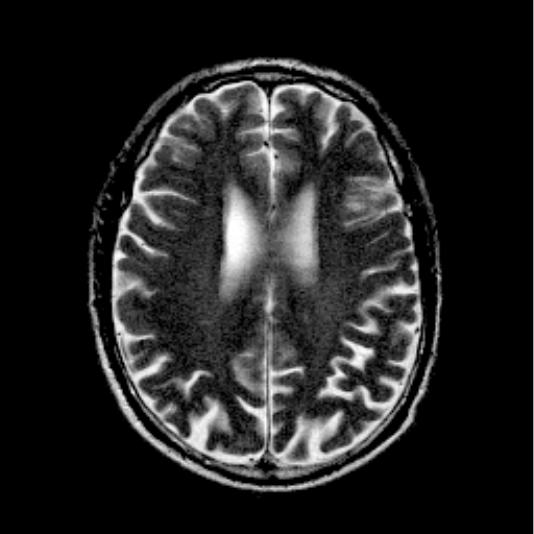}
		&\includegraphics[width=0.074\textwidth]{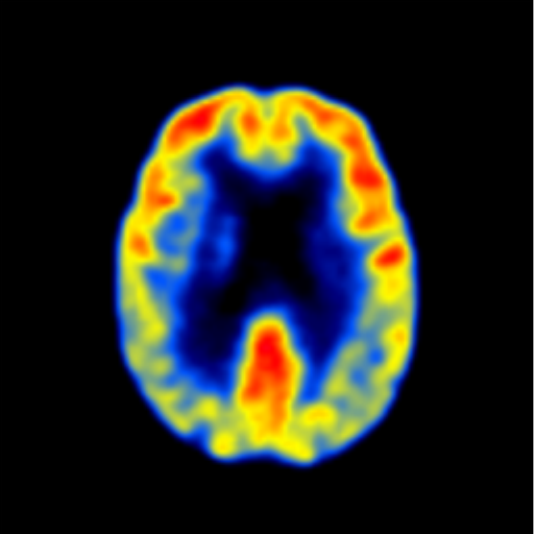}
		&\includegraphics[width=0.074\textwidth]{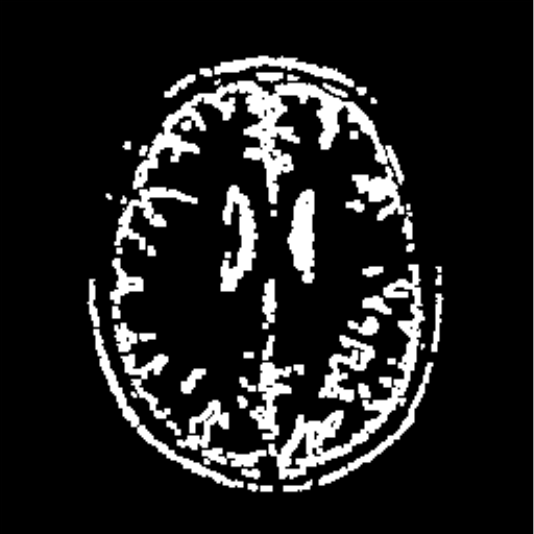}
		&\includegraphics[width=0.074\textwidth]{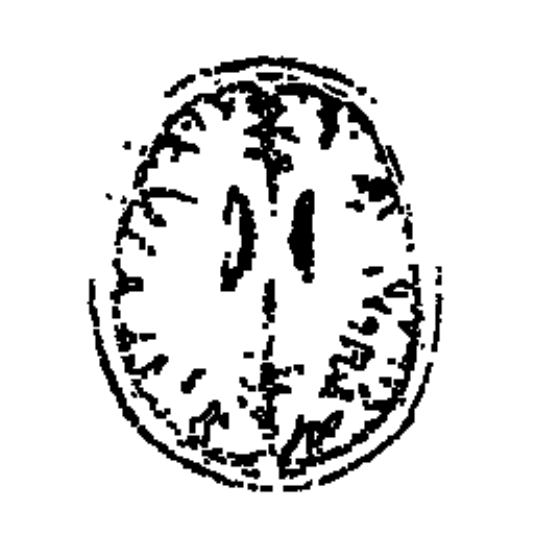}
		&\includegraphics[width=0.074\textwidth]{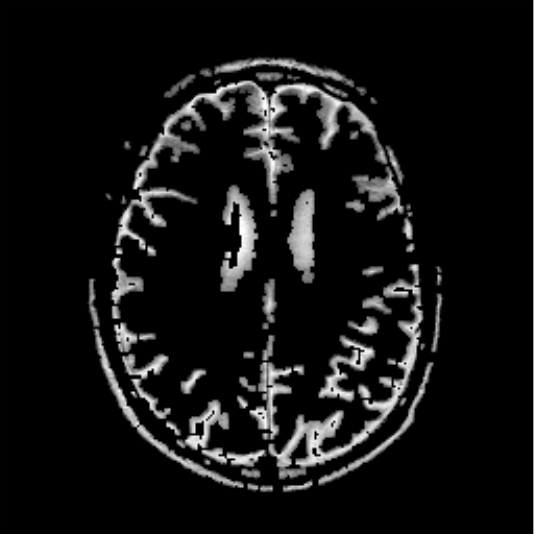}
		&\includegraphics[width=0.074\textwidth]{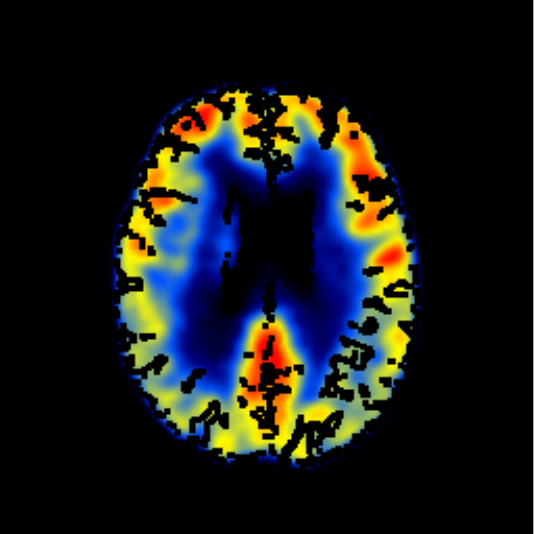}
		\\
		\includegraphics[width=0.074\textwidth]{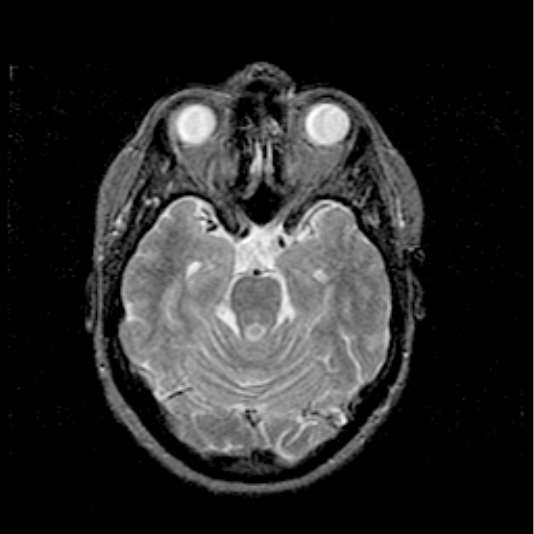}
		&\includegraphics[width=0.074\textwidth]{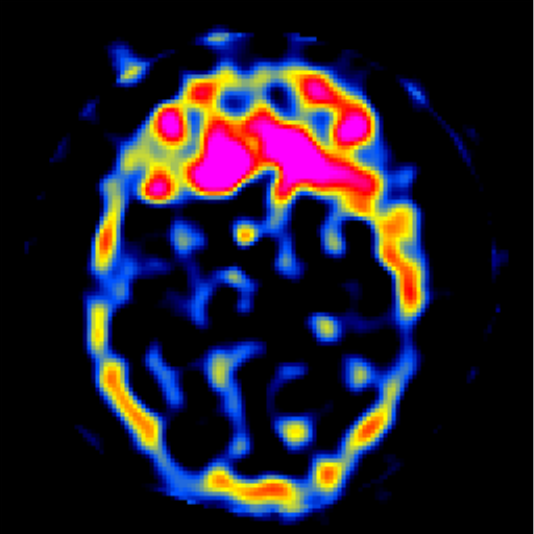}
		&\includegraphics[width=0.074\textwidth]{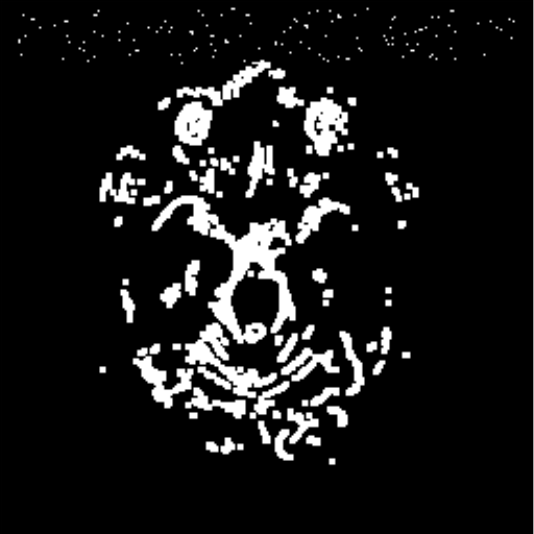}
		&\includegraphics[width=0.074\textwidth]{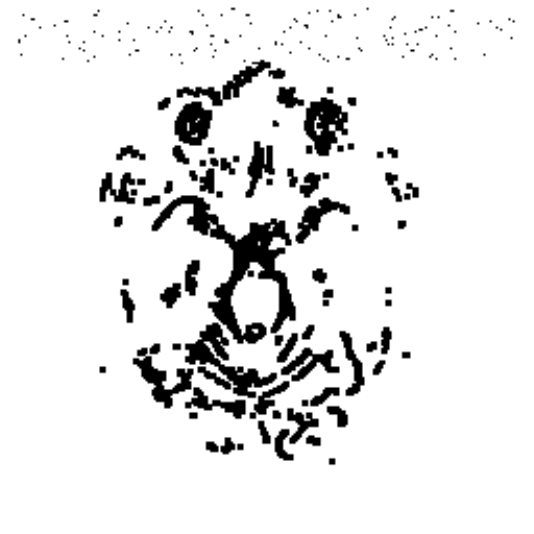}
		&\includegraphics[width=0.074\textwidth]{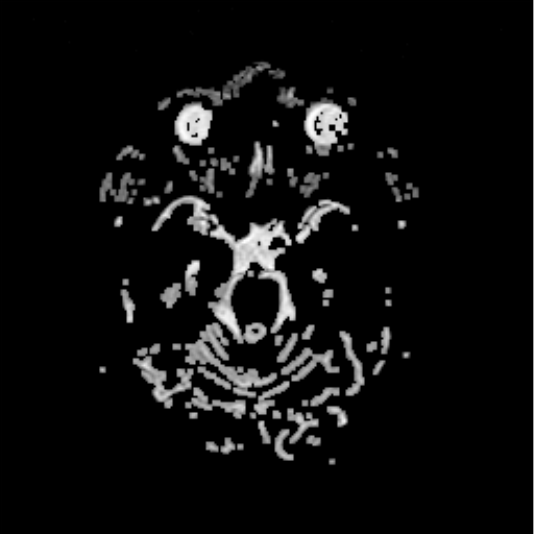}
		&\includegraphics[width=0.074\textwidth]{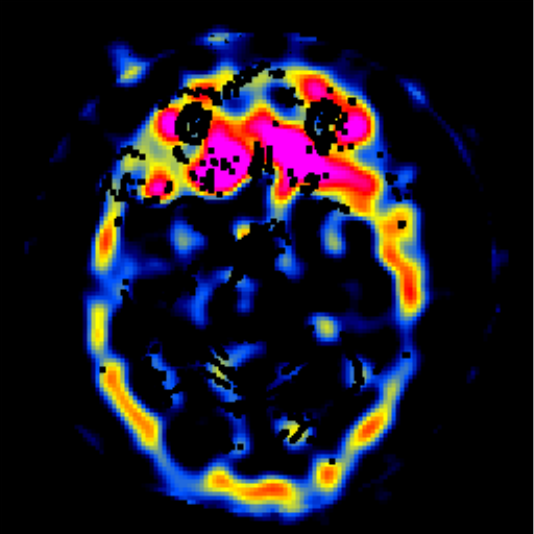}
		\\
		\tiny $\rm{I}_{MRI}$&\tiny$\rm{I}_{Fun}$&\tiny$\mathcal{M}_{m}$&\tiny$\mathcal{M}_{f}$&\tiny$\rm{I}_{MRI}\odot\mathcal{M}_{m}$&\tiny$\rm{I}_{Fun}\odot\mathcal{M}_{f}$
	\end{tabular}
	\caption{Typical examples of masks for medical image fusion task.}
	\label{fig:M2}
\end{figure}

\begin{equation}
	\mathcal{L}_{fun} = \sum_{i=1}^{N}w_{i}\frac{\parallel{\nu_i}-{\nu_i}^{+}\parallel_{1}}{\sum_{m}^{M}\parallel{\nu_i}-{\nu_i}^{m-}\parallel_{1}},
	\label{eq5}
\end{equation} 
where $\nu_i$ denotes the functional features of the fused image, which is defined as $G_i({\rm I}_{F}\odot\mathcal{M}_{f})$. $\nu_i^{+}$ and $\nu_i^{m-}$ are the positive and negative samples, formulated as $\nu_i^{+}=G_i({\rm I}_{Fun}\odot\mathcal{M}_{f})$, $\nu_i^{m-}=G_i({\rm I}^{m}_{MRI}\odot\mathcal{M}_{f})$, respectively. Superscript $m$ means the $m$th negative sample.

\section{Experiments}
\subsection{Experimental Settings}
\subsubsection{Datasets}
The infrared and visible image pairs we utilize to evaluate our method are collected from the TNO~\footnote{https://figshare.com/articles/TNO Image Fusion Dataset/1008029} and the RoadScene~\cite{U2Fusion2020}, which are publicly available. The above mentioned datasets are clarified below.

\noindent\textbf{TNO dataset }The TNO is a widely used dataset for infrared and visible image fusion. We adopt TNO as the benchmark to train our network for its high quality images with distinctive scenarios.

\noindent\textbf{RoadScene dataset }The RoadScene includes realistic driving scenes (\emph{e.g.}, vehicles, pedestrians and road symbol signs). It contains 221 representative image pairs with no uniform resolution, collected from authentic driving videos.

\begin{figure*}
	\centering
	\setlength{\tabcolsep}{1pt} 
	
	\includegraphics[width=0.98\textwidth]{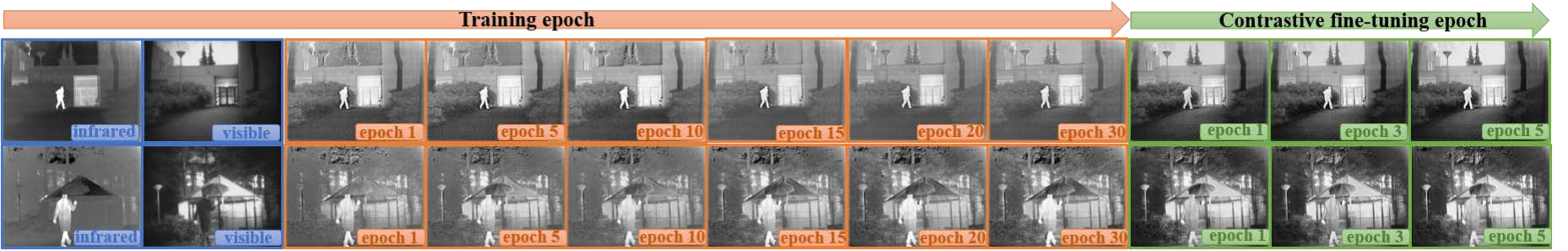}
	
	\caption{Visual illustration of training epochs incrementally on the TNO dataset. It is important to note that during the fine-tuning phase, the contrastive constraints exhibit a robust capability in recovering distinct foreground thermal targets and maintaining abundant background details.}
	\label{fig:training } 
\end{figure*}

\subsubsection{Training details}
Our entire fusion framework is trained on the TNO dataset through two phases: training and fine-tuning. The overall training strategy can be found in Algorithm~\ref{alg:correction1}.

In the training stage, only the self-adaptive loss is applied to update the network parameters (i.e., the contrastive constraints are not involved in this phase). Concretely, the self-adaptive factors $\sigma^{a,b}$ and $\gamma^{a,b}$ are first computed by measuring the average gradient and entropy of an image pair,  note that this will not affect network parameters. Thereafter, the network is penalized by the self-adaptive loss. As for data pre-processing and other hyper-parameters, we  select 46 pairs of images and convert them to greyscale ones. To make full use of the gradient and entropy of each image for self-adaptive training loss, 1410 image patches of size 64$\times$64 were cropped from source images, this enables the network to better perceive subtle gradients and entropy. Then, the training patches are normalized to [-1, 1] and fed into our network. The Adam was selected as optimizer and the learning rate was set to 0.0001 with the batch size of 30.

In the fine-tuning phase, both self-adaptive loss and the contrastive constraints participate in updating the network weights, as shown in Algorithm~\ref{alg:correction1}. Data used in this step include merely 18 images with salient masks from TNO. As in the previous stage, 1410 images of size 64$\times$64 are cropped. For the contrastive constraint loss, we use one positive sample with three negative samples (one corresponded with the positive patch and two randomly selected from other negative patches). The network is updated for 5 epochs, the optimizer, learning rate and batch size settings are the same as that in the first phase. The tuning parameter $\alpha$ is set to an empirical value of 20. The visual illustration of fusion performance over training epoch is given in Figure~\ref{fig:training }.

Likewise, to train our model for medical imaging, two stages are required: training and fine-tuning. 2662 PET patches and 4114 SPECT patch images are selected from Atlas\footnote{http://www.med.harvard.edu/aanlib} dataset. All images are cropped to size 64$\times$64 and normalized to [-1, 1] as training set. We select Adam as the optimizer and learning rate is set to 0.0001 in both stages. In the training stage, the model is trained for 3 epochs with batch size of 30. In the fine tuning stage, the model is trained for 1 epoch with batch size of 10. The self-adaptive loss and contrastive loss settings are the same as the IVIF task. The tuning parameter $\alpha$ is set to 20.

\begin{algorithm}[t]
	\caption{Two-stage Training}\label{alg:correction1}
	\begin{algorithmic}[1]
		\Require Source images \{$\mathbf{x}$,$\mathbf{y}$\}, contrastive loss $\mathcal{L}_{\{ir,vis\}}$, self-adaptive loss $\mathcal{L_P}$ and other necessary hyper-parameters.
		\Ensure Optimal parameter $\bm{\omega}^{*}$. 
		\While{not converge}
		\State \% Train without contrastive loss 
		\For{$\mathbf{N}$ iterations}
		\State \% Determine self-adaptive factors
		\State $\bm{\sigma}^{a,b}, \bm{\gamma}^{a,b}\leftarrow\mathcal{G}(\mathbf{x},\mathbf{y}),\mathcal{E}(\mathbf{x},\mathbf{y})$
		\State \% Compute self-adaptive loss
		\State $\mathcal{L_P}\leftarrow\ \mathcal{L_S}(x,y,\mathcal{N}(\mathbf{x},\mathbf{y};{\bm{\omega}}); \sigma^{a},\sigma^{b})+\mathcal{L_N}(x,y,\mathcal{N}(\mathbf{x},\mathbf{y};{\bm{\omega}}); \gamma^{a},\gamma^{b})$
		\State \% Update trainable parameters
		\State ${\bm{\omega}}\leftarrow{\bm{\omega}}-\nabla_{{\bm{\omega}}}{\mathcal{L_P}}$
		\EndFor
		\State \% Finetune by adding constrastive loss
		\For{$\mathbf{M}$ iterations}
		\State $\bm{\sigma}^{a,b}, \bm{\gamma}^{a,b}\leftarrow\mathcal{G}(\mathbf{x},\mathbf{y}),\mathcal{E}(\mathbf{x},\mathbf{y})$
		\State $\mathcal{L_P}\leftarrow\ \mathcal{L_S}(x,y,\mathcal{N}(\mathbf{x},\mathbf{y};{\bm{\omega}}); \sigma^{a},\sigma^{b})+\mathcal{L_N}(x,y,\mathcal{N}(\mathbf{x},\mathbf{y};{\bm{\omega}}); \gamma^{a},\gamma^{b})$
		\State ${\bm{\omega}}\leftarrow{\bm{\omega}}-\nabla_{{\bm{\omega}}}({\mathcal{L_P}}+\mathcal{L}_{ir}+\mathcal{L}_{vis})$
		\EndFor
		\EndWhile

	\end{algorithmic}
\end{algorithm}

\subsubsection{Evaluation metrics}

To quantitatively evaluate the fusion performance, in this letter, we select six commonly used metrics for image quality measurement, including EN~\cite{Roberts2008Assessment}, AG~\cite{eskicioglu1995image}, SF~\cite{eskicioglu1995image}, SD~\cite{aslantas2015new}, SCD~\cite{ma2019infrared} and VIF~\cite{han2013new}. Their details are given as follows.

\noindent\textbf{Entropy (EN)} EN measures the abundance of information included in an image, a larger value indicates that the fusion strategy performs better. The formulation is given as Eq.~\ref{EN}.

\noindent\textbf{Average gradient (AG)}
AG measures the gradient information of the fused image, which can reflect the details of the fused image. The formulation is given as Eq.~\ref{AG}.

\noindent\textbf{Spatial frequency (SF)} SF is a criterion that reflects the change of gray level in an image. An image with a higher SF value possesses more textural details. It is obtained based on horizontal and vertical gradient information. The mathematical expression can be described as follows:

\begin{equation}
	\rm SF = \sqrt{H^{2}+V^{2}}
	\label{eq4}
\end{equation} 

where $H$ and $V$ are:

\begin{equation}
	H = \sqrt{\frac{1}{MN}\sum_{i=1}^{M}\sum_{j=2}^{N}{\lvert I_F(i,j)-I_F(i,j-1)\rvert}^{2}}
\end{equation} 

\begin{equation}
	V = \sqrt{\frac{1}{MN}\sum_{i=2}^{M}\sum_{j=1}^{N}{\lvert I_{F}(i,j)-I_{F}(i-1,j) \rvert}^{2}}
	\label{eq6}
\end{equation} 

where $M$ and $N$ are the width and height of the estimated image.

\noindent\textbf{Standard deviation (SD)} SD is utilized to measure whether an image has abundant information and high contrast. A larger SD value suggests more characteristics included in an image. It is given as follows:

\begin{equation}
	SD = \frac{1}{MN}\sum_{i=1}^{M}\sum_{j=1}^{N}{\lvert I_{F}(i,j)-\mu \rvert}^{2}
	\label{eq7}
\end{equation} 

where $\mu$ is the average pixel value, $M$ and $N$ are the width and height of the estimated image.

\noindent\textbf {The sum of the correlations of differences(SCD)}
SCD is an index based on image correlation. First, the definition of correlation about source image $I_{X}$ and fused image $I_{F}$ is given as follows:

\begin{tiny}
	\begin{equation}
		r(I_{X}, I_{F})=\frac{\sum_{i=1}^{M} \sum_{j=1}^{N}(I_{X}(i, j)-\overline{I_{X}})(I_{F}(i, j)-\overline{I_{F}})}{\sqrt{\sum_{i=1}^{M} \sum_{j=1}^{N}(I_{X}(i, j)-\overline{I_{X}})^{2}\left(\sum_{i=1}^{M} \sum_{j=1}^{N}(I_{F}(i, j)-\overline{I_{F}})^{2}\right)}}
	\end{equation}
\end{tiny}

\begin{figure*}
	\centering
	
	\setlength{\tabcolsep}{1pt}
	\begin{tabular}{cccccc}
		
		\includegraphics[width=0.13\textwidth,height=0.07\textheight]{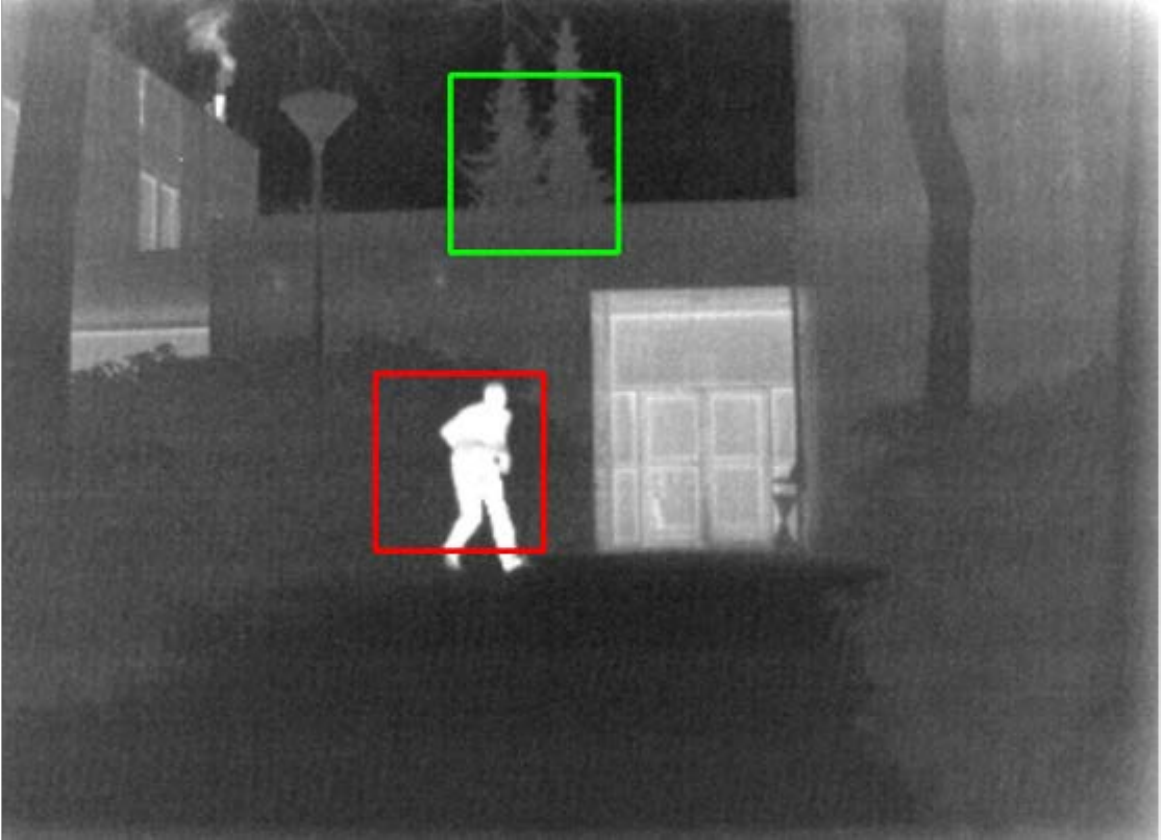}
		&\includegraphics[width=0.17\textwidth,height=0.07\textheight]{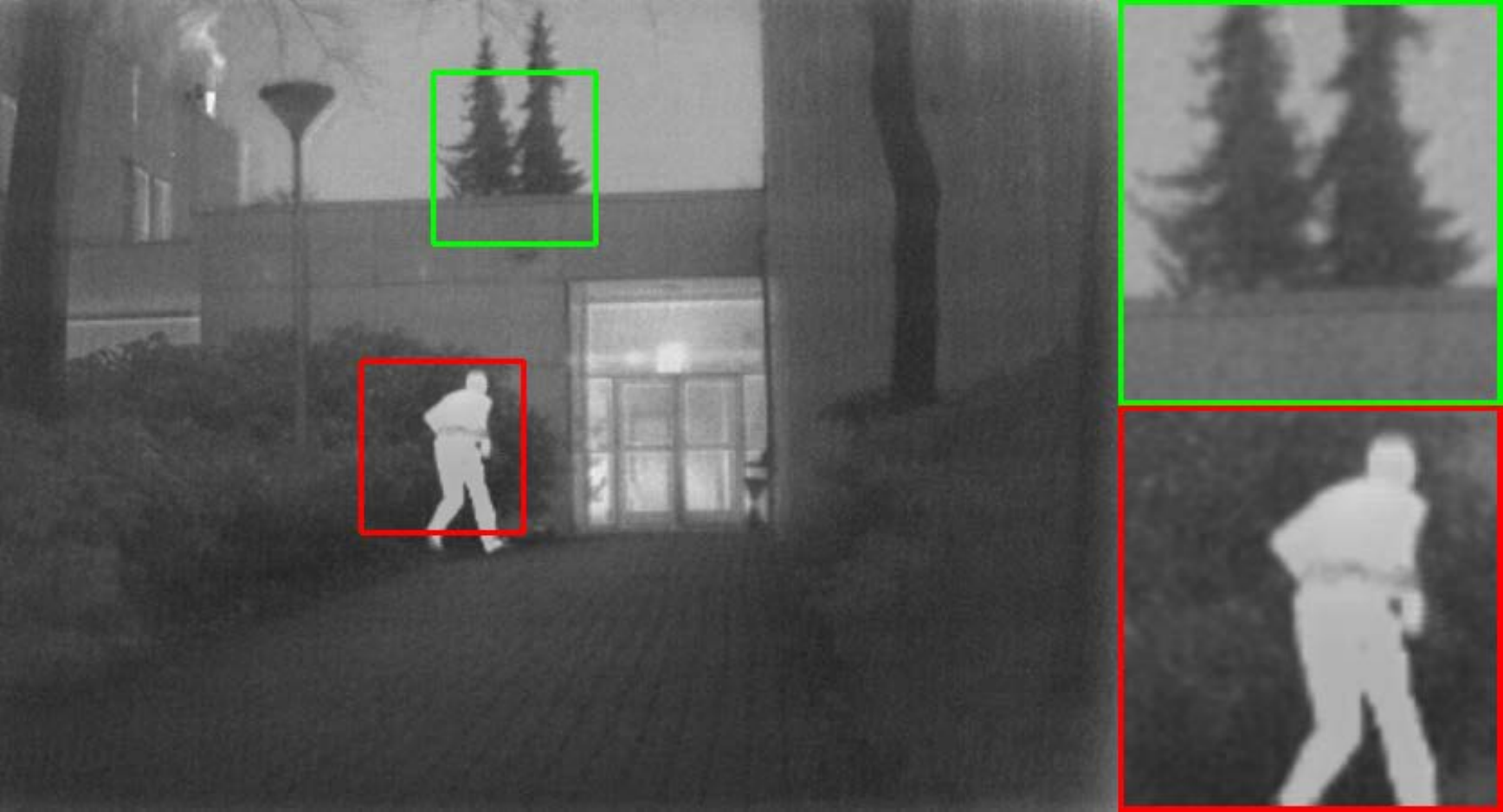}
		&\includegraphics[width=0.17\textwidth,height=0.07\textheight]{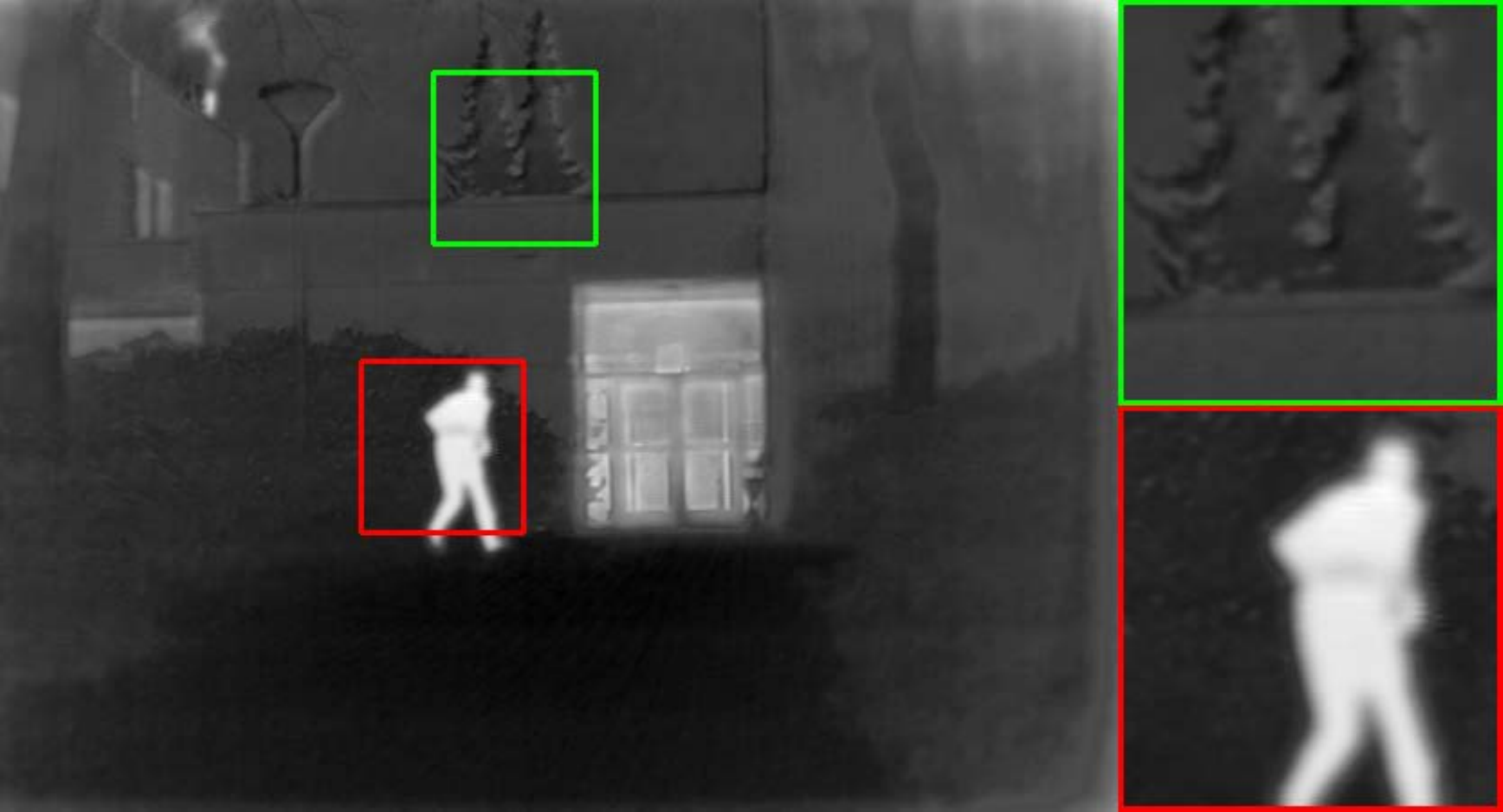}
		&\includegraphics[width=0.17\textwidth,height=0.07\textheight]{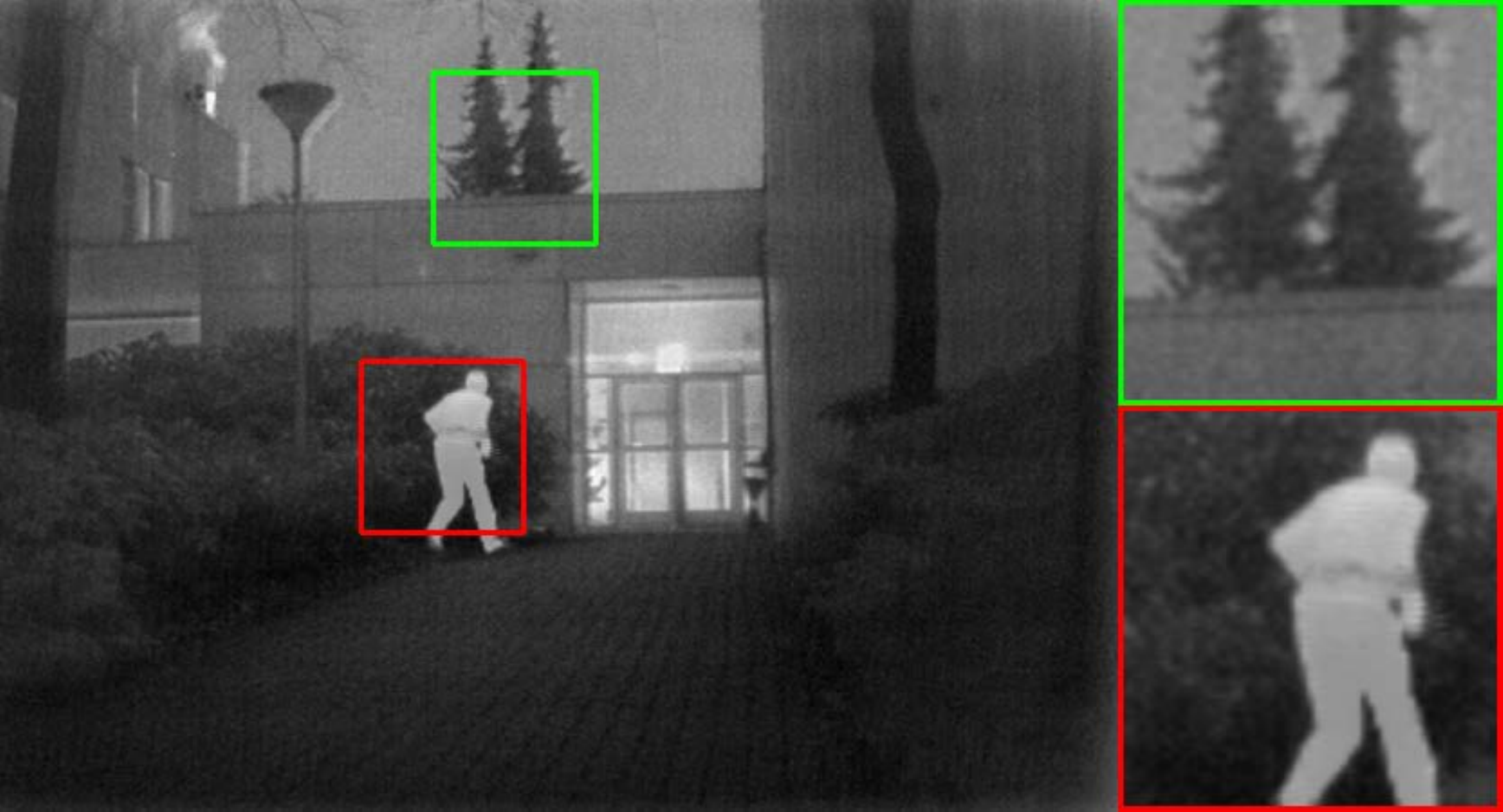}
		&\includegraphics[width=0.17\textwidth,height=0.07\textheight]{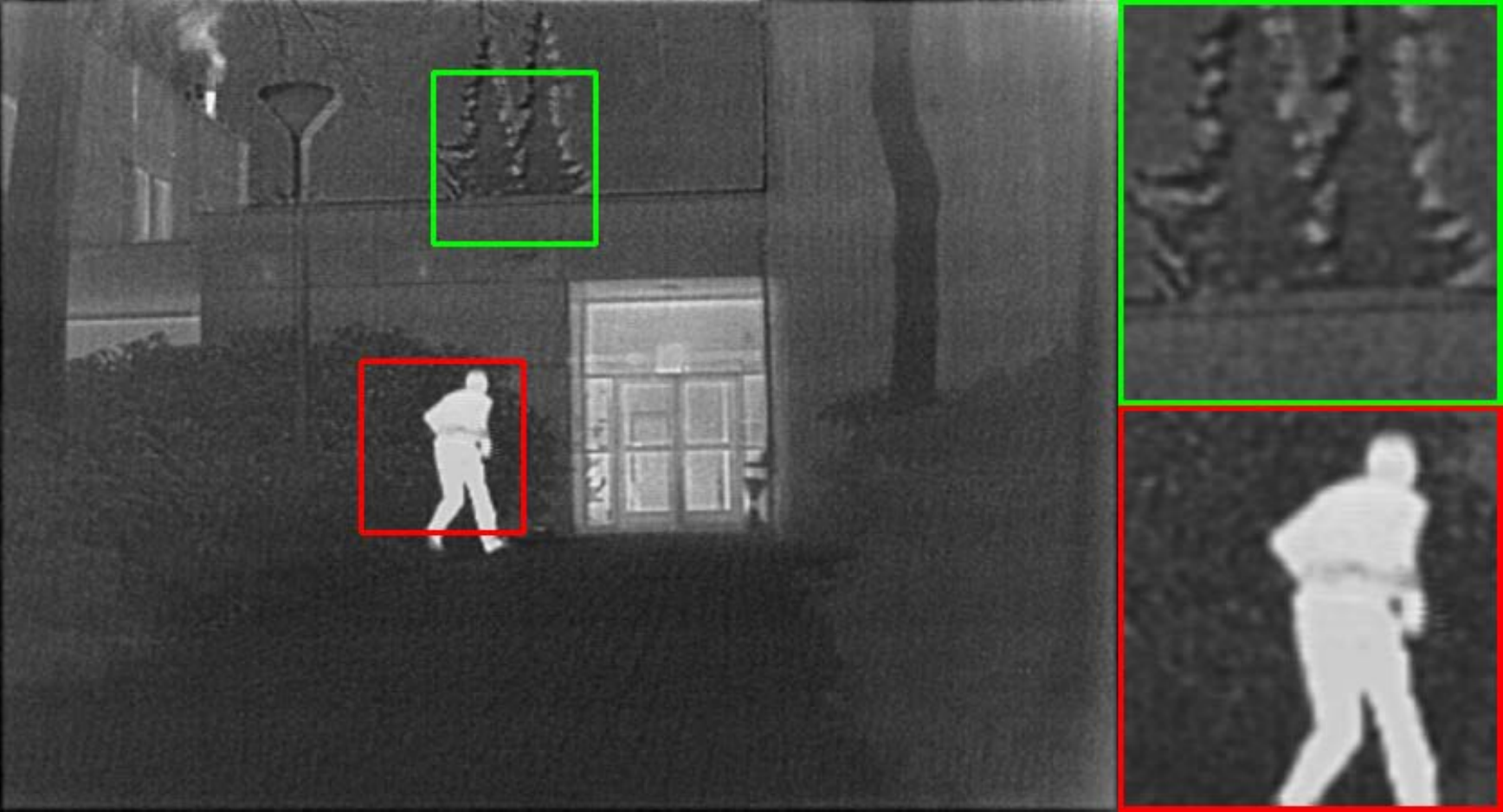}
		&\includegraphics[width=0.17\textwidth,height=0.07\textheight]{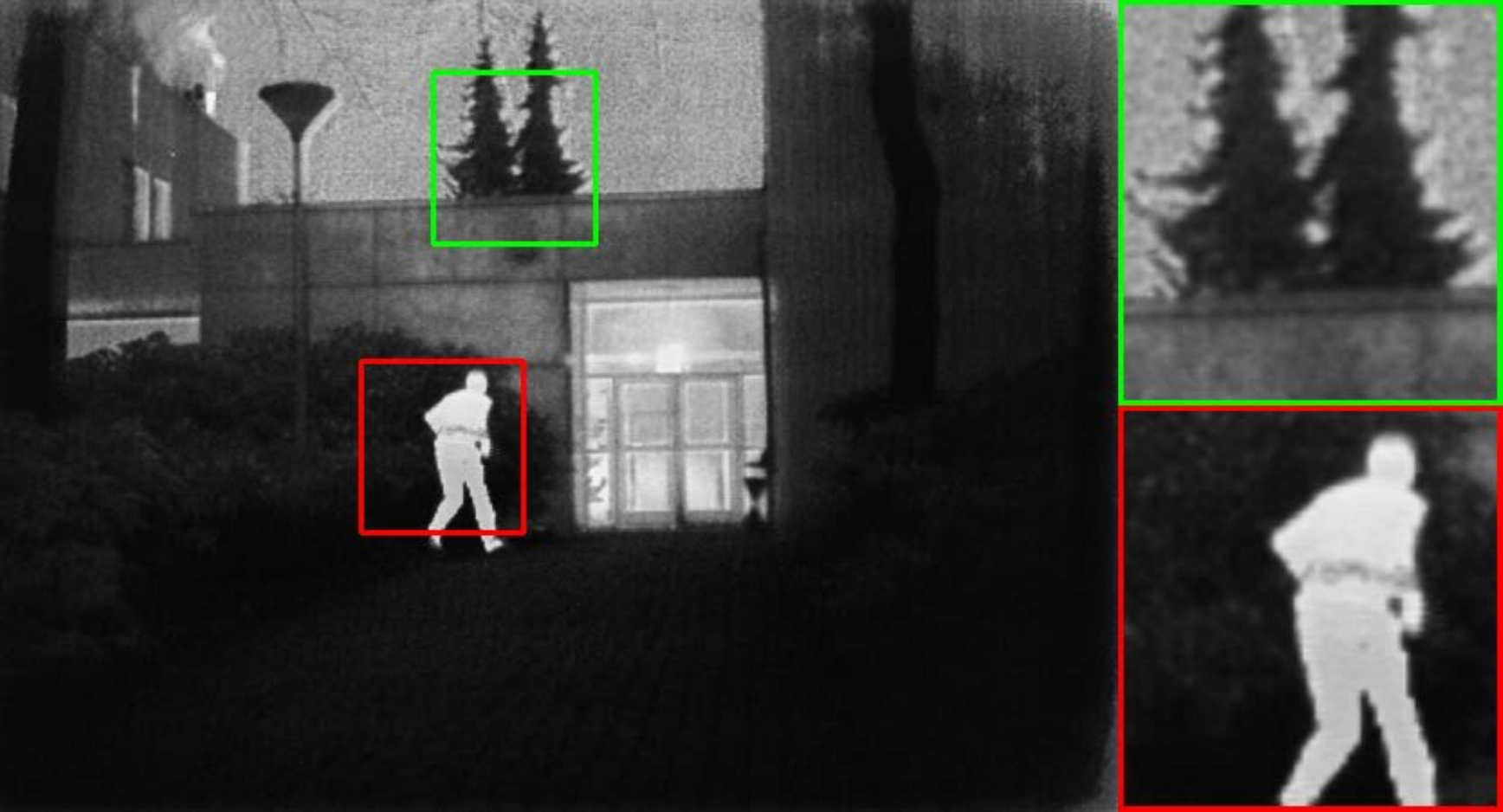}
		\\
		Infrared &SMoA&FusionGAN&DenseFuse&SDNet&DIDFuse
		\\
		\includegraphics[width=0.13\textwidth,height=0.07\textheight]{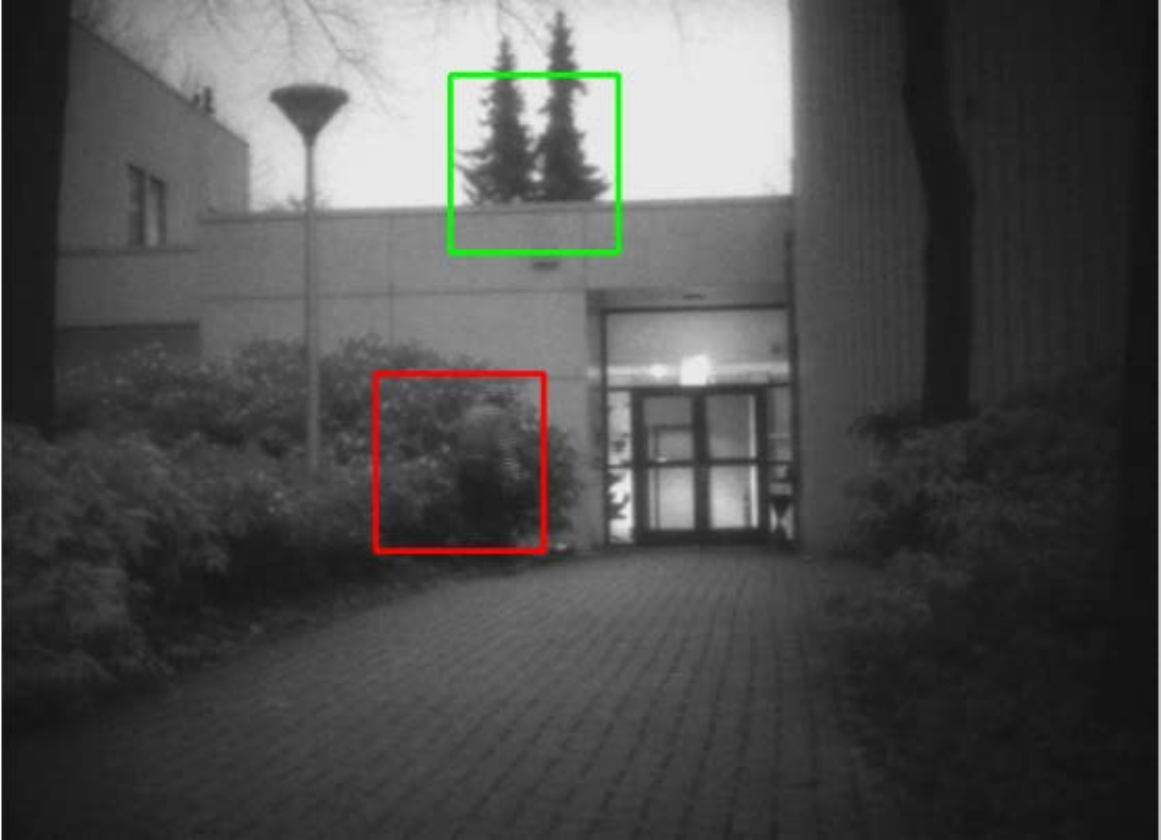}
		&\includegraphics[width=0.17\textwidth,height=0.07\textheight]{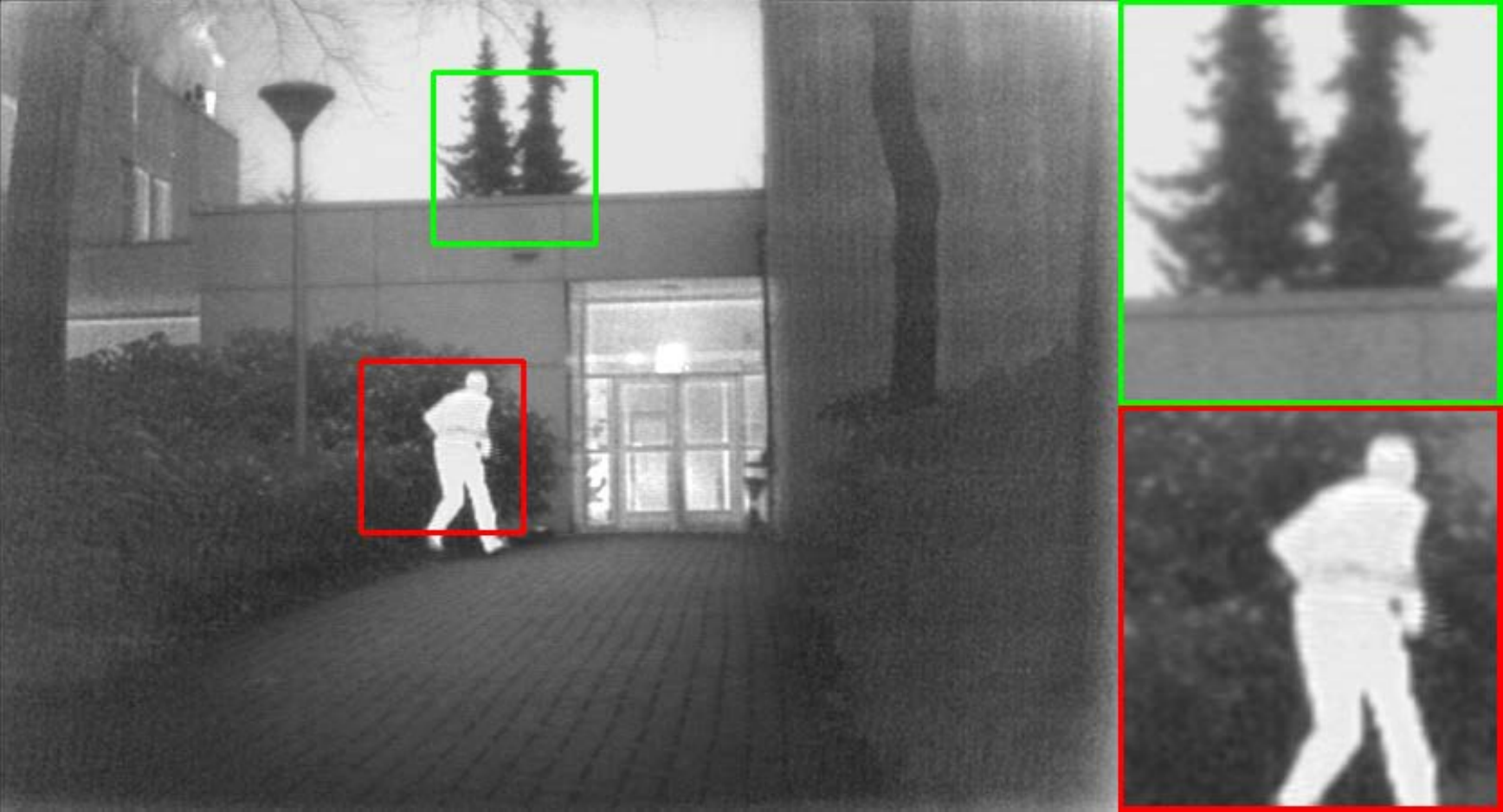}
		&\includegraphics[width=0.17\textwidth,height=0.07\textheight]{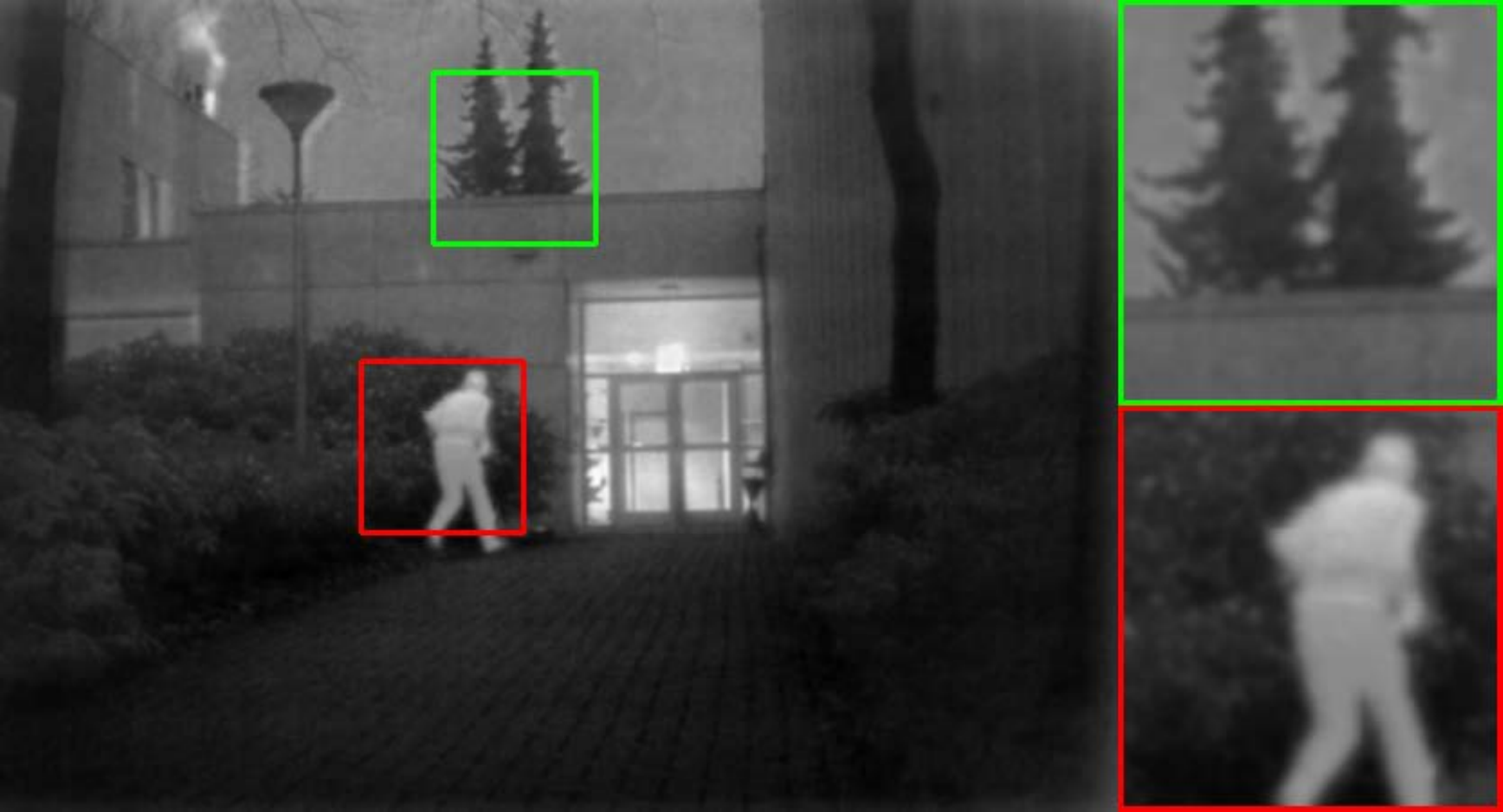}
		&\includegraphics[width=0.17\textwidth,height=0.07\textheight]{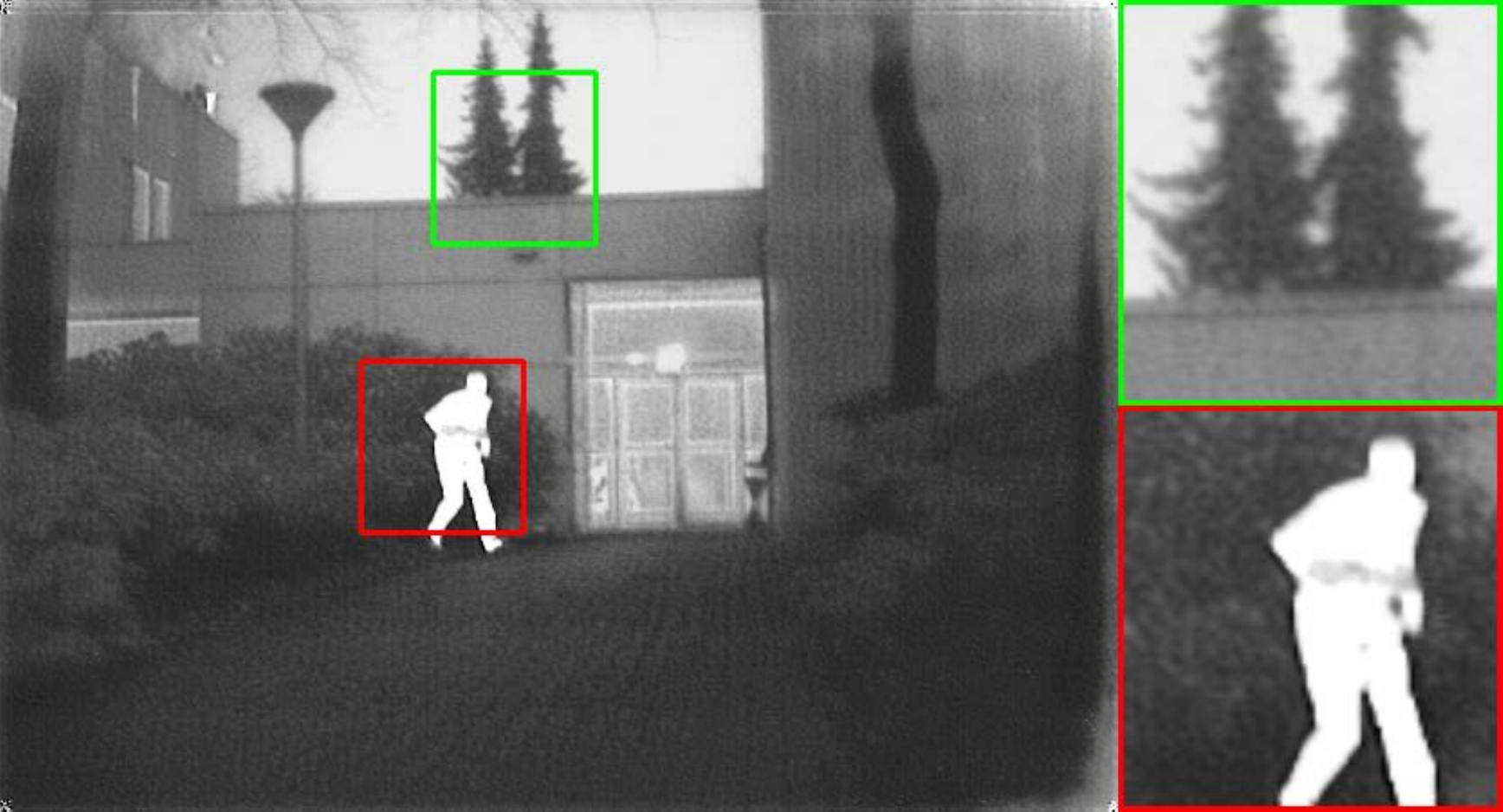}
		&\includegraphics[width=0.17\textwidth,height=0.07\textheight]{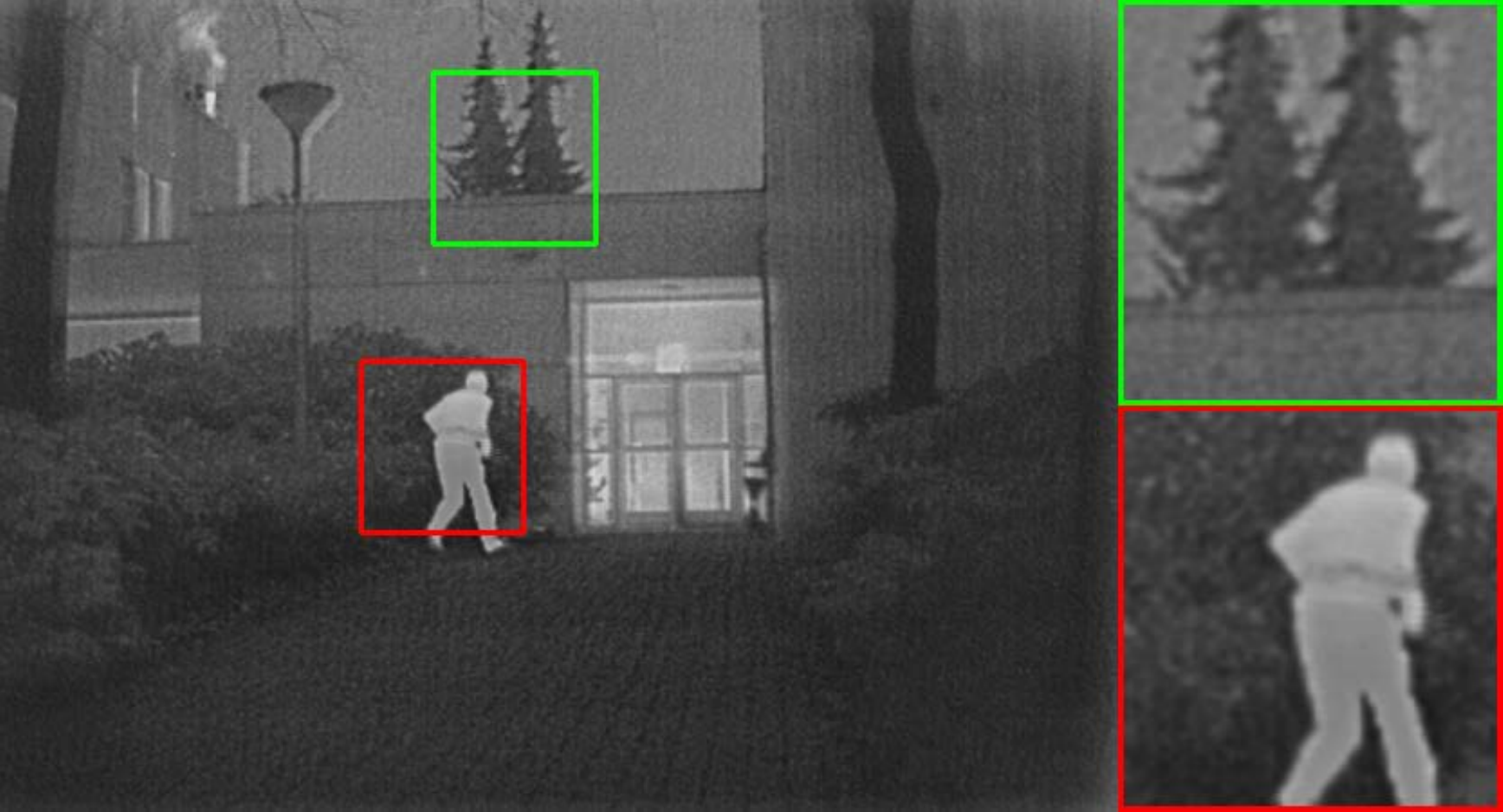}
		&\includegraphics[width=0.17\textwidth,height=0.07\textheight]{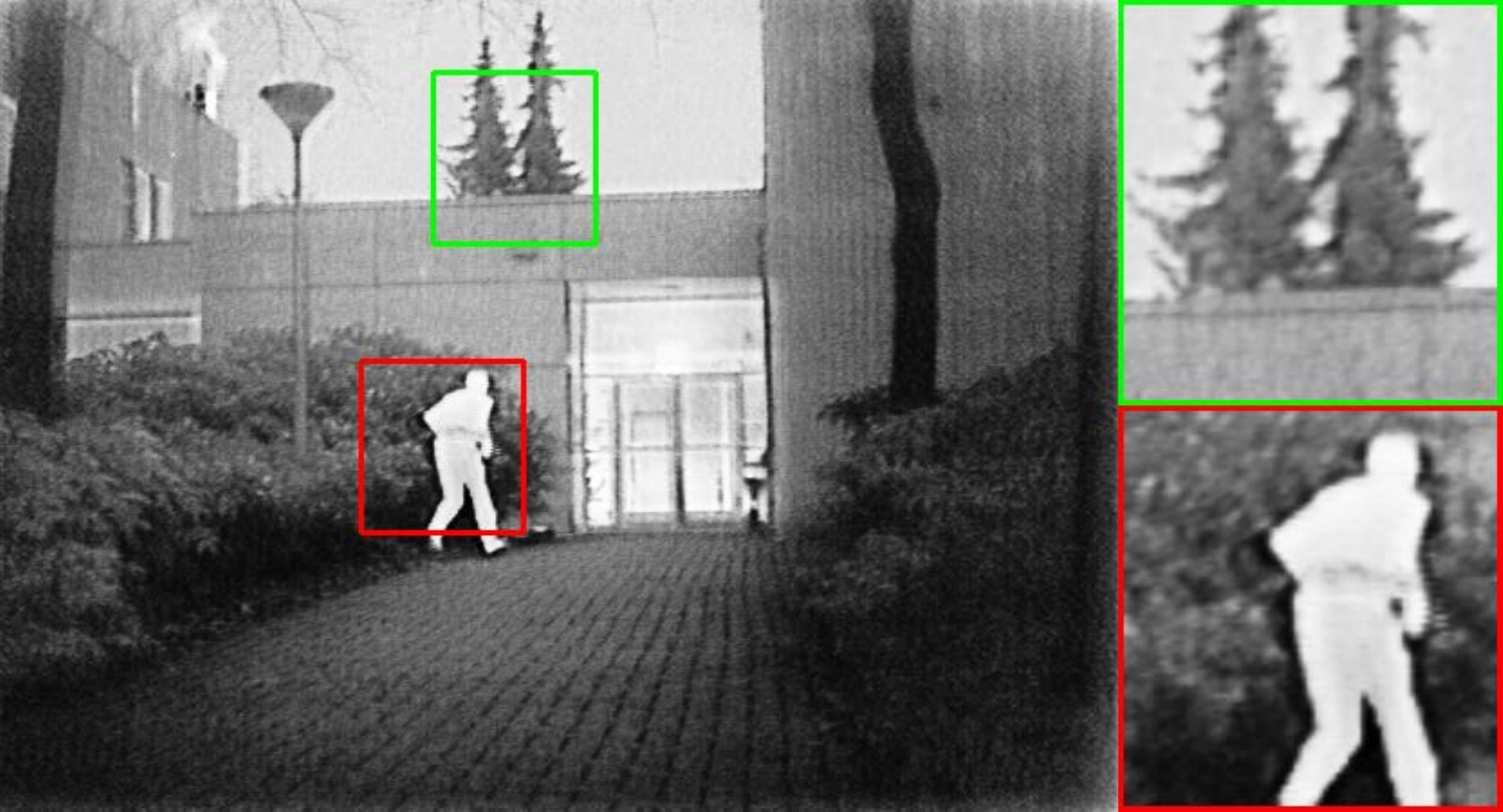}
		\\
		Visible&SwinFusion&RFN&TarDAL&U2Fusion&Ours
		\\
		\includegraphics[width=0.13\textwidth,height=0.07\textheight]{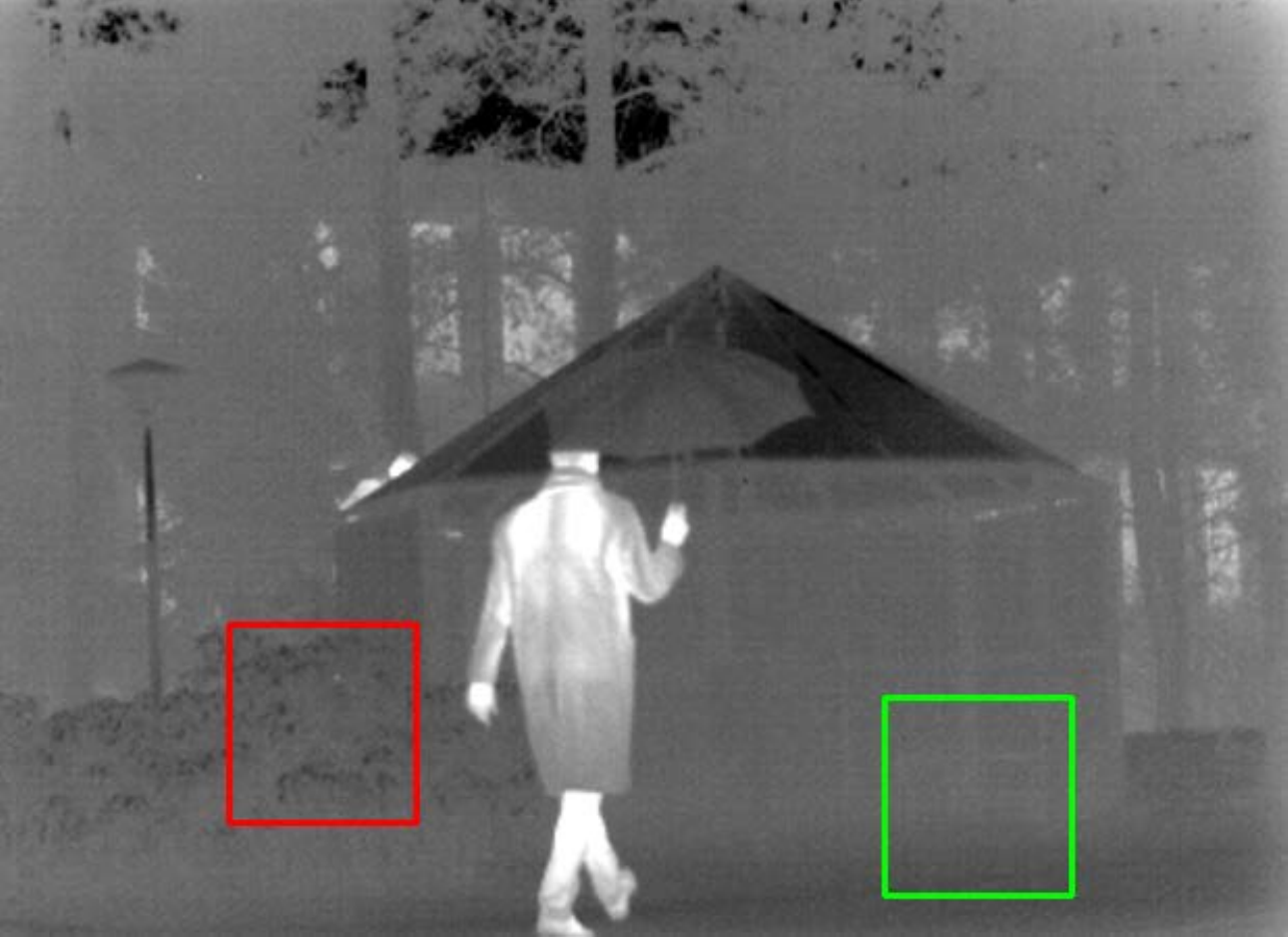}
		&\includegraphics[width=0.17\textwidth,height=0.07\textheight]{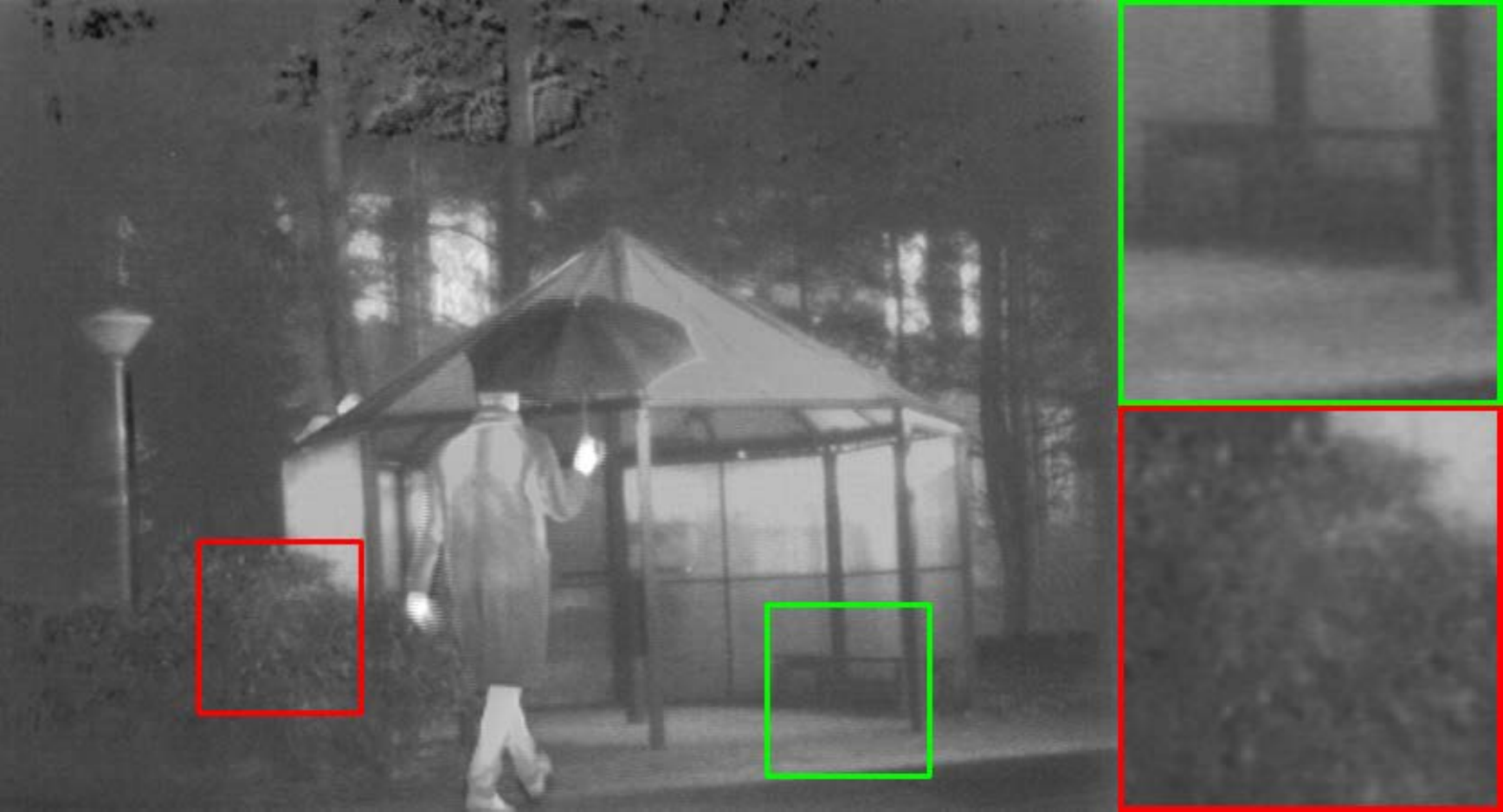}
		&\includegraphics[width=0.17\textwidth,height=0.07\textheight]{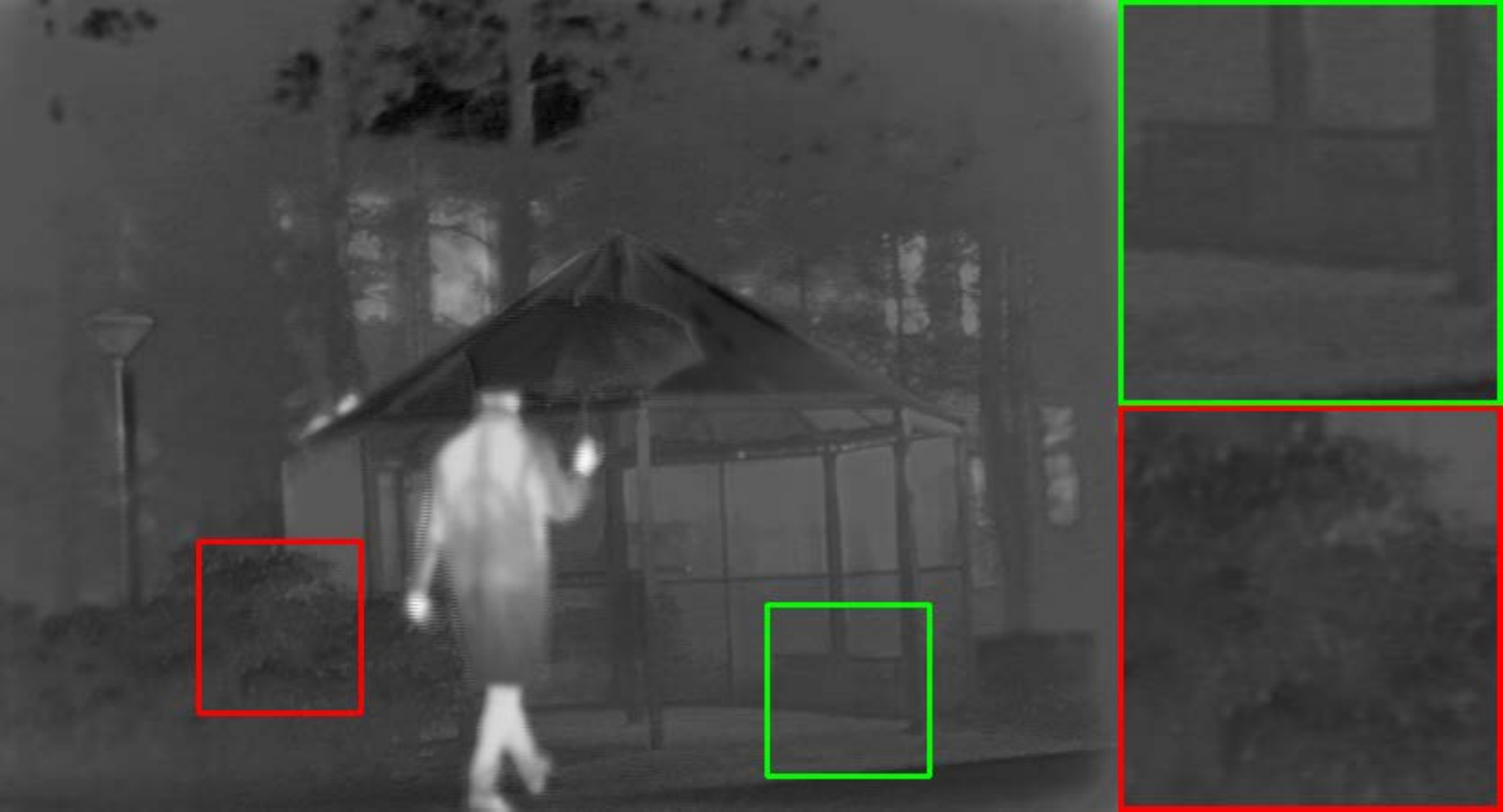}
		&\includegraphics[width=0.17\textwidth,height=0.07\textheight]{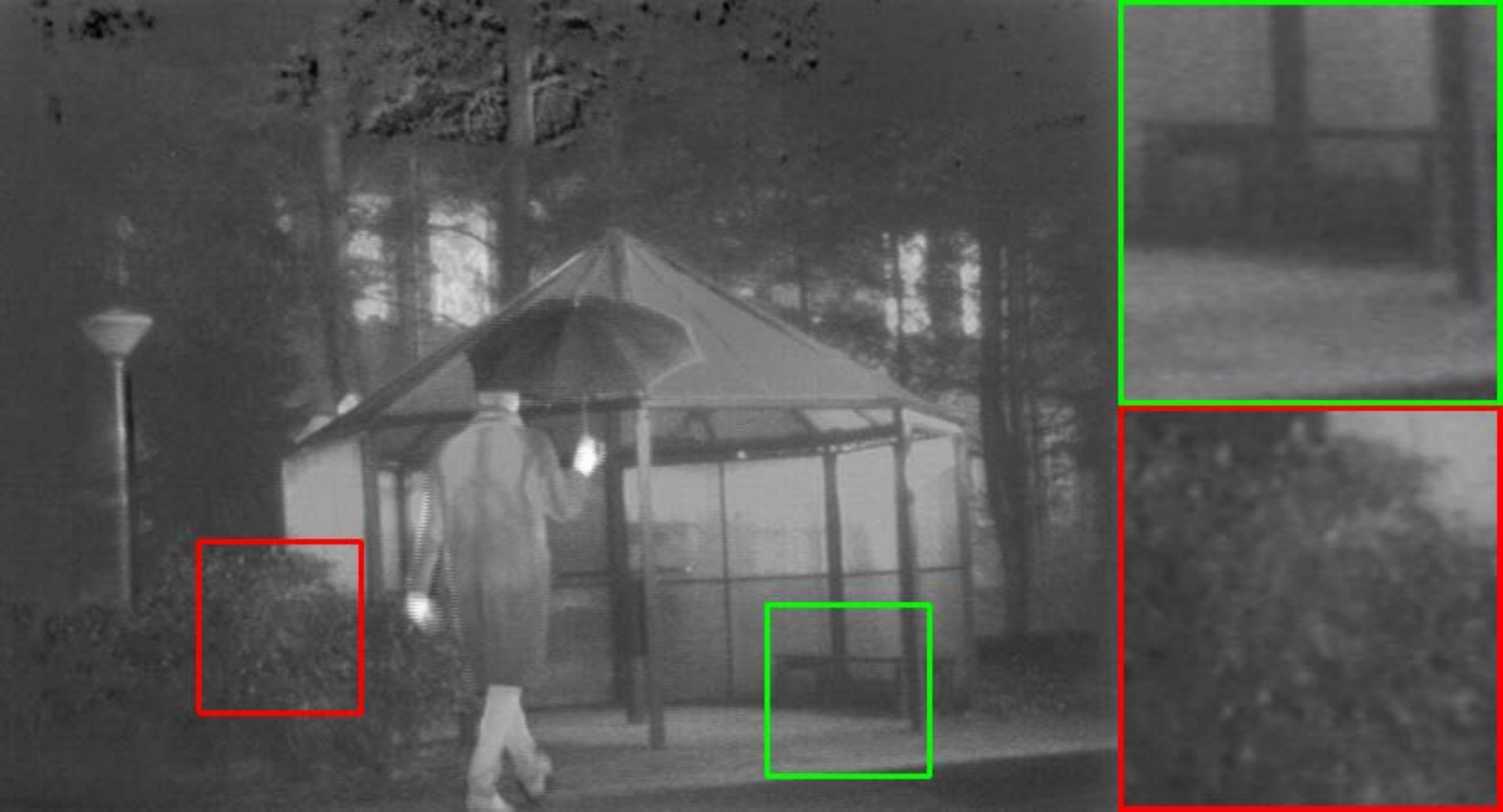}
		&\includegraphics[width=0.17\textwidth,height=0.07\textheight]{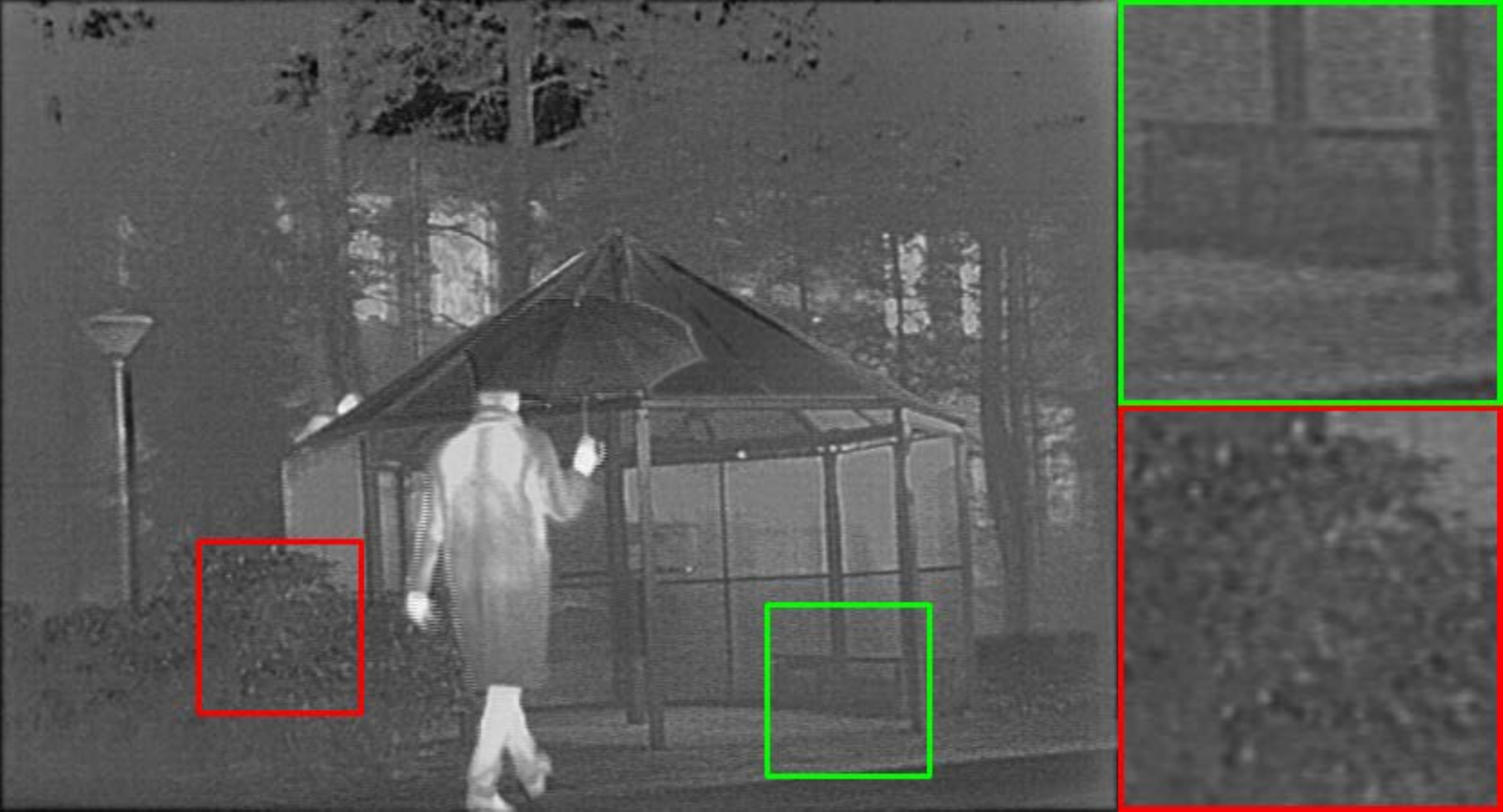}
		&\includegraphics[width=0.17\textwidth,height=0.07\textheight]{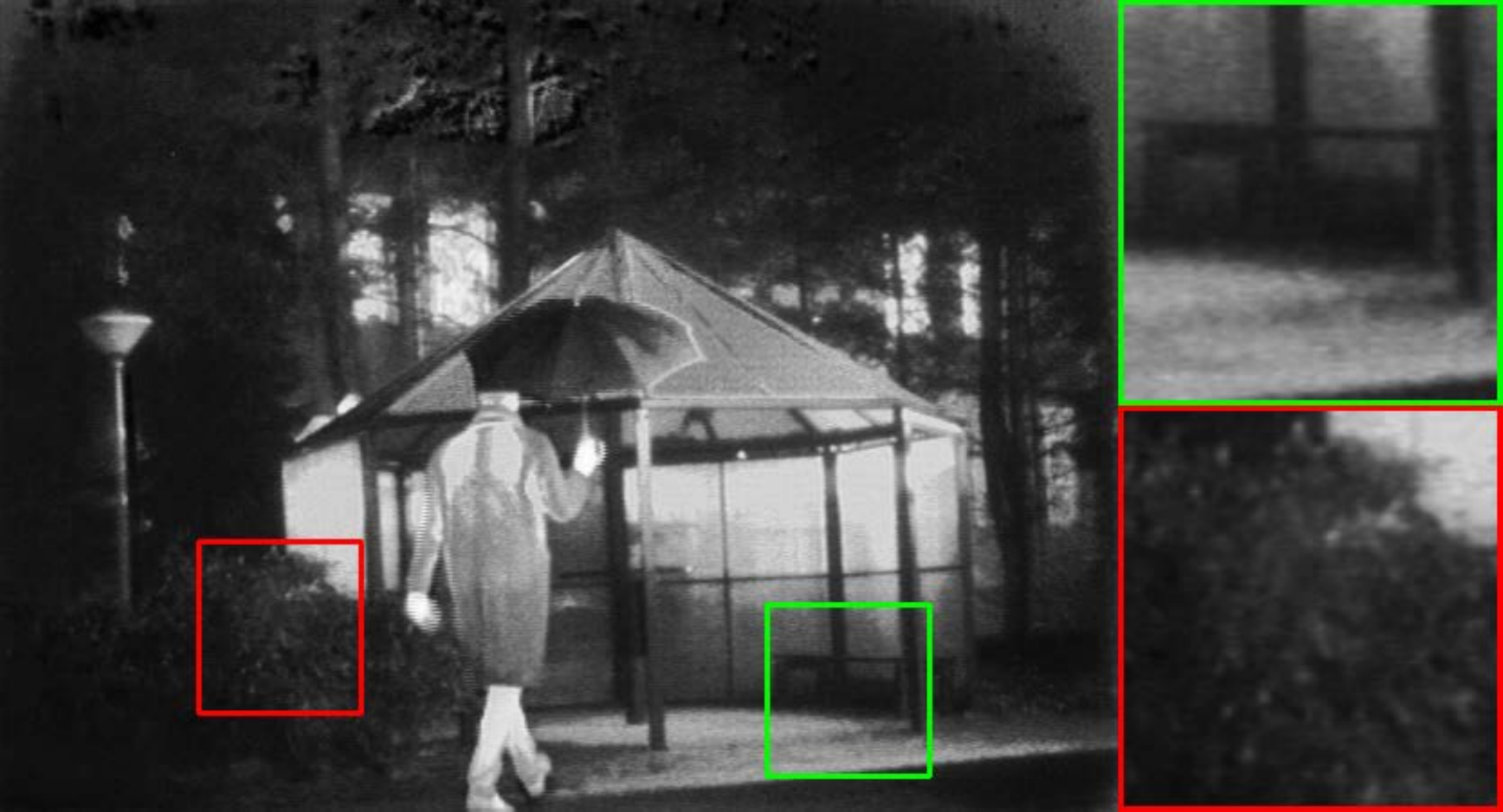}
		\\
		Infrared&SMoA&FusionGAN&DenseFuse&SDNet&DIDFuse
		\\
		\includegraphics[width=0.13\textwidth,height=0.07\textheight]{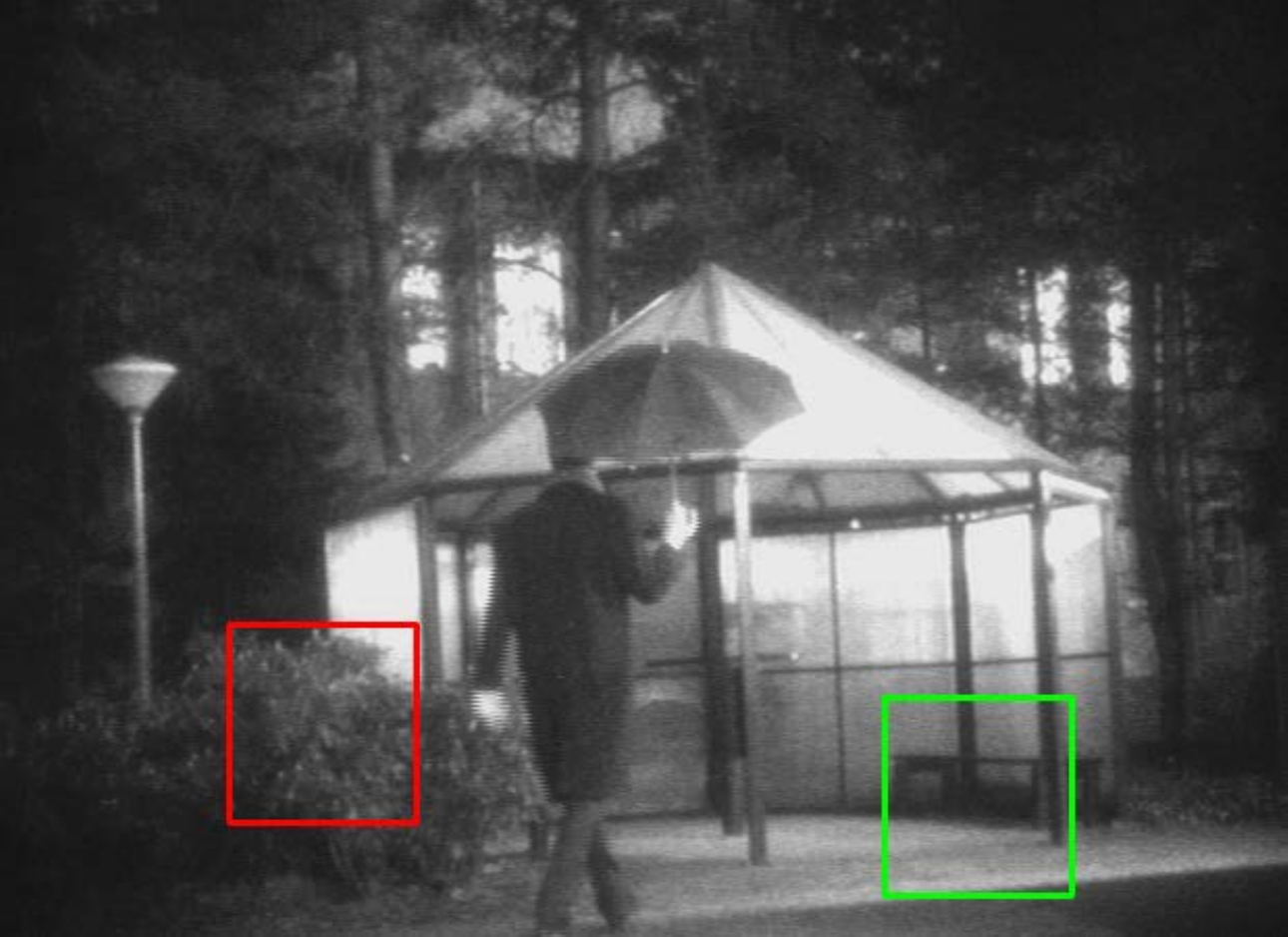}
		&\includegraphics[width=0.17\textwidth,height=0.07\textheight]{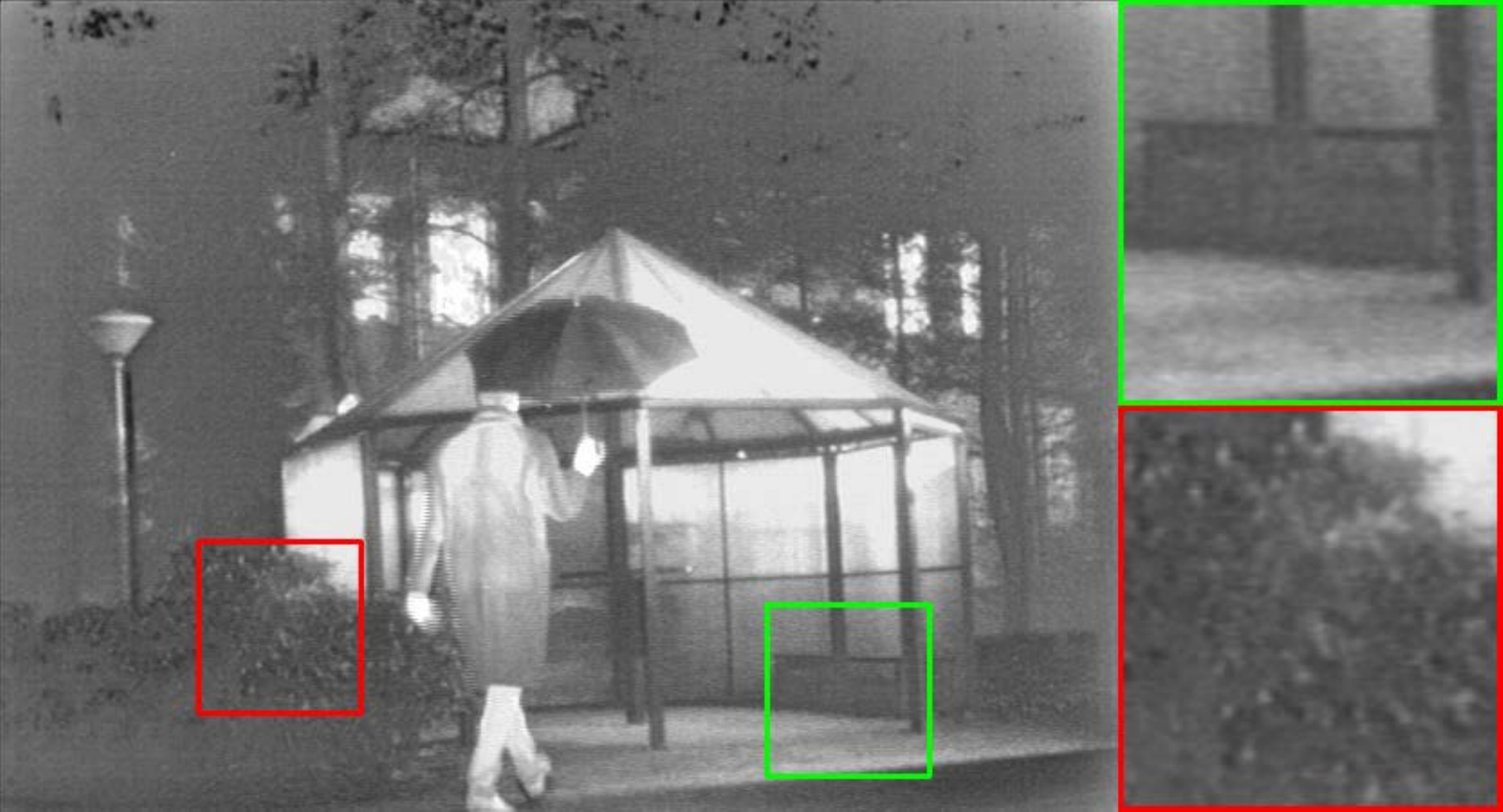}
		&\includegraphics[width=0.17\textwidth,height=0.07\textheight]{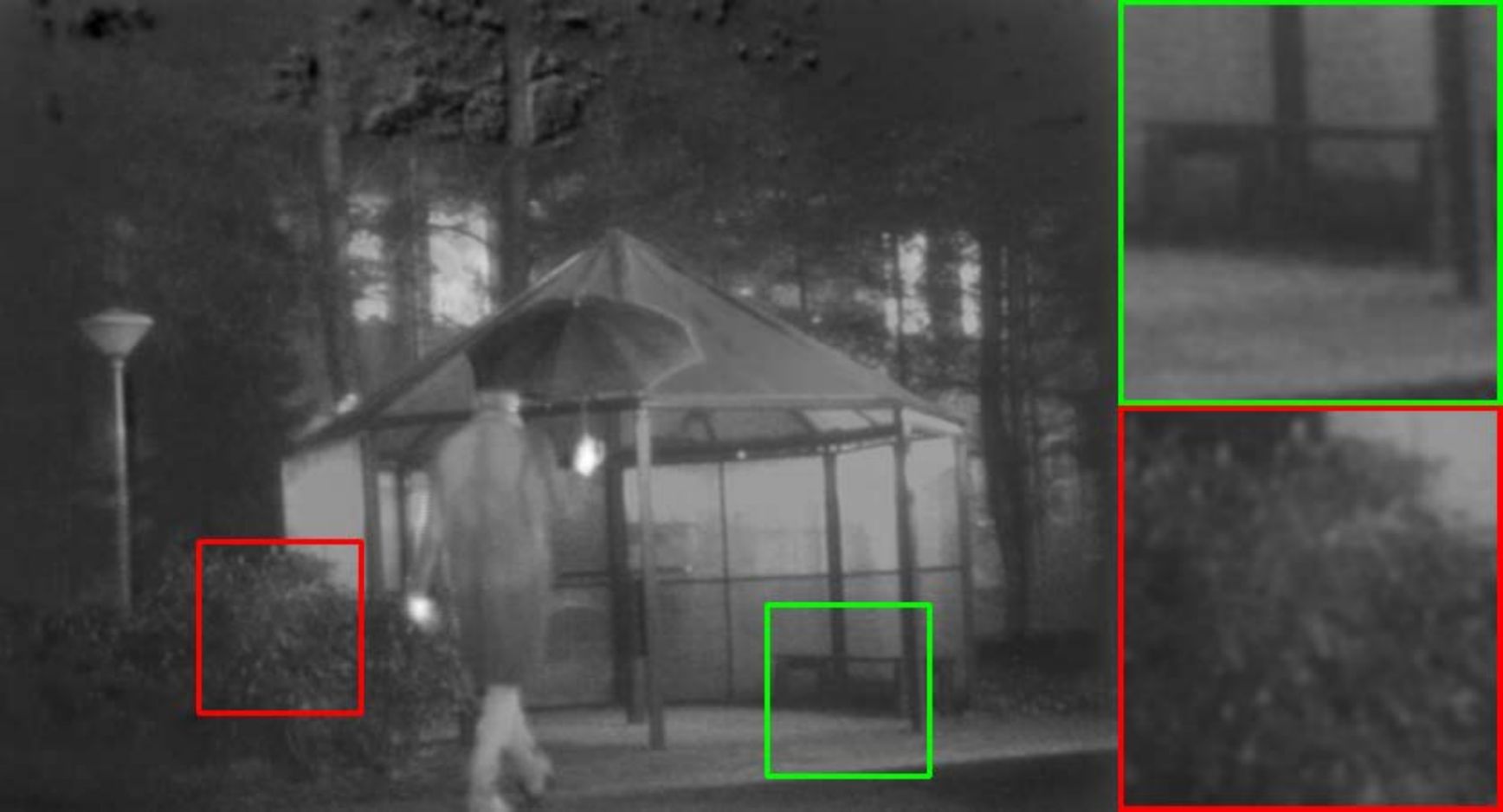}
		&\includegraphics[width=0.17\textwidth,height=0.07\textheight]{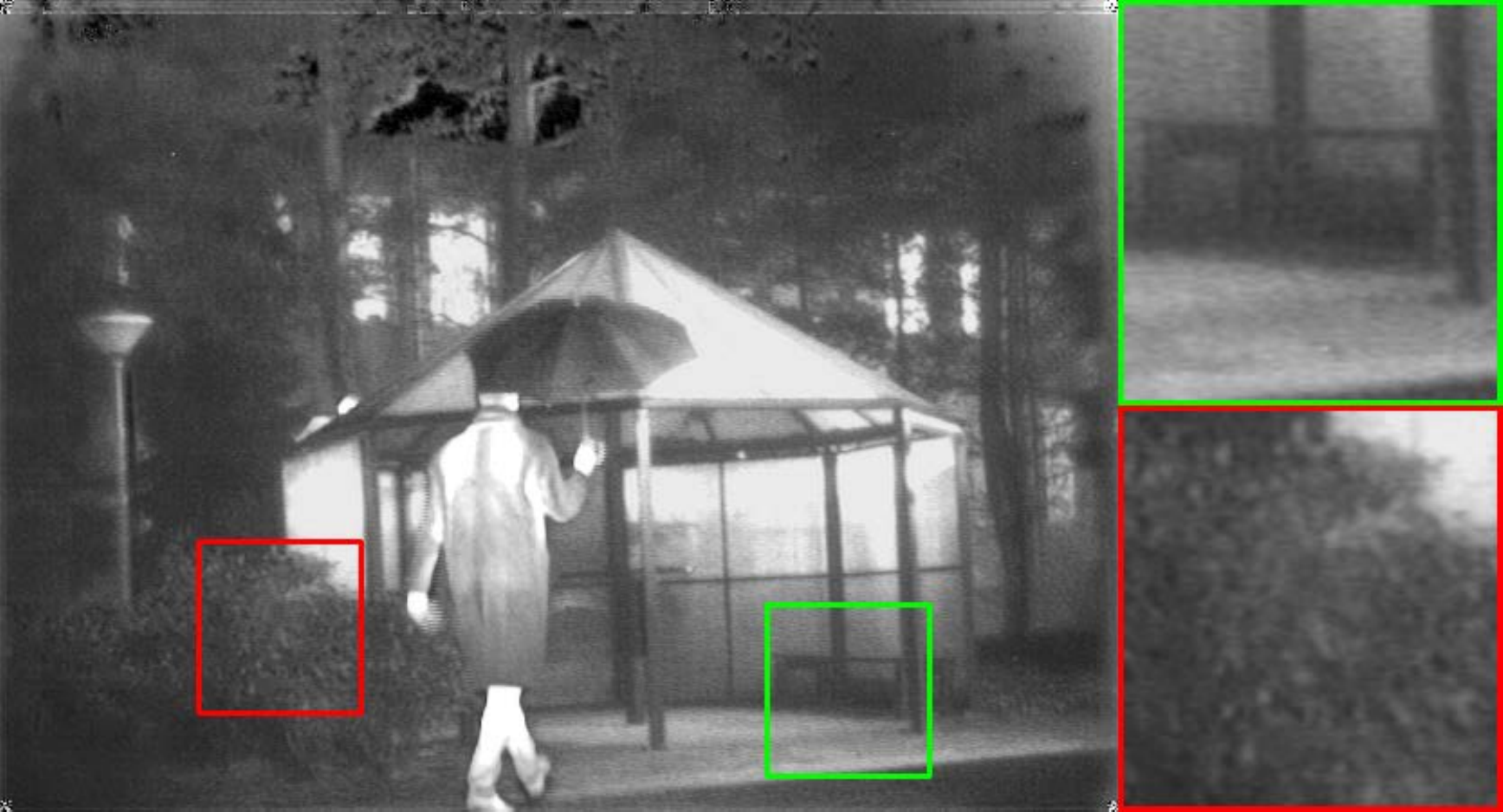}
		&\includegraphics[width=0.17\textwidth,height=0.07\textheight]{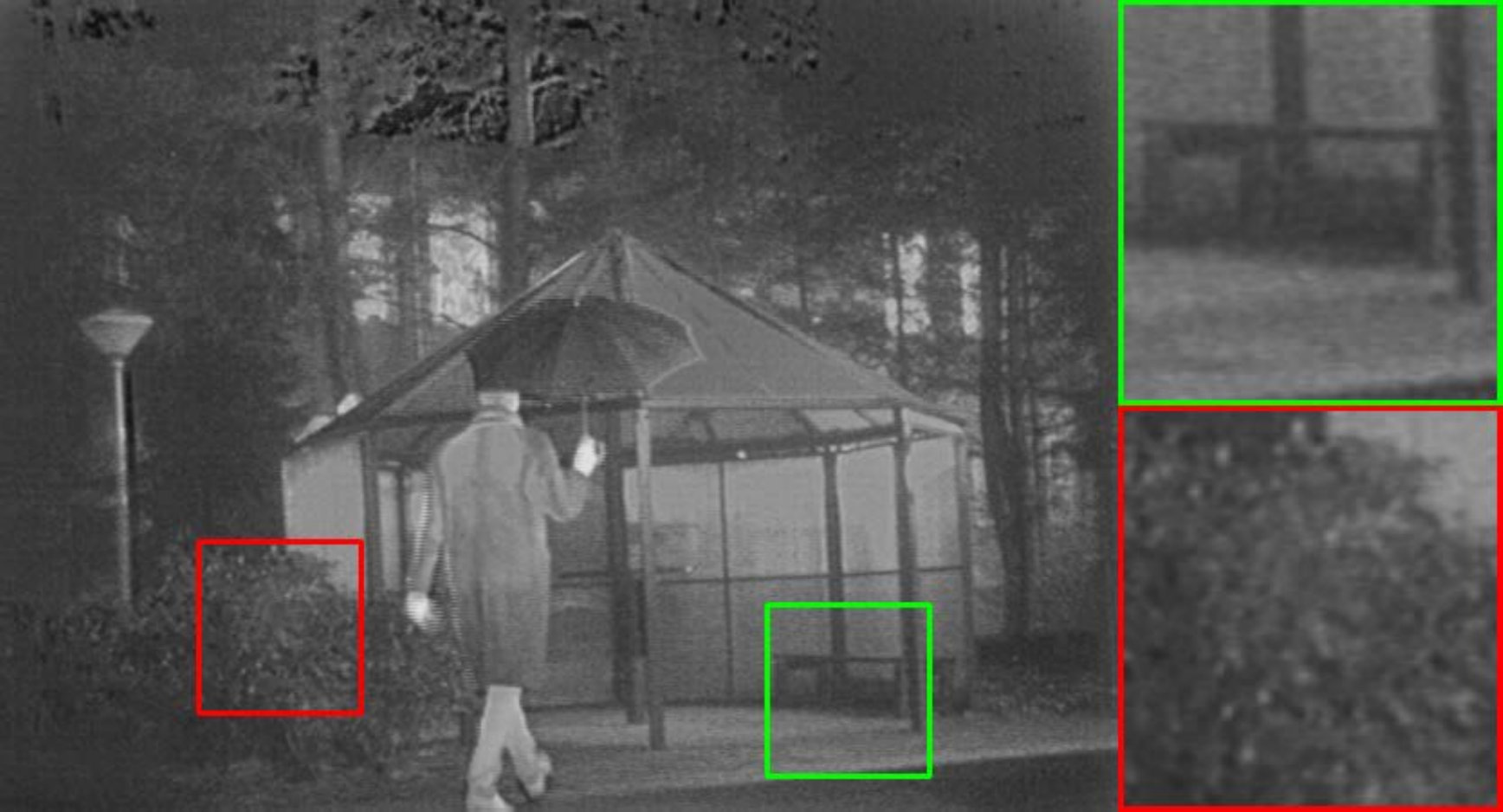}
		&\includegraphics[width=0.17\textwidth,height=0.07\textheight]{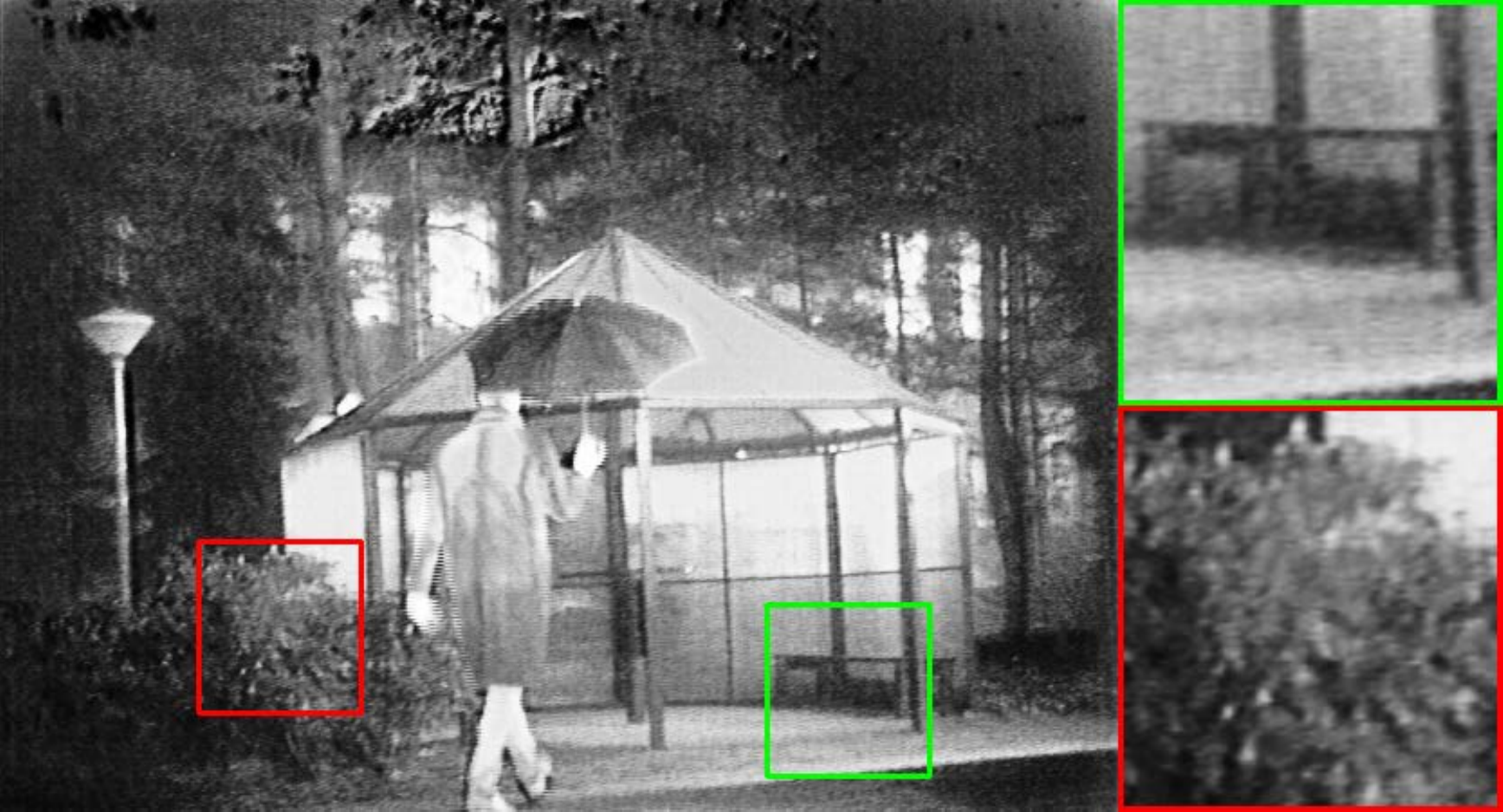}
		\\
		Visible&SwinFusion&RFN&TarDAL&U2Fusion&Ours
		\\
	\end{tabular}
	\caption{Visual comparison between different methods on TNO~dataset.  Our method is better in preserving detailed textural details and distinctive thermal target, especially in the zoomed-in patches.}
	\label{fig:TNO}
\end{figure*}

The difference between the fused image and the source images as $D_{V,F}$ and $D_{R,F}$, which correspond the correlation between the source image and the difference image respectively. SCD is defined as follows:

\begin{equation}
	S C D=r\left(I_{V}, D_{V,F}\right)+r\left(I_{R}, D_{R,F}\right) .
\end{equation}

\noindent\textbf {Visual information fidelity for fusion (VIF)} VIF evaluates image quality based on its fidelity, which determines whether an image is visually-friendly. It assesses the amount of valid information fused from the source image. A larger value indicates better quality. VIF is defined as follows:

\begin{small}
	\begin{equation}
		VIF(I_{V},I_{R},I_{F}) = \sum_{k}p_{k}\frac {{\sum}_{b}FVID_{s,b}(I_{V},I_{R},I_{F})}{{\sum}_{b}FVIND_{s,b}(I_{V},I_{R},I_{F})}
		\label{eq8}
	\end{equation} 
\end{small} 

where $FVID_{s,b}$ is the fusion visual information with distortion, and $FVIND_{s,b}$ denotes the fusion visual information without distortion, in the $b$th block, $s$th sub-band.

\begin{figure*}
	\centering
	\setlength{\tabcolsep}{1pt} 
	
	\includegraphics[width=1.01\textwidth, height=0.385\textheight]{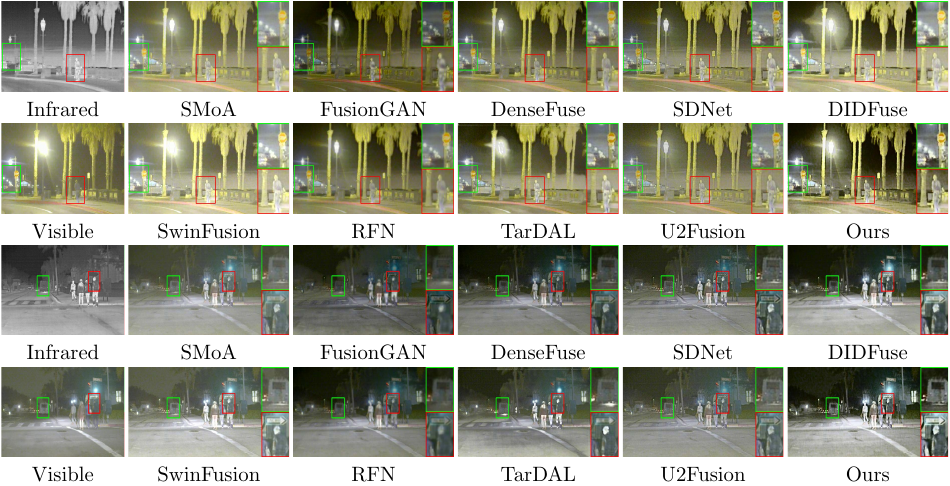}
	\caption{Visual comparison between different methods on Roadscene dataset. Our method has advantage in retaining the most information, \emph{e.g.,} the pedestrians in the patches. Best viewed on screen.}
	\label{fig:Road}
\end{figure*}

\subsection{Results and Analysis on IVIF}
\noindent\textbf{Qualitative Comparisons on TNO} In Figure \ref{fig:TNO}, we compare our CoCoNet with several SOTA methods:
DenseFuse~\cite{li2018densefuse}, 
FusionGAN~\cite{ma2019fusiongan}, PMGI~\cite{PMGI}, DIDFuse~\cite{zhao2020didfuse}, GANMcC~\cite{GANMcC}, RFN~\cite{rfn2021}, MFEIF~\cite{MFEIF2021}, U2Fusion~\cite{U2Fusion2020}, SwinFusion~\cite{Ma2022SwinFusion}, SDNet~\cite{zhang2021sdnet}, SMoA~\cite{liu2021smoa} and TarDAL~\cite{liu2022target} on samples from the TNO dataset. Owing to the proposed contrastive learning, overall, our results render higher contrast and the foreground targets are much brighter (\emph{e.g.}, trees, leaves, and foreground objects framed by the red and green boxes). As indicated in the first picture, the salient object (boxed by a red frame) is sharper and clearer, thanks to the infrared contrast and semantic features extracted from multi-level attention. DenseFuse and DIDFuse can also give clear thermal features, however, their images are not bright enough thus causing degradation of the final visual effects. FusionGAN and RFN fail to provide clear targets (the human in their results appear blurry). On the other hand, we can retain vivid textural details from the visible image as well. As displayed in the second compared picture, the red box showcases abundant and clear details of the leaves, while FusionGAN, DIDFuse and SMoA provide nearly no detailed textures. From a comprehensive view, the proposed method obtains the highest quality of both infrared targets and visible features, and fuses both modalities in a natural manner. Although DenseFuse and FusionGAN can also produce relatively desirable foreground targets, they fail to give clear background information, their images tend to be whether dark or blurry. In a word, our CoCoNet achieves the best balance of both saliency and vivid details.

\begin{table*}[htbp]
	\caption{Quantitative comparison of fourteen methods on the TNO and Roadscene Datasets.}
	\centering
	\resizebox{\textwidth}{!}{
		\begin{tabular}{@{}cccccccc|ccccccc@{}}
			\toprule
			&                    & \multicolumn{6}{c|}{\textbf{TNO}}                                         & \multicolumn{6}{c}{\textbf{RoadScene}}                                    &  \\ \midrule
			& Method                   & \textbf{EN} & \textbf{SD} & \textbf{VIF}   & \textbf{SF}   & \textbf{AG} & \textbf{SCD}& \textbf{EN} & \textbf{SD} & \textbf{VIF}   & \textbf{SF}   & \textbf{AG} & \textbf{SCD} &  \\ \midrule
			& \textbf{DenseFuse}
			&   6.961$\pm$0.318  &   9.835$\pm$0.914  &   0.783$\pm$0.206  &   0.0383$\pm$0.0127  &   3.742$\pm$1.251  &   1.726$\pm$0.251  &   7.143$\pm$0.262  &   9.875$\pm$0.813  &   0.808$\pm$0.107  &   0.0384$\pm$0.0121  &   3.815$\pm$1.313  &   1.728$\pm$0.149  &     \\
			& \textbf{FusionGAN}
			&   6.704$\pm$0.335  &   9.057$\pm$1.101  &   0.608$\pm$0.159  &   0.0289$\pm$0.0097  &   2.764$\pm$0.788  &   1.194$\pm$0.546  &   7.033$\pm$0.304  &   9.994$\pm$0.813  &   0.617$\pm$0.077  &   0.0318$\pm$0.0093  &   3.050$\pm$0.979  &   0.995$\pm$0.552  &     \\
			& \textbf{PMGI}
			&   7.130$\pm$0.204  &   9.984$\pm$0.800  &   0.829$\pm$0.229  &   0.0371$\pm$0.0108  &   3.725$\pm$1.152  &   1.620$\pm$0.381  &   7.293$\pm$0.195  &   10.139$\pm$0.596  &   \textcolor{red}{0.854$\pm$0.097}  &   0.0389$\pm$0.0096  &   3.995$\pm$1.113  &   1.357$\pm$0.558  &     \\
			& \textbf{DIDFuse}
			&   7.208$\pm$0.375  &   \textcolor{blue}{10.200$\pm$1.125}  &   0.800$\pm$0.184  &   \textcolor{blue}{0.0524$\pm$0.0199}  &   4.982$\pm$1.950  &   \textcolor{blue}{1.769$\pm$0.191}  &   7.355$\pm$0.219  &   \textcolor{red}{10.907$\pm$0.659}  &   \textcolor{blue}{0.851$\pm$0.107}  &   0.0481$\pm$0.0155  &   4.723$\pm$1.653  &   1.766$\pm$0.150  &     \\
			& \textbf{GANMcC}
			&   6.881$\pm$0.337  &   9.580$\pm$0.640  &   0.676$\pm$0.189  &   0.0265$\pm$0.0070  &   2.719$\pm$0.763  &   1.590$\pm$0.324  &   7.212$\pm$0.274  &   10.023$\pm$0.640  &   0.757$\pm$0.081  &   0.0325$\pm$0.0076  &   3.403$\pm$0.942  &   1.588$\pm$0.260  &     \\
			& \textbf{RFN}
			&   7.146$\pm$0.245  &   9.946$\pm$0.838  &   0.785$\pm$0.188  &   0.0244$\pm$0.0078  &   2.896$\pm$1.063  &   1.728$\pm$0.256  &   \textcolor{blue}{7.306$\pm$0.257}  &   9.978$\pm$0.763  &   0.803$\pm$0.101  &   0.0273$\pm$0.0078  &   2.882$\pm$1.026  &   \textcolor{blue}{1.773$\pm$0.135}  &     \\
			& \textbf{MFEIF}
			&   6.910$\pm$0.444  &   9.740$\pm$1.182  &   0.788$\pm$0.146  &   0.0327$\pm$0.0138  &   3.291$\pm$1.628  &   1.668$\pm$0.232  &   6.993$\pm$0.288  &   10.008$\pm$0.709  &   0.827$\pm$0.100  &   0.0321$\pm$0.0091  &   3.202$\pm$1.038  &   1.590$\pm$0.181  &     \\
			& \textbf{U2Fusion}
			&   6.650$\pm$0.364  &   9.524$\pm$1.044  &   0.672$\pm$0.147  &   0.0374$\pm$0.0148  &   3.938$\pm$1.660  &   1.484$\pm$0.311  &   6.990$\pm$0.321  &   9.878$\pm$0.813  &   0.749$\pm$0.113  &   0.0455$\pm$0.0165  &   4.686$\pm$1.841  &   1.381$\pm$0.377  &     \\
			& \textbf{SwinFuse}
			&   7.015$\pm$0.463  &   9.866$\pm$1.117  &   \textcolor{blue}{0.887$\pm$0.134}  &   0.0494$\pm$0.0229  &   4.882$\pm$2.534  &   1.623$\pm$0.209  &   6.944$\pm$0.357  &   10.405$\pm$0.692  &   0.849$\pm$0.101  &   0.0414$\pm$0.0136  &   3.872$\pm$1.447  &   1.601$\pm$0.199  &     \\
			& \textbf{SDNet}
			&   6.890$\pm$0.309  &   9.403$\pm$1.171  &   0.749$\pm$0.195  &   0.0512$\pm$0.0182  &   \textcolor{blue}{4.991$\pm$1.772}  &   1.493$\pm$0.312  &   7.240$\pm$0.246  &   10.039$\pm$0.682  &   0.806$\pm$0.095  &   \textcolor{blue}{0.0516$\pm$0.0156}  &   \textcolor{blue}{5.199$\pm$1.760}  &   1.420$\pm$0.305  &     \\
			& \textbf{SMoA}
			&   6.782$\pm$0.454  &   9.476$\pm$1.187  &   0.719$\pm$0.135  &   0.0318$\pm$0.0130  &   3.037$\pm$1.478  &   1.603$\pm$0.267  &   6.908$\pm$0.297  &   9.678$\pm$0.813  &   0.765$\pm$0.102  &   0.0315$\pm$0.0092  &   3.124$\pm$0.991  &   1.487$\pm$0.239  &     \\
			& \textbf{TarDAL}
			&   \textcolor{blue}{7.238$\pm$0.427}  &   9.929$\pm$1.102  &   0.803$\pm$0.119  &   0.0522$\pm$0.0117  &   4.645$\pm$1.415  &   1.550$\pm$0.304  &   7.210$\pm$0.245  &   10.111$\pm$0.784  &   0.790$\pm$0.086  &   0.0475$\pm$0.0068  &   4.044$\pm$0.989  &   1.339$\pm$0.295  &     \\
			& \textbf{Ours}
			&   \textcolor{red}{\textbf{7.776$\pm$0.081}}   &   \textcolor{red}{\textbf{10.591$\pm$0.477}}   &   \textcolor{red}{\textbf{0.893$\pm$0.268}}   &   \textcolor{red}{\textbf{0.0831$\pm$0.0280}}   &   \textcolor{red}{\textbf{7.967$\pm$2.696}}   &   \textcolor{red}{\textbf{1.794$\pm$0.091}}   &   \textcolor{red}{\textbf{7.693$\pm$0.094}}   &   \textcolor{blue}{\textbf{10.674$\pm$0.509}}   &   \textcolor{red}{\textbf{0.854$\pm$0.091}}   &   \textcolor{red}{\textbf{0.0718$\pm$0.0193}}   &   \textcolor{red}{\textbf{6.537$\pm$2.001}}   &   \textcolor{red}{\textbf{1.817$\pm$0.051}}   &     \\
			
			\bottomrule
		\end{tabular}
	}
	\label{T_NUM}
\end{table*}

\noindent\textbf{Qualitative Comparison on RoadScene} We also display the visual comparison of our method and the state-of-the-arts on typical realistic driving scenes (\emph{e.g.}, roads, symbol signs and pedestrians) in Figure \ref{fig:Road}. Generally, all of these methods can fuse the thermal radiation and textural structure details from the source image to some extent. However, RFN and FusionGAN tend to have blur edges. Our method, TarDAL and SDNet all produce visual-friendly fusion images, among which our salient targets (\emph{i.e.}, the pedestrians framed by red boxes) are much brighter and more conspicuous. Unfortunately, RFN fails to combine as much infrared information as other methods, and this causes their objects less clear. On the other hand, our CoCoNet also achieves the best performance in terms of vivid structure details. In the second compared picture, one can see that the truck indicated by the green box renders much better textures even in a relatively low-light scene. Nevertheless, FusionGAN and PMGI are not able to tackle such problems, they appear dark and fuzzy in the boxed region. We give this credit to the proposed multi-level attention, in which we fuse more high-level features and rearrange the importance of different feature channels. Because of this, our images are sharper and clearer. Although TarDAL can also give good quality fusion result, the output tends to be influenced by the glow, thus degrading the visibility. Overall, we have reached the first class in road scenes and our method is more robust in various scenarios.

\begin{figure*}
	\centering
	\setlength{\tabcolsep}{1pt} 
	\begin{tabular}{c}
		
		\includegraphics[width=0.98\textwidth]{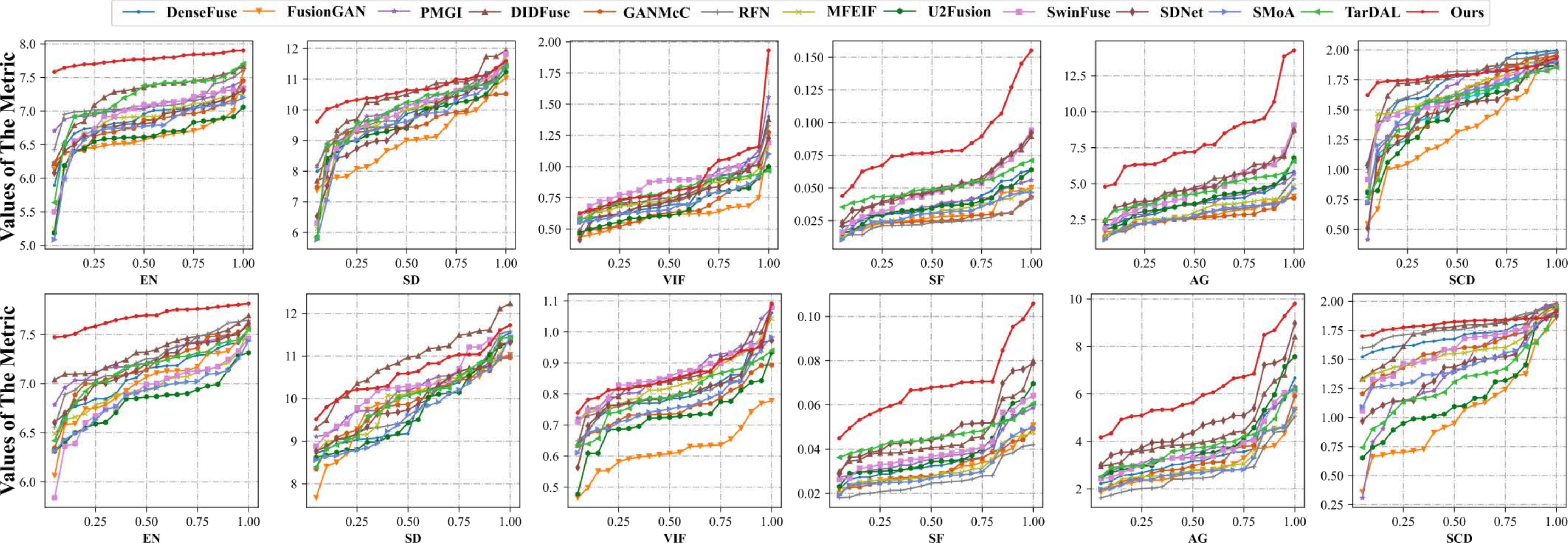}	
	\end{tabular}
	\caption{Quantitative comparison of five metrics, i.e., EN, SD, VIF, SF, AG and SCD, on 20 image pairs from the TNO dataset (the first row) and other 20 image pairs from the RoadScene dataset (the second row). The thirteen state-of-the-art methods are used for comparison. A single point (x, y) on the curve denotes that there are (100${\times}$x)\% percent of image pairs that have metric values no more than y. }
	\label{fig:TNO_NUM}
\end{figure*}

\noindent\textbf{Quantitative Comparisons on TNO} 
We provide evaluations of all methods on several important metrics discussed in the above section. The results are shown in Figure \ref{fig:TNO_NUM} . It is obvious that on TNO dataset, we achieve the highest results on all six metrics, which demonstrates that the proposed CoCoNet is able to make full use of vital features from the source image.
Besides, in Table \ref{T_NUM}, we also display the average value and standard deviation of each metric to demonstrate our overall performance. For TNO dataset, it's notable that on SF and AG, we show an overwhelming advantage over the second best method, \emph{e.g.}, SwinFuse and DIDFuse by even 58$\%$ higher score. This further demonstrates that due to the proposed self-adaptive learning, we are able to generate images with more grey levels, hence more informative features. DIDFuse also achieves relatively excellent results in terms of SD and SCD. TarDAL produces satisfying results on EN.

\noindent\textbf{Quantitative Comparisons on RoadScene} We can process not only military scenes, but also complex driving scenarios. Thanks to the contrast learning, the fused image contains distinctive grey levels with high contrast. Figure \ref{fig:TNO_NUM} also displays a quantitative comparison evaluated on RoadScene dataset at the second row. Generally, we achieve the best on EN, SF, AG, SCD and VIF and SOTA results on SD.
In Table \ref{T_NUM}, an evaluation of all methods is listed.  This showcases that our method can fuse the most valid information from source images, and our visibility and sharpness are of the top level as well. It's worth mentioning that the proposed CoCoNet is 58$\%$ higher than the second best SDNet on SF. This indicates much more information with abundant characteristics is included in our results even in complex real scenes, due to self-adaptive learning strategy.

\subsection{Ablation Study}
In this part, we discuss the necessity of different modules in our proposed CoCoNet. 

\begin{figure}
	\centering
	\setlength{\tabcolsep}{1pt}
	\begin{tabular}{c}
		\includegraphics[width=0.48\textwidth]{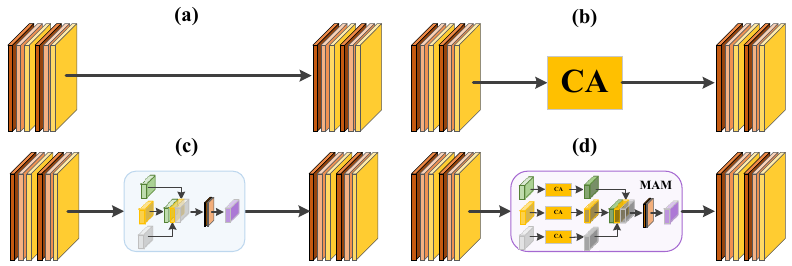}
		\\
	\end{tabular}
	\caption{Visual display of different architectures of MAM, where (a) skip connection only, (b) skip connection with channel attention, (c) skip connection with channel attention and backbone features, (d) our full MAM.}
	\label{fig:mamcpr}
\end{figure}

\noindent\textbf{Effectiveness of MAM} To ablate the use of multi-level attention mechanism, we remove either features from pretrained VGG backbone, channel attention, or both, to generate three variations of the MAM, respectively. In Figure \ref{fig:mamcpr}, we illustrate these variations along with  our full MAM, where (a) removes both channel attention and pretrained VGG backbone, (b) removes only pretrained backbone, (c) removes only channal attention from MAM and (d) our full MAM.

\begin{table*}[!htb]
	\caption{Quantitative comparison of different model architectures.}
	\centering
	
	\renewcommand\arraystretch{1.1} 
	\setlength{\tabcolsep}{1.2mm}
	\begin{tabular}{|c|cccccc|cccccc|}
		\hline
		\multirow{2}{*}{ Model}&\multicolumn{6}{c|}{ \textbf{TNO Dateset}}&\multicolumn{6}{c|}{ \textbf{Roadscene Dataset} }\\
		\hhline{~|*{12}{|-}} 
		& EN & SD & VIF & SF ~& AG  & SCD & EN & SD & VIF & SF ~& AG  & SCD\\
		\hline
		$\mathrm{w/o~_{ca\&vgg}}$& \small 6.564& \small8.965&\small 0.705&\small 0.0531&\small 4.166&\small \textcolor{blue}{\textbf{1.736}}&\small 6.637&\small 9.431 &\small 0.698 &\small 0.0430 &\small 4.31 & \small 1.511\\
		\hline 
		$\mathrm{w/o~_{vgg}}$& \small6.786&\small 9.288& \small0.774&\small 0.0596&\small 5.645&\small 1.681&\small 6.839&\small 9.611 &\small 0.769 &\small 0.0514 &\small 5.033 &\small 1.524 \\
		\hline 
		$\mathrm{w/o~_{ca}}$&\small \textcolor{blue}{\textbf{7.325}}&\small \textcolor{blue}{\textbf{9.511}}&\small \textcolor{blue}{\textbf{0.823}}&\small \textcolor{blue}{\textbf{0.0675}}&\small \textcolor{blue}{\textbf{6.536}}
		&\small 1.703&\small \textcolor{blue}{\textbf{7.288}}&\small \textcolor{blue}{\textbf{9.980}}&\small \textcolor{blue}{\textbf{0.833}} &\small \textcolor{blue}{\textbf{0.0591}} &\small \textcolor{blue}{\textbf{5.653}} &\small \textcolor{blue}{\textbf{1.623}} \\
		\hline 
		
		\textbf{Ours} 
		&\small \textcolor{red}{\textbf{7.776}}   
		&\small \textcolor{red}{\textbf{10.591}}   
		& \small \textcolor{red}{\textbf{0.893}}   
		& \small \textcolor{red}{\textbf{0.0831}}   
		&\small \textcolor{red}{\textbf{7.967}}   
		&\small \textcolor{red}{\textbf{1.794}}   
		&\small \textcolor{red}{\textbf{7.693}}   
		& \small \textcolor{red}{\textbf{10.674}}  
		&\small \textcolor{red}{\textbf{0.854}}   
		&\small \textcolor{red}{\textbf{0.0718}}   
		&\small \textcolor{red}{\textbf{6.537}}   
		&\small \textcolor{red}{\textbf{1.817}}   \\
		\hline 		
	\end{tabular}
	\label{tab:structure}
\end{table*}

\begin{figure*}
	\centering
	\setlength{\tabcolsep}{1pt} 
	
	\includegraphics[width=0.99\textwidth, height=0.29\textheight]{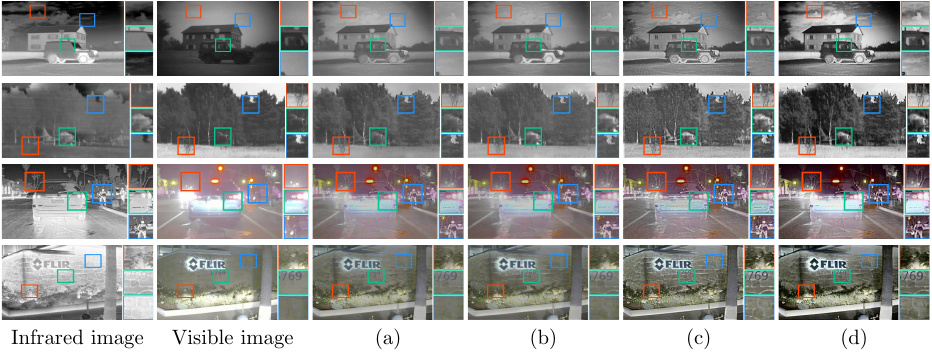}	
	\caption{Visual comparison between different model architectures of the proposed MAM, where (a) skip connection only, (b) skip connection with channel attention, (c) skip connection with channel attention and backbone features, (d) our full MAM. The full MAM is better in preserving informative features, especially in the zoomed-in patches. }
	\label{fig:modlevisual}
\end{figure*}

\begin{figure}
	\centering
	\setlength{\tabcolsep}{1pt}
	\begin{tabular}{c}
		\includegraphics[width=0.45\textwidth]{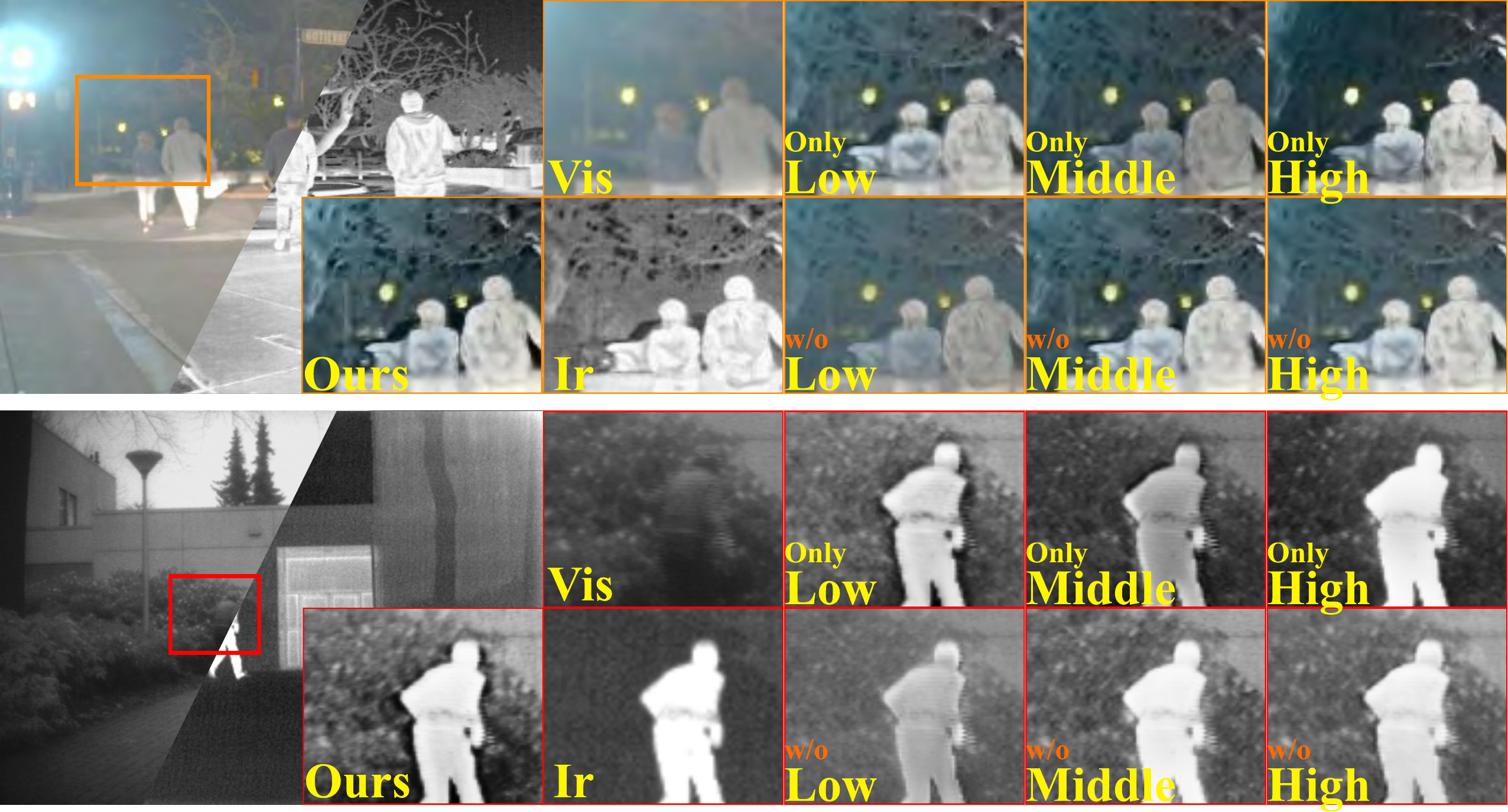}
		\\
	\end{tabular}
	\caption{Visual display of variants that utilize different VGG features.}
	\label{fig:vgg_visual}
\end{figure}

\begin{figure*}
	\centering
	\setlength{\tabcolsep}{1pt} 
	
	\includegraphics[width=0.92\textwidth, height=0.23\textheight]{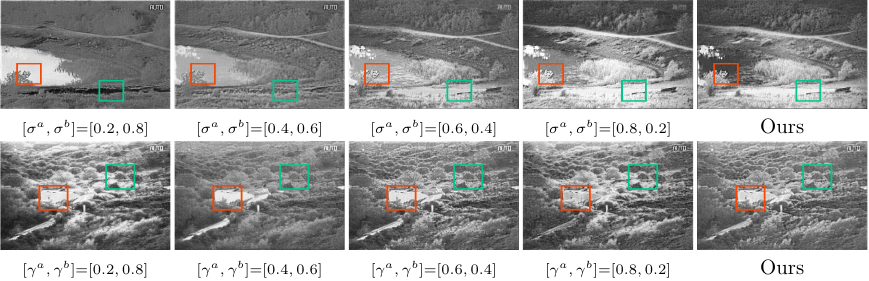}	
	\caption{	Visual comparison among different parameter setting of $\sigma$ and $\gamma$. The top row shows the results with manually-craft  $\sigma$ and the proposed self-adaptive weight, the second row shows the results on $\gamma$ correspondingly. Note that the self-adaptive weight is better in preserving detailed textural details and considerable contrast, especially in the zoomed-in patches. Best viewed on screen. }
	\label{fig:ssim-weight}
\end{figure*}

The numerical results are reported in Table \ref{tab:structure}, where $\mathrm{w/o~_{ca\&vgg}}$, $\mathrm{w/o~_{vgg}}$, $\mathrm{w/o~_{ca}}$ and $ours$ correspond to the aforementioned (a), (b), (c) and (d), respectively. It's notable that removing the pretrained VGG features causes a significant drop in metrics on both TNO and RoadScene datasets, which proves the importance of complementary high level features extracted from the pretrained backbone. To prove the importance of channel attention, we also notice a boost of performance from $\mathrm{w/o~_{ca\&vgg}}$ to $\mathrm{w/o~_{vgg}}$ and from $\mathrm{w/o~_{ca}}$ to $ours$ on EN, SF, SD and VIF. This demonstrates that channel attention rearranges the weights of different channels obtained from both modalities, benefiting the overall textural details of the fused image.

Moreover, visual comparisons of $\mathrm{w/o~_{ca\&vgg}}$ to $ours$ are also provided in Figure \ref{fig:modlevisual}. It is obvious that by adding the pretrained VGG features, the acuteness of the fused image is greatly improved. Consequently, column (d) and column (c) are much sharper than (b) and (a), \emph{e.g.}, the alphabets indicated by green boxes are clearer in column (c) and (d), and the global structure information is more evident. By reorganizing feature channels, channel attention enables our network to further remove undesirable noises and retain richer details, which is reflected on the differences between column (c) and (d). 

\begin{table*}[!htb]
	\caption{Quantitative comparison of different features from VGG.}
	\centering
	\small
	\renewcommand\arraystretch{1.1} 
	\setlength{\tabcolsep}{1.2mm}
	\begin{tabular}{|c|cccccc|cccccc|}
		\hline
		\multirow{2}{*}{ Model}&\multicolumn{6}{c|}{ \textbf{TNO Dateset}}&\multicolumn{6}{c|}{ \textbf{Roadscene Dataset} }\\
		\hhline{~|*{12}{|-}}
		& EN & SD & VIF & SF ~& AG  & SCD & EN & SD & VIF & SF ~& AG  & SCD\\
		\hline 
		Only Low & \color{blue}7.387 & \color{blue}9.977 & 0.797 & \color{blue}0.0694 & \color{blue}6.806 & 1.774 & \color{blue}7.491 & \color{blue}10.178 & 0.768 & \color{blue}0.0648 & \color{blue}6.155 & \color{blue}1.652\\
		\hline 
		Only Middle & 7.078 & 9.694 & 0.764 & 0.0497 & 4.695 & 1.688 & 7.155 & 9.973  & 0.753 & 0.0483 & 4.315 & 1.390 \\
		\hline 
		Only High & 7.081 & 9.642 & 0.781 & 0.0504 & 4.719 & 1.621 & 7.255 & 9.980  & \color{blue}0.787 & 0.0488 & 4.375 & 1.424 \\
		\hline 
		w/o Low & 6.796 & 9.386 & 0.712 & 0.0444 & 4.163 & 1.544 & 6.878 & 9.516  & 0.704 & 0.0396 & 3.570 & 1.025 \\
		\hline 
		
		w/o Middle &7.338 & 9.678 & 0.756 & 0.0685 & 6.715 & 1.693 & 7.394 & 9.873  & 0.728 & 0.0638 & 6.050 & 1.475 \\
		\hline 
		
		w/o High & 7.294 & 9.555 &\color{blue} 0.802 & 0.0579 & 5.762 & \color{blue}1.788 & 7.413 & 9.830  & 0.775 & 0.0582 & 5.571 & 1.641 \\
		\hline 
		\textbf{Ours} 
		& \textcolor{red}{\textbf{7.776}}   
		& \textcolor{red}{\textbf{10.591}}   
		& \textcolor{red}{\textbf{0.893}}   
		& \textcolor{red}{\textbf{0.0831}}   
		& \textcolor{red}{\textbf{7.967}}   
		& \textcolor{red}{\textbf{1.794}}   
		& \textcolor{red}{\textbf{7.693}}   
		& \textcolor{red}{\textbf{10.674}}  
		& \textcolor{red}{\textbf{0.854}}   
		& \textcolor{red}{\textbf{0.0718}}   
		& \textcolor{red}{\textbf{6.537}}   
		& \textcolor{red}{\textbf{1.817}}   \\
		\hline 		
	\end{tabular}
	\label{tab:VGG}
\end{table*}

\noindent\textbf{Ablation of Features from VGG}~ Considering that the low/middle/high-level features (64/128/256 channels) extracted by VGG have different characteristics, we created six new variants for experimentation by using these three features separately, including those containing only low/middle/or high-level features, as well as those without them.

Figure~\ref{fig:vgg_visual} provides visual comparison images. It can be observed that the features exhibit similar trends in both datasets. Low-level features typically have a strong response to edge and color information in the image. The variants without low-level features lose a large amount of detail information and produce color deviation. The variants without middle-level and high-level features exhibit some degradation compared to the complete model. Among the variants with only one type of feature, the variant with only low-level features achieves relatively good visualization effect, while the variant with only high-level features has excellent overall contrast. The fusion result of the complete model maintains rich detail information (such as the branches in the upper scene and the bushes in the lower scene), while also achieving prominent highlights for significant individuals and high contrast for the scene.

In Table~\ref{tab:VGG}, we provided quantitative comparison results on two datasets. The variant with only the lowest-level features achieved the second place in most metrics, while the variant without the lowest-level features performed the worst. This indicates that low-level features provide more edge and color information, thus better describing the image. On top of the low-level features, adding middle/high-level features actually had a negative impact. The proposed complete method, which integrates all levels of features, achieved the best performance in all metrics.

\noindent\textbf{Effectiveness of self-adaptive learning} Additional experiments are conducted on TNO and RoadScene datasets to validate the effectiveness of the self-adaptive learning. In Figure \ref{fig:ssim-weight}, we visualize performance of several training weights designed in a hand-crafted manner and the proposed self-adaptive weights measured based on internal characteristics of the source image. Specifically, ${\sigma}$ and ${\gamma}$ are the weights for SSIM and MSE, respectively. We display fused results where several possible fixed weight combinations are adopted (e.g., ${\sigma^a}$ ranging from 0 to 1 while making sure the sum of ${\sigma^a}$ and ${\sigma^b}$ to be 1) to compare with the self-adaptive strategy (presented in the last column). As Figure \ref {fig:ssim-weight} suggests, the self-adaptive manner achieves much higher contrast, stressing both salient thermal target and vivid details from the visible image by using average gradient and entropy information, as indicated by the green and res boxes. The grass details are clearer with brighter pixels, and the building structure preserves fined edges. Overall, automatically learned weights can generate images with better global contrast.

\begin{figure*}
	\centering
	\setlength{\tabcolsep}{1pt}
	\begin{tabular}{c}
		\includegraphics[width=0.95\textwidth]{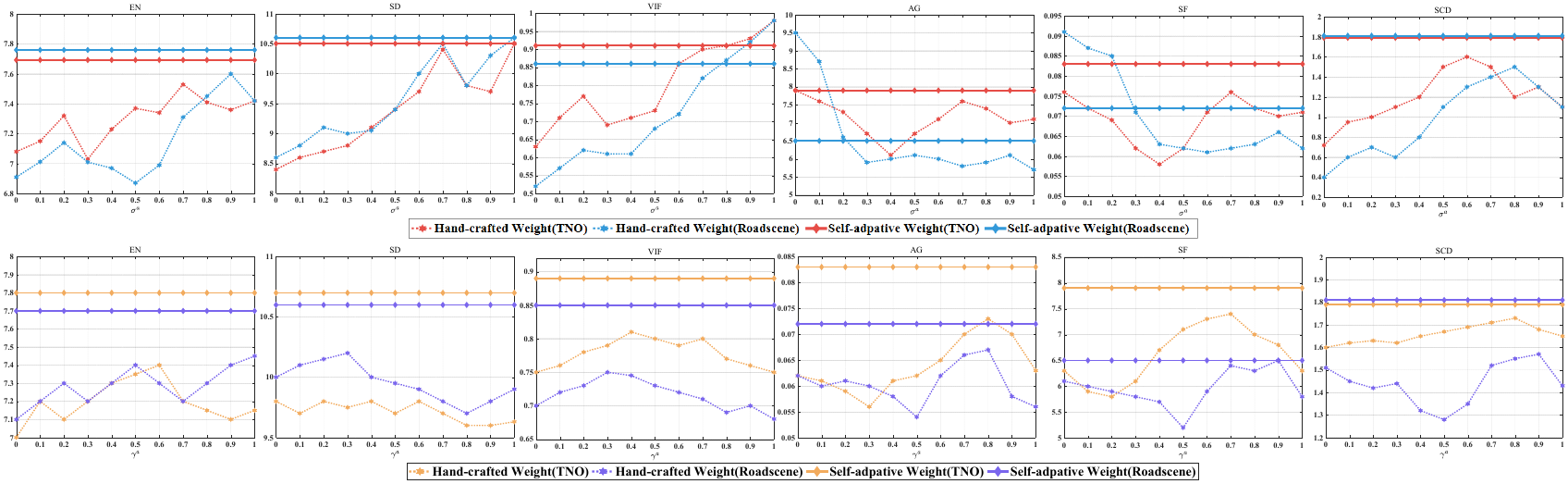}
		\\
	\end{tabular}
	\caption{Quantitative comparison between hand-crafted parameter and our self-adaptive weight on two datasets in terms of $\sigma,\gamma$, respectively. The top row is tested on TNO dataset while the bottom row is verified on Roadscene dataset. In the diagram, dotted lines represent the metric value using manually crafted weight parameters, tuning from 0 to 1. To make the comparison more obvious, the solid line denotes the quantitative value using self-adaptive weight.}
	\label{fig:pic}
\end{figure*}

\begin{figure}
	\centering
	\setlength{\tabcolsep}{1pt} 
	
	\includegraphics[width=0.48\textwidth]{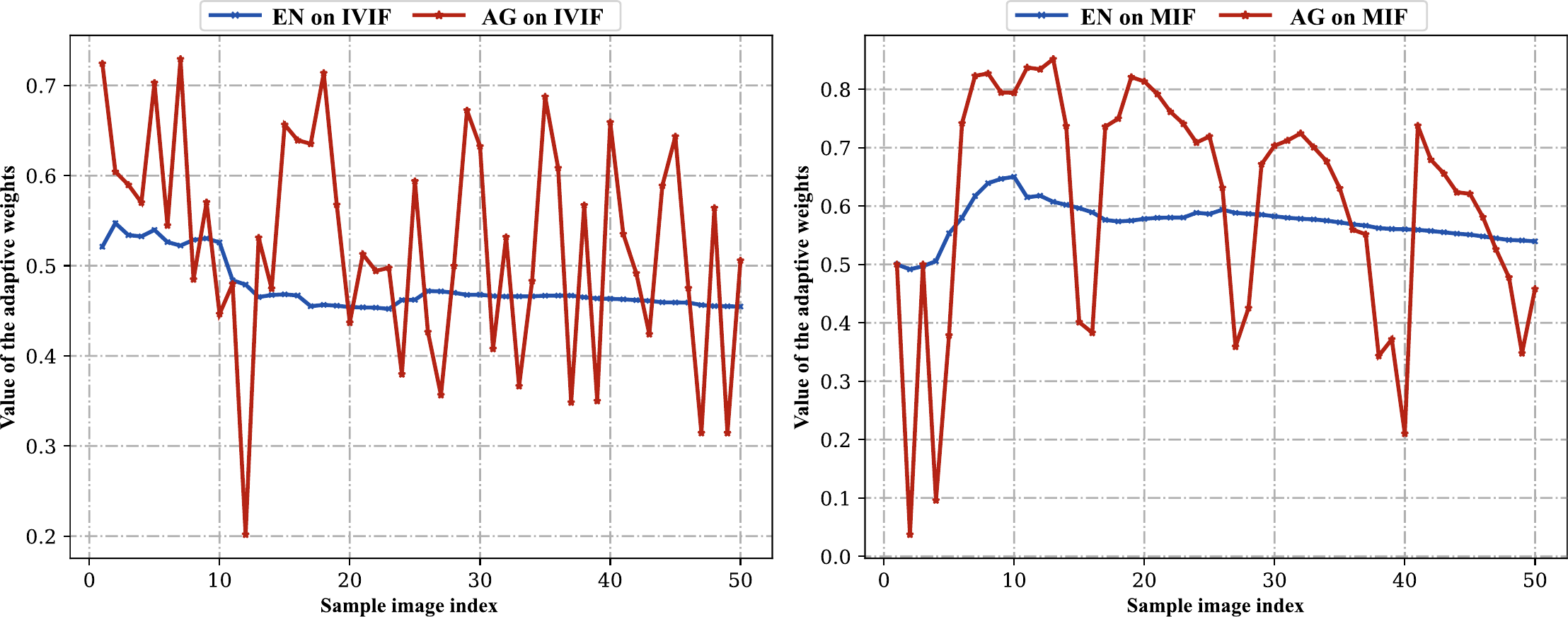}
	
	\caption{Value change of the adaptive weights on different tasks.}
	\label{fig:egag} 
\end{figure}

\begin{figure*}
	\centering
	\setlength{\tabcolsep}{1pt} 
	
	\includegraphics[width=0.94\textwidth]{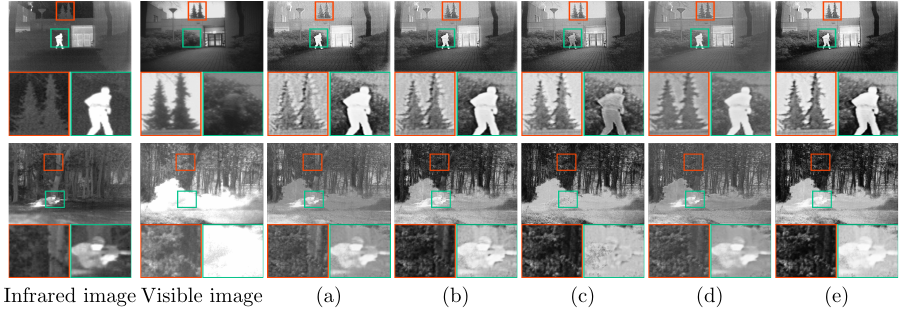}
	\caption{Visual comparison of different settings of contrastive regularization, where (a) no contrastive regularization, (b) target constrain only, (c) detail constrain only, (d) contrast with random positive and negative samples, and (e) the proposed contrastive regularization. Our method is better in preserving both salient targets and faithful details, especially in the zoomed-in patches. Best viewed on screen.}
	\label{fig:contristive}
\end{figure*}

Besides, we plot quantitative results in Figure \ref {fig:pic} to further demonstrate our effectiveness. Results in the first raw report the performance of six metrics where ${\sigma}$ is fixed to a certain value between 0 and 1, while ${\gamma}$ is self-adaptive. Likewise, in the second row, ${\gamma}$ is hand-crafted, while ${\sigma}$ is learned based on source images. We notice that for both SSIM and MSE weights, hand-crafted manner fails to adapt to characteristics in various images, thus reporting lower scores on six evaluation metrics in most cases. On TNO dataset, our strategy achieves around 0.2 points higher than fixed SSIM weights on SCD, which is higher than the best hand-crafted weights. Despite that hand-crafted weights outperform ours on certain points (e.g., fixed weights are slightly better on VIF when $\sigma$ is set to 0.8), we generate images of higher qualities by employing features from a single image itself, which is robust to source images with diverse characteristics. 

In Figure~\ref{fig:egag}, we have visualized the specific weight values of randomly selected 50 image pairs from the TNO and the MRI-PET datasets. It is shown that the $\sigma$ line and $\gamma$ line fluctuate on both datasets, which indicates the dynamic nature of the distinctive structures between modalities. This further proves that under our loss setting, fixed $\sigma$ and $\gamma$ weights could not fully exploit the structural information/pixel distribution of the sources for fusion.

\noindent\textbf{Effectiveness of contrastive learning} Experiments are further conducted to verify the effectiveness of the proposed contrastive regularization, the visual comparisons can be found in Figure \ref {fig:contristive}, where (a) means results with neither target nor detail constraint, column (b) denotes results with target constrain only and (c) denotes results with detail constrain only. To better showcase the importance of sample selection in the contrastive regularization, in column (d), we contrast images using samples that are not corresponded to each other (i.e., image patches for contrast are randomly selected from the whole dataset). The full contrastive learning is given in column (e). Obviously, the lack of both infrared and visible contrast brings unwanted noises to the fused results, causing the edge blurry (e.g., in column (a), the tree details in the red boxes are less clearer with noisy pixels). In column (b), we present results with infrared regularization only. Despite enhancement of the salient thermal target, we still notice degeneration of image quality on the visible details, making it hard to render vivid textural information. This proves that removing regularization on the visible part could harm textural details in the fusion process. In column (c), we display results with only visible constraints by removing the infrared enhancement. It is obvious that the thermal target (e.g.,the standing person indicated by the green box) is abundant in visible details but less bright for the absence of enough guidance of infrared features. Based on results from column (d), we argue that patch sampling is the most significant in contrastive regularization. In absence of correct guidance, the fused image is degraded, with its pixel darker and less clearer, e.g., the person is less brighter and the image appears blurry in general. In conclusion, both target and detail constrains are indispensable to achieve our initial goal, that is, combining both salient thermal targets from the infrared image and vivid background details from the visible image.

\begin{figure}
	\centering
	\setlength{\tabcolsep}{1pt}
	\begin{tabular}{ccc}
		\includegraphics[width=0.15\textwidth]{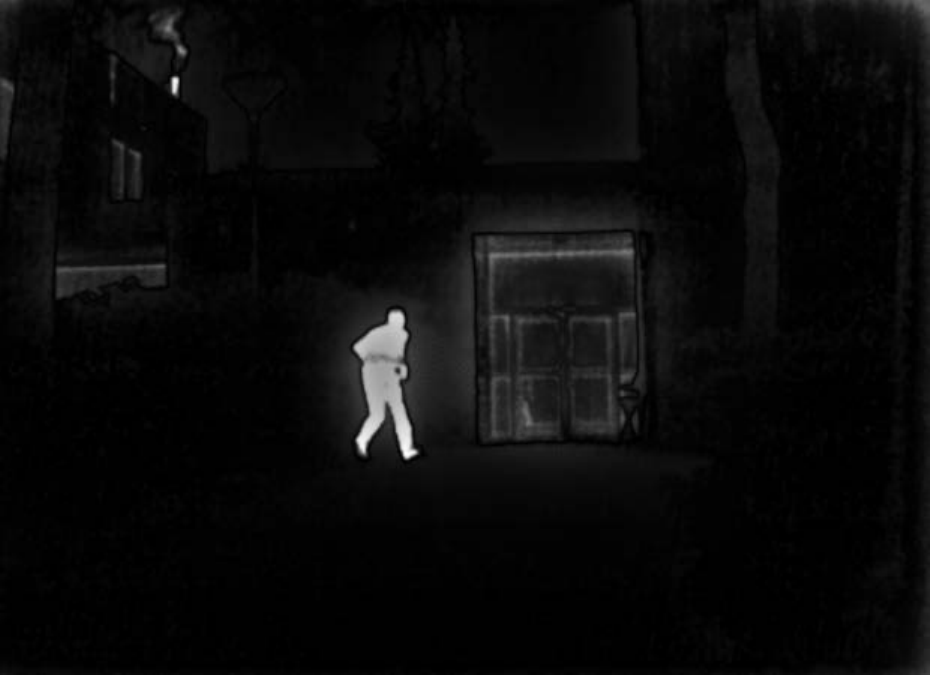}
		&\includegraphics[width=0.15\textwidth]{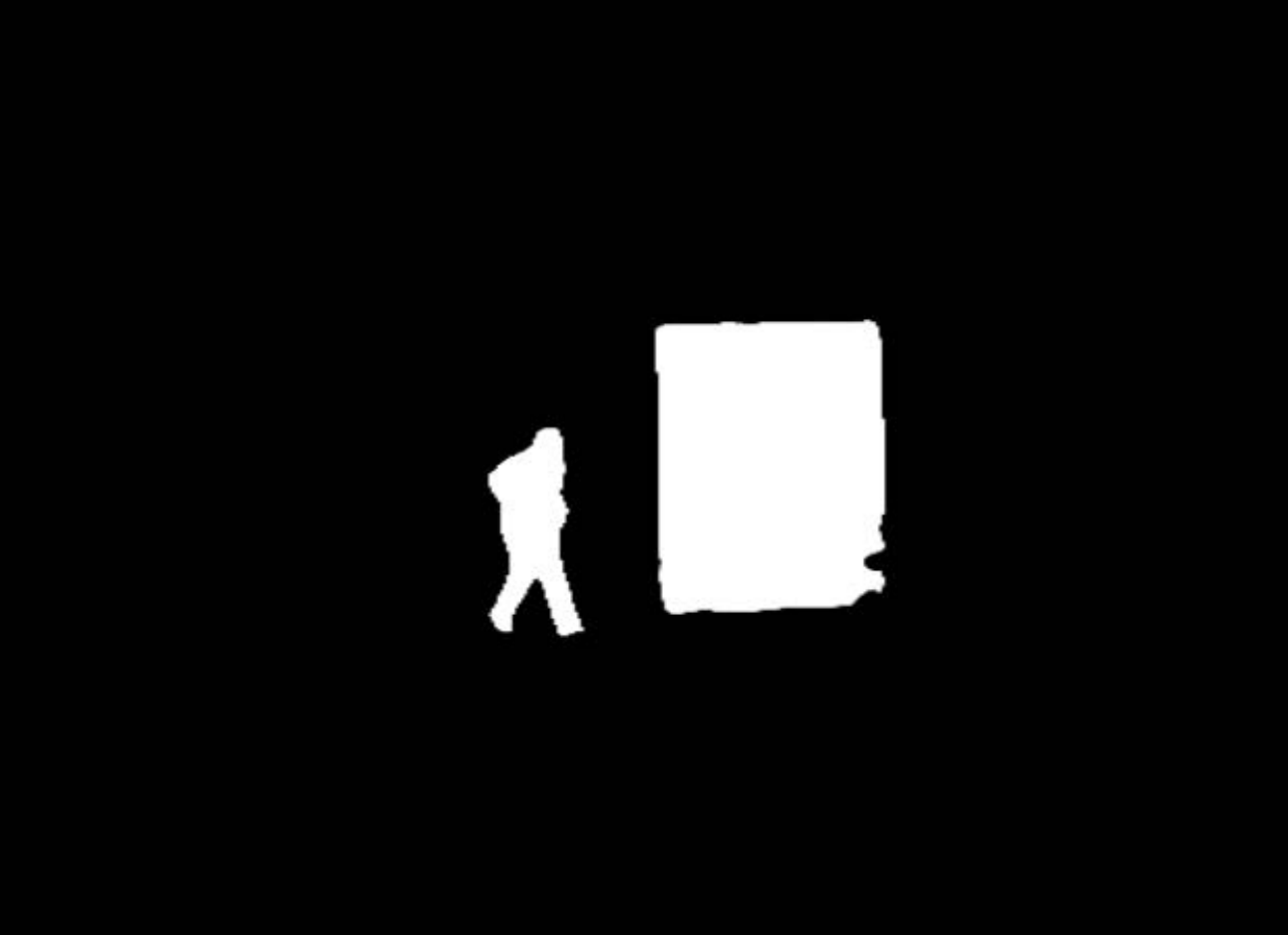}
		&\includegraphics[width=0.15\textwidth]{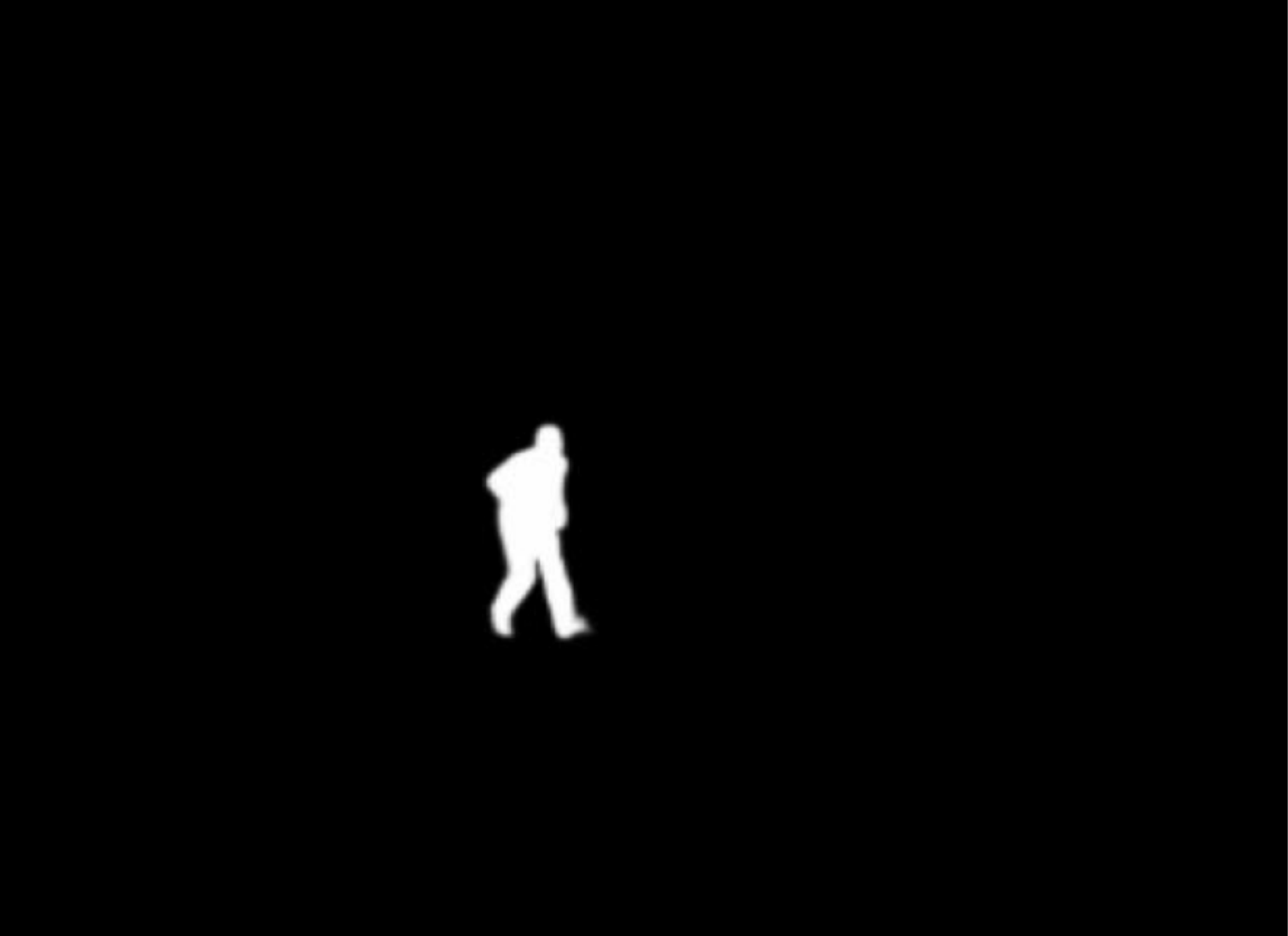}
		\\
		$\mathcal{TM}$&$\mathcal{UM}$&$\mathcal{M}$
		\\
		\includegraphics[width=0.15\textwidth]{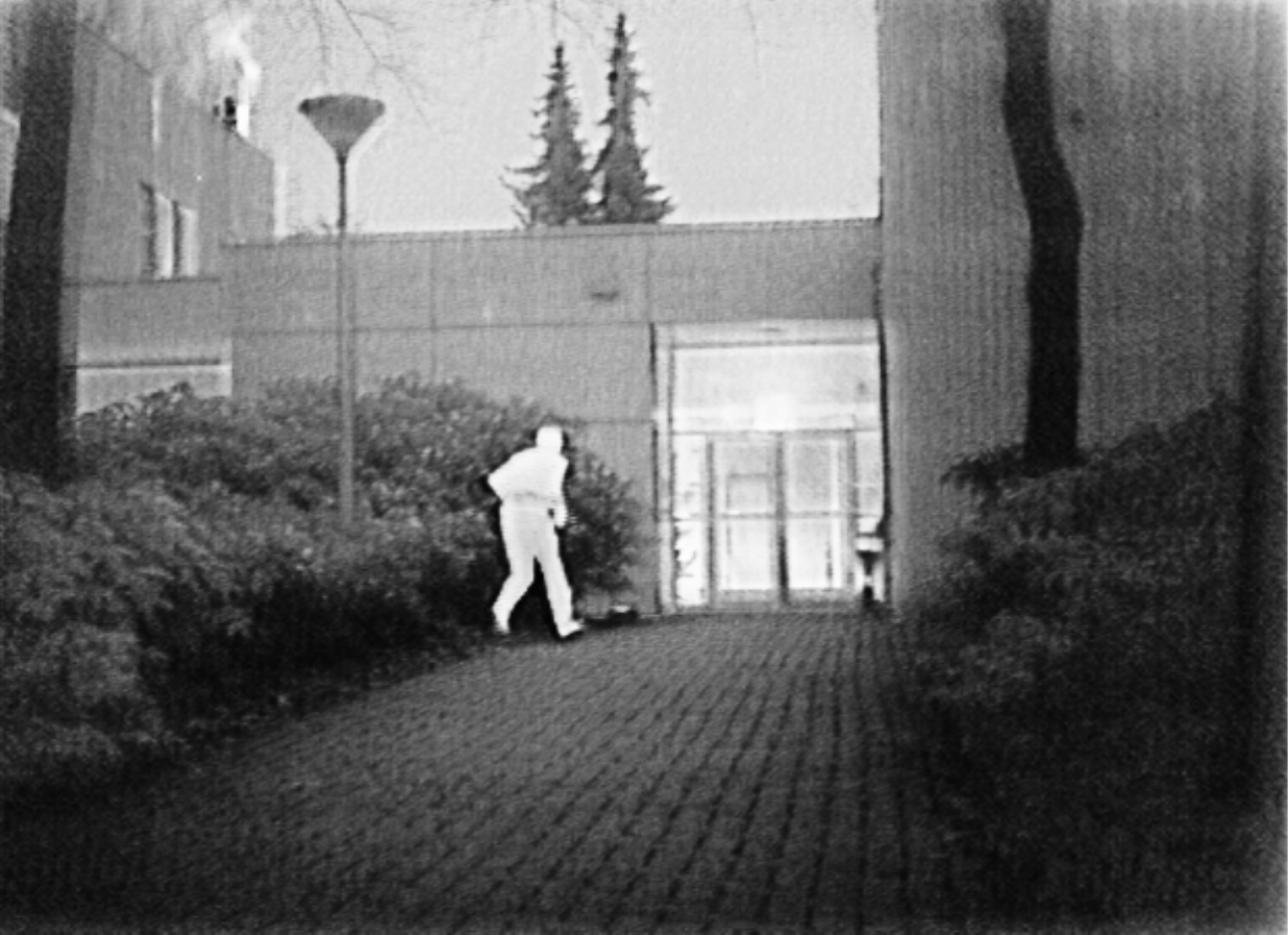}
		&\includegraphics[width=0.15\textwidth]{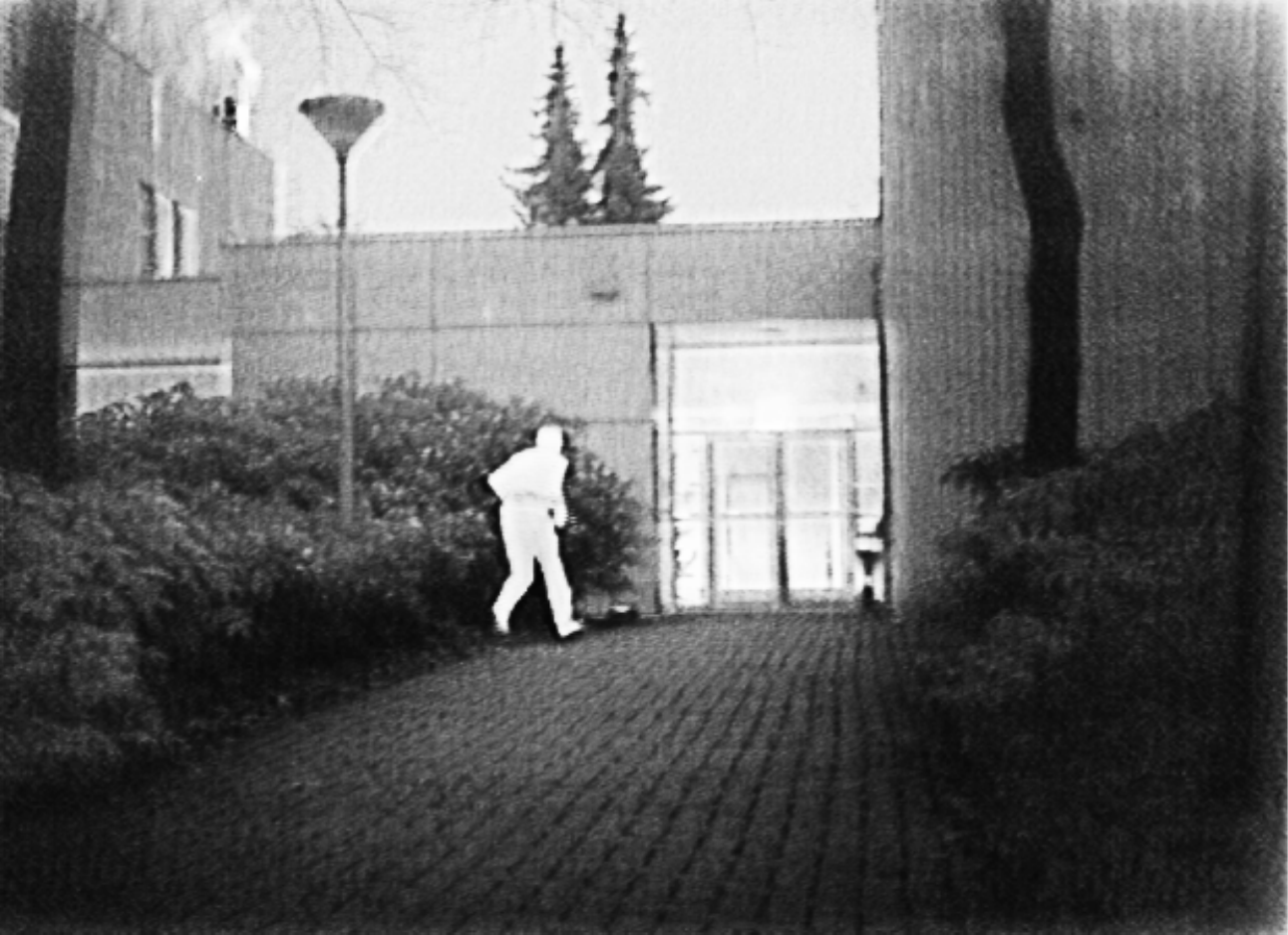}
		&\includegraphics[width=0.15\textwidth]{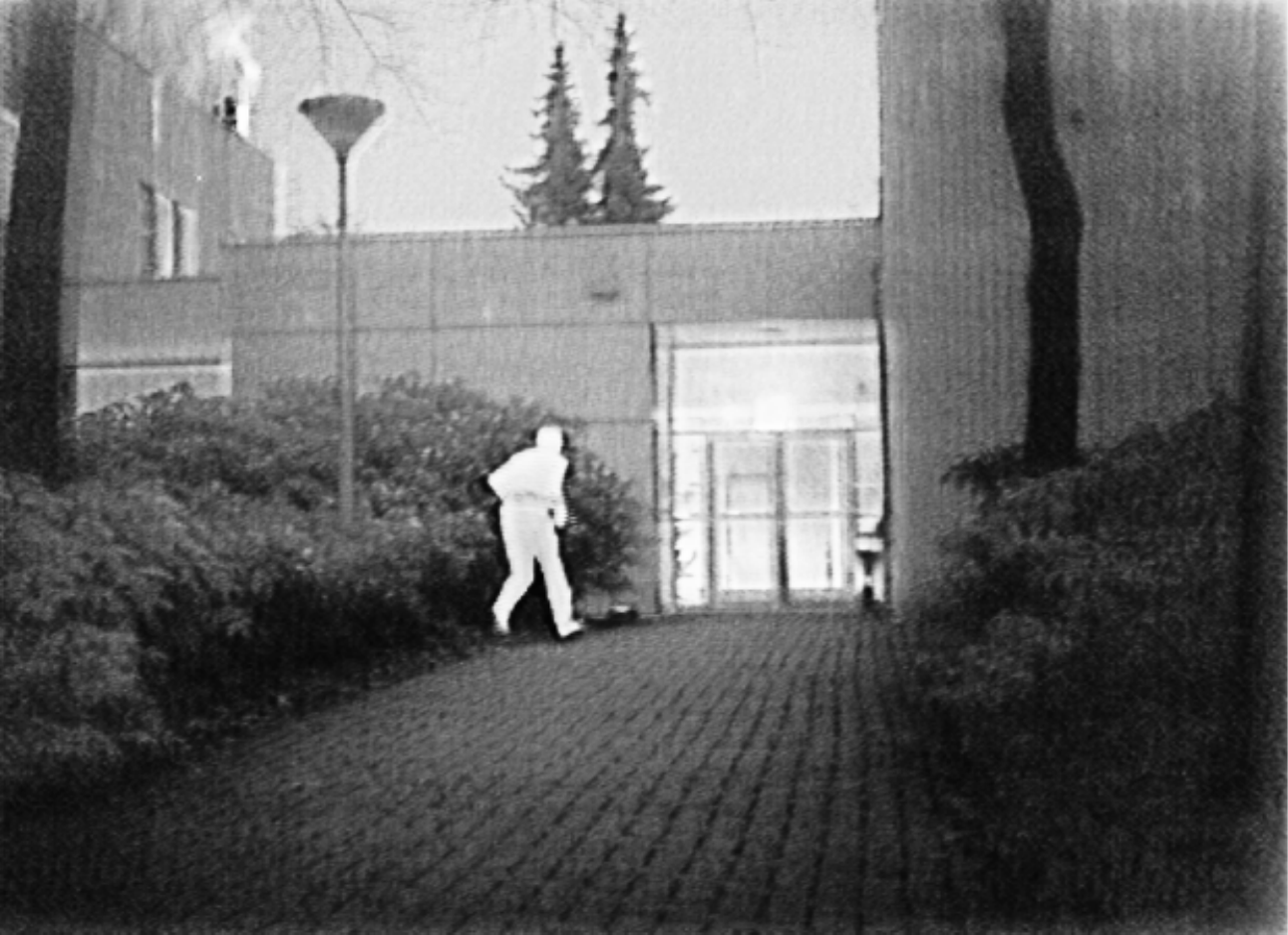}
		\\
		$\mathcal{TM}$-fused&$\mathcal{UM}$-fused&Ours
	\end{tabular}
	\caption{Different masks and their fusion results.}
	\label{fig:MsFusion}
\end{figure}

\begin{table*}
	\caption{Quantitative comparison on model complexity and running time.}
	\centering 
	\renewcommand\arraystretch{1.1} 
	\setlength{\tabcolsep}{0.5mm}
	\begin{tabular}{|c|ccccccccccccc|}
		\hline
		Method&\footnotesize DenseFuse&\footnotesize FusionGAN&\footnotesize PMGI&\footnotesize DIDFuse&\footnotesize GANMcC&\footnotesize RFN&\footnotesize MFEIF&\footnotesize U2Fusion&\footnotesize SwinFusion&\footnotesize SDNet&\footnotesize SMoA&\footnotesize TarDAL&\footnotesize Ours\\
		\hline
		\footnotesize TP(M)&\footnotesize 0.925&\footnotesize 0.074&\footnotesize \textcolor{red}{\textbf{0.042}}&\footnotesize 0.261&\footnotesize 1.864&\footnotesize 10.936&\footnotesize 0.158&\footnotesize 0.659&\footnotesize 0.974&\footnotesize \textcolor{blue}{\textbf{0.067}}&\footnotesize 0.223&\footnotesize 0.297&\footnotesize 9.13\\
		\hline 	
		\footnotesize FLOPS(G)&\footnotesize 497.96&\footnotesize 48.96&\footnotesize 745.21&\footnotesize \textcolor{red}{\textbf{18.71}}&\footnotesize 1002.56&\footnotesize 676.09&\footnotesize \textcolor{blue}{\textbf{25.32}}&\footnotesize 366.34&\footnotesize 471.04&\footnotesize 37.35&\footnotesize 61.869&\footnotesize 82.37&\footnotesize 115.37\\
		\hline 
		
		\footnotesize Time(s)&\footnotesize 0.124&\footnotesize 0.251&\footnotesize 0.182&\footnotesize 0.055&\footnotesize 0.246&\footnotesize 0.239&\footnotesize 0.045&\footnotesize 0.123&\footnotesize 1.345&\footnotesize \textcolor{blue}{\textbf{0.045}}&\footnotesize 8.071&\footnotesize \textcolor{red}{\textbf{0.002}}&\footnotesize 0.052\\
		\hline 
	\end{tabular}
	\label{tab: time}	
\end{table*}

\begin{figure*}
	\centering
	\setlength{\tabcolsep}{1pt} 
	
	\includegraphics[width=0.95\textwidth]{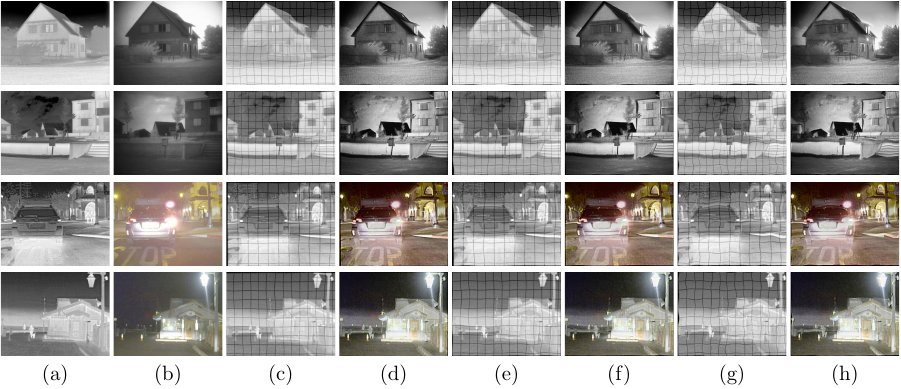}
	\caption{Visual results of our fusion method on unregistered image pairs
		on the TNO and Roadscene dataset, respectively. From right to left: visible image, infrared image, SMI, SMI result, MMI, MMI result, EMI, EMI result. SMI, MMI and EMI denotes tiny misaligned, moderate misaligned and extreme misaligned infrared image, respectively. Best viewed on screen.}
	\label{fig:regis}
\end{figure*}
\noindent\textbf{Analysis of Different masks}
We generated the $\mathcal{M}$ in various manners in order to investigate how the $\mathcal{M}$ affects the proposed method.
All three kinds of masks are shown in the first line in Figure~~\ref{fig:MsFusion}. $\mathcal{TM}$ is the feature weight map generated by a human detection algorithm~\cite{montabone2010human} guided by the saliency mechanism. In practice, we set a threshold to convert it since it was originally not a 0-1 mask. $\mathcal{UM}$ comes from an unsupervised saliency detector~\cite{nguyen2019deepusps}, which also marks the door highlighted by room temperature in the infrared image.

We used these three different masks for fusion, and the results are shown in the second line  in Figure~~\ref{fig:MsFusion}. From the visual perspective, different masks do not affect the visualization features of the proposed method (such as real details and high contrast). The performance of the proposed method is almost independent of how we obtain the masks. 

\begin{figure*}
	\centering
	\setlength{\tabcolsep}{1pt} 
	
	\includegraphics[width=0.95\textwidth]{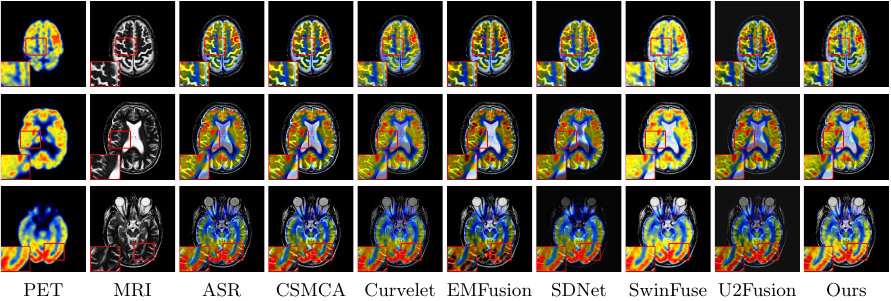}
	\caption{Visual comparison of different methods on MRI-PET task. Best viewed on screen.}
	\label{fig:MIFpet}
\end{figure*}

\subsection{Computational Complexity Analysis}
Beyond qualitative and numerical analysis, model size and running speed are significant in real-world applications as well. Therefore, we verify the memory consumption and computation efficiency of the proposed model. Note that the results shown in this part might be slightly different from the original paper due to different settings of running platform, hyper-parameters, etc. Specifically, FLOPs and training parameters are calculated by setting input size as 64$\times$64. For running time, we select ten images from TNO with size 64$\times$64 to compute the average time. Table \ref {tab: time} displays quantitative results on model size, FLOPs and running time among several state-of-the-art methods. Our model is faster than all methods except for SDNet and MFEIF. Although we are slower due to the simplicity of their frameworks, our model still achieves a running time under 0.1s, which is two times faster than DenseFuse and U2Fusion, three times faster than PMGI, and four times faster than FusionGAN, GANMcC and RFN. Moreover, in spite that our CoCoNet is more complicated than FusionGAN, we enjoy faster speed, which demonstrates the strength of the proposed architecture. Specially, MAM module needs large parameters to calculate and integrate the multi-level features for better feature extraction and transmission. Although removing the MAM module can alleviate the computational complexity to some extent, it may lose some vital information during the fusion process, resulting in loss of detail or blurred target on the fused result.

\subsection{Limitations}
It is pretty challenging to acquire pixel-level aligned infrared and visible image pairs in most real-world scenarios. Existing work~\cite{tang2022superfusion,huang2022reconet} alleviates this by introducing additional registration~\cite{ma2019locality,LiuLZFL20,LiuLFZHL22} module. To discuss the impact of fusing these slight misaligned image pairs, we initially synthesize these slight misaligned source images through performing random affine and elastic translations with different degrees(i.e., tiny, moderate, and extreme) on the TNO, Roadscene datasets, respectively. Then we use the proposed method for merging these misaligned source images. Visual results are shown in Figure~\ref{fig:regis}, note that our method can deal with the slightly misaligned images, which preserves a large part of vital information. However, halos and artifacts emerge on the fused results when pixel deviation is significant.

\begin{figure*}
	\centering
	\setlength{\tabcolsep}{1pt} 
	
	\includegraphics[width=0.95\textwidth]{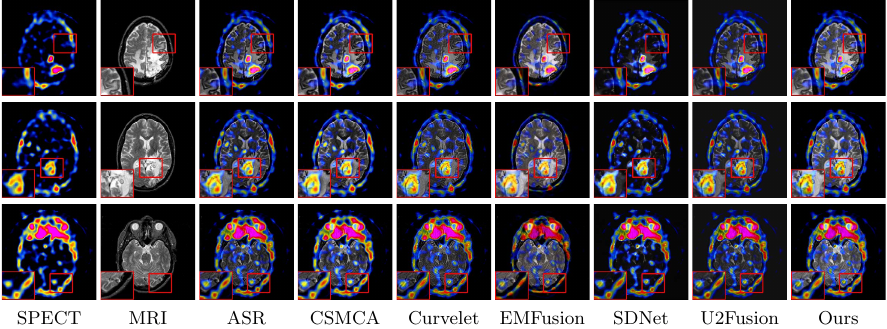}
	\caption{{Visual comparison of different methods on MRI-SPECT task. Best viewed on screen.}}
	\label{fig:MIFspect}
\end{figure*}

\subsection{Results and Analysis on MIF}
We compare the proposed method with some state-of-the-art approaches, including three traditional methods (i.e., self-adaptive sparse representation based ASR~\cite{liu2015simultaneous}, convolution sparse representation and morphological principal component analysis based CSMCA~\cite{liu2019medical} and contour wave transform based Curvelet~\cite{Yang2008Multimodality}), three CNN-based method (i.e., EMFusion~\cite{xu2021emfusion}, SDNet~\cite{zhang2021sdnet} and U2Fusion~\cite{U2Fusion2020}) and transformer based SwinFuse~\cite{Ma2022SwinFusion}. As SwinFusion is only able to deal with the MRI-PET image pairs, so that its results  only participate in the comparison of MRI-PET fusion task.

\begin{table*}[htbp]
	\caption{Quantitative comparison of MRI-PET and MRI-SPECT image fusion tasks.}
	\centering
	\resizebox{\textwidth}{!}{
		\begin{tabular}{@{}cccccccc|ccccccc@{}}
			\toprule
			&                    & \multicolumn{6}{c|}{\textbf{MRI-PET}}                                         & \multicolumn{6}{c}{\textbf{MRI-SPECT}}                                    &  \\ \midrule
			& Method                   & \textbf{EN} & \textbf{SD} & \textbf{VIF}   & \textbf{SF}   & \textbf{AG} & \textbf{SCD}& \textbf{EN} & \textbf{SD} & \textbf{VIF}   & \textbf{SF}   & \textbf{AG} & \textbf{SCD} &  \\ \midrule
			& \textbf{ASR}
			&3.755$\pm$0.491	&8.658$\pm$0.768	&0.573$\pm$0.056	&0.1027$\pm$0.0164	&6.570$\pm$1.306	&1.246$\pm$0.158
			&3.922$\pm$0.291    &8.445$\pm$0.336 &0.452$\pm$0.017	&0.0714$\pm$0.0033 &\textcolor{blue}{4.891$\pm$0.339} &\textcolor{blue}{1.060$\pm$0.031}&
			\\
			& \textbf{CSMCA}
			& 3.937$\pm$0.524	&8.649$\pm$0.743	&0.645$\pm$0.066	&\textcolor{blue}{0.1071$\pm$0.0173}	&\textcolor{blue}{6.919$\pm$1.392}	&1.318$\pm$0.170
			&\textcolor{blue}{4.175$\pm$0.315}	&8.431$\pm$0.335	&\textcolor{blue}{0.472$\pm$0.018}	&\textcolor{red}{0.0782$\pm$0.0043}	&\textcolor{red}{5.454$\pm$0.427}	&\textcolor{red}{1.130$\pm$0.031}
			&     \\
			& \textbf{Curvelet}
			& 4.206$\pm$0.579	&8.599$\pm$0.777	&0.551$\pm$0.052	&0.0971$\pm$0.0151	&6.204$\pm$1.279	&1.152$\pm$0.170
			&\textcolor{red}{4.237$\pm$0.280}	&\textcolor{blue}{8.644$\pm$0.365}	&\textcolor{red}{0.502$\pm$0.019}	&\textcolor{blue}{0.0740$\pm$0.0046}	&4.885$\pm$0.405	&1.032$\pm$0.028
			&     \\
			& \textbf{EMFusion}
			&4.165$\pm$0.584	&8.516$\pm$0.594	&\textcolor{blue}{0.667$\pm$0.056}	&0.1009$\pm$0.0174	&6.662$\pm$1.412	&1.214$\pm$0.201
			&3.928$\pm$0.260	&8.378$\pm$0.485	 &0.23$\pm$0.011	    &0.0637$\pm$0.0029	&4.374$\pm$0.315	&0.585$\pm$0.038
			&     \\
			& \textbf{SDNet}
			&4.150$\pm$0.551	    &8.089$\pm$0.824	&0.424$\pm$0.046	&0.0626$\pm$0.0077	&3.840$\pm$0.588	&0.824$\pm$0.093
			&4.084$\pm$0.294	&6.639$\pm$0.237	&0.439$\pm$0.012	&0.0621$\pm$0.0054	&3.305$\pm$0.329	&1.004$\pm$0.022
			&     \\
			& \textbf{U2Fusion}
			&4.088$\pm$0.588	&7.928$\pm$0.693	&0.444$\pm$0.028	&0.0632$\pm$0.0076	&4.627$\pm$0.851	&0.616$\pm$0.261
			&3.972$\pm$0.272	&7.279$\pm$0.214	&0.256$\pm$0.011	&0.0543$\pm$0.0033	
			&3.800$\pm$0.323	&0.843$\pm$0.036
			&     \\
			& \textbf{SwinFuse}
			&\textcolor{blue}{4.468$\pm$0.664}	&\textcolor{blue}{8.697$\pm$0.758}	&\textcolor{red}{0.722$\pm$0.074}	&0.0772$\pm$0.0156	&5.902$\pm$1.454	&\textcolor{red}{1.726$\pm$0.078}
			&-&-&-&-&-&-
			&     \\
			& \textbf{Ours}
			&\textcolor{red}{4.498$\pm$0.359}	&\textcolor{red}{8.737$\pm$0.694}	&0.645$\pm$0.057	&\textcolor{red}{0.1081$\pm$0.0166}	&\textcolor{red}{7.049$\pm$1.387}	&\textcolor{blue}{1.454$\pm$0.240}
			&4.142$\pm$0.221	&\textcolor{red}{8.871$\pm$0.452}	&0.404$\pm$0.022	&0.0653$\pm$0.0029	&4.658$\pm$0.317	&0.874$\pm$0.028
			&     \\
			\bottomrule
		\end{tabular}
	}
	\label{T_MIF}
\end{table*}

\subsubsection{MRI-PET fusion task}
\noindent\textbf{Qualitative Comparisons}
The comparisons of visualization on typical MRI-PET images are shown in Figure~\ref{fig:MIFpet}. Benefiting from the multi-scale attention mechanism, the proposed method can accurately retain high-value information in different modalities. In addition, the contrastive learning using MRI segmentation masks, which successfully solves the problem that MRI image information is covered by intensity information of other modalities, prevents the loss of effective details. The three traditional methods have similar performance. They focus on the preservation of details in MRI, but ignore the restoration of color information in another modality (the frame part of the first two sets of images). SDNet and U2Fusion are difficult to effectively extract and save information, causing serious loss of details and colors, while EMFusion causes serious color distortion (the frame part of the last set of images). SwinFusion retains color information well, but it does not solve the problem of MRI details being covered. To sum up, the proposed method achieves the balance between retaining color and detail information, and achieves the optimal visualization effect.

\noindent\textbf{Quantitative Comparisons}
To give a better view of the performance of CoCoNet on MIF task, we provide the evaluations of all methods on metrics discussed in the above section. The results are shown in the left of Table~\ref{T_MIF} . It is obvious that the proposed method achieves the highest results on EN, SD, SF, AG and SOTA level on SCD. SwinFusion achieves the highest VIF and SCD results with its characteristic highlighted visualization.

\subsubsection{MRI-SPECT fusion task}
\noindent\textbf{Qualitative Comparisons}
The comparisons of visualization on typical MRI-SPECT images are shown in Figure~\ref{fig:MIFspect}. Different from PET, intensity information of SPECT image is more sparse. The MRI segmentation mask contrastive learning used can better preserve the details of the MRI image with less interference.
Other deep learning methods suffer information loss in different degrees and modalities: SDNet and U2Fusion enable to save the details of MRI images, while EMFusion has a serious defect in color information (the frame part of the first set and the last set of images). Traditional methods retain intensity information, but suffer from MRI partial darkening (the frame part of the first two sets of images). 

\noindent\textbf{Quantitative Comparisons}
The evaluations of all methods on metrics are shown in the right side of Table~~\ref{T_MIF}. All deep learning methods have disastrous performance on six metrics, but we have achieved SOTA performance upon all depth learning methods. Based on the actual analysis, the reasons are as follows: since the original resolution of SPECT images is low, the images need to be upsampled before participating in fusion. In this process, many non-uniform pixel noises are introduced,while convolutional neural network can better process them to produce smooth results. In the calculation of indicators, the noise interference produces results unfavorable to the depth learning method, such as EN, SF and AG. Our method better retains the information of the two modalities, so it is superior to other deep learning methods in terms of metrics.

\begin{figure*}
	\centering
	
	\setlength{\tabcolsep}{1pt} 
	
	\includegraphics[width=0.99\textwidth]{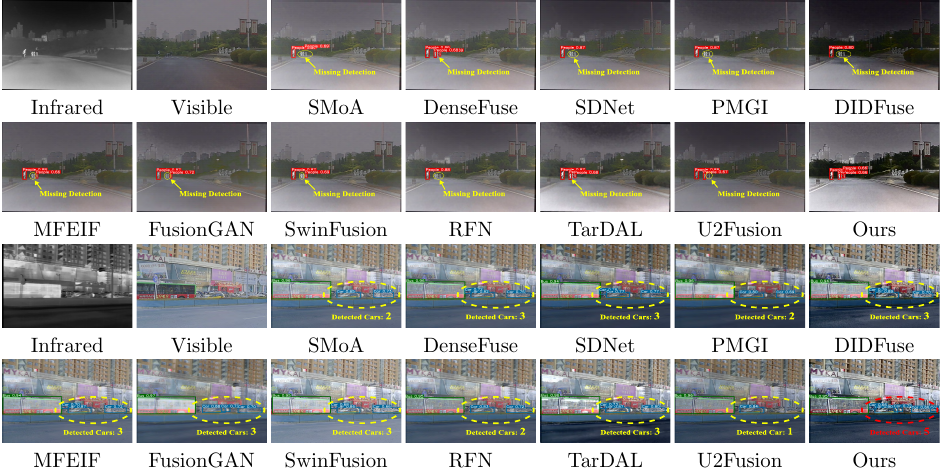}
	\caption{Visual comparison of our CoCoNet with state-of-the-art approaches on the M$^3$FD dataset.}
	\label{fig:detection}
\end{figure*}

\subsection{Task-driven evaluation}
The utilization of fused images encompasses not only the enhancement of fundamental visual observations but also plays a pivotal role in boosting the performance of their subsequent high-level vision tasks. Unfortunately, the majority of existing evaluation methods primarily focus on the assessment of visual quality and relevant statistical indicates pertaining to these images. In this section, we advocate for a task-driven evaluation that surpasses the limitations inherent in traditional evaluation paradigms. To achieve this objective, we employ a two-fold approach involving object detection and semantic segmentation tasks applied to the fused images. This methodology enables the comparative analysis of various infrared and visible image fusion techniques concerning their subsequent high-level vision tasks.

\begin{table*}[!htb]
	\caption{Quantitative results of object detection on the M$^3$FD datasets with retrained detectors (YOLOv5) for each method. The best result is in {\textcolor{red}{\textbf{red}}} whereas the second best one is in {\textcolor{blue}{\textbf{blue}}}.}
	\centering
	\renewcommand\arraystretch{1.1} 
	\setlength{\tabcolsep}{2.8mm}
	{
		\begin{tabular}{|c|cccccc|c|c|}
			\hline
			\multirow{2}{*}{ Method}&\multicolumn{8}{c|}{ M$^3$FD dataset}\\
			\hhline{~|*{8}{|-}} 
			~& People& Car& Bus& Motor& Truck & Lamp & mAP@.5& mAP@.5:.95\\
			\hline
			SMoA& 65.9& \color{blue}88.5& 95.2& 69.5& 71.7& \color{blue}90.7& 80.3& 53.5\\
			\hline
			DenseFuse&66.1 &88.3&95.2&69.6&73.0&\color{blue}90.7&80.5&53.9   \\
			\hline
			SDNet& \color{red}67.4& \color{blue}88.5& 95.2& 69.4& 68.9& 89.6& 79.8& 53.8\\
			\hline
			PMGI& \color{blue}67.2& \color{red}88.8& 95.2& 67.5& 71.5& 88.3& 79.9& 53.7\\
			\hline
			DIDFuse& 66.6& 88.1& \color{red}96.8& 69.8& 72.9& 89.7& \color{blue}80.8& 54.1\\
			\hline
			GANMcC& 65.6& 88.3& 96.1& 70.4& 73.3& 88.8& 80.4& 53.6\\
			\hline
			MFEIF& 65.1& 88.1& 95.4& 68.9& \color{red}74.7& 89.8& 80.3& 53.8 \\
			\hline
			FusionGAN& 64.2& 81.4& 88.8& 69.2& 65.4& 89.5& 76.4& 51.2\\
			\hline
			SwinFusion& 65.7& 87.9& \color{blue}96.3& 68.5& \color{blue}74.0& 89.2& 80.3& 53.6\\
			\hline
			RFN&66.6& 88.1& 95.8& 67.9& 71.3& \color{red}91.2& 80.1& 53.9\\
			\hline
			TarDAL&65.7& 88.4& 96.1& \color{blue}71.2& 73.2& \color{blue}90.7& \color{red}80.9& \color{red}54.5\\
			\hline
			U2Fusion&65.7& 88.3& 95.2& 68.5& 73.9& 88.8& 80.1& 54.1\\
			\hline
			Ours&65.1& \color{blue}88.5& 94.6& \color{red}71.8& 73.8& 90.6& 80.7& \color{blue}54.2 \\
			\hline 
		\end{tabular}
	}
	
	\label{tab:M3FD_detect}
\end{table*}

\subsubsection{Evaluation on object detection}
Object detection is a traditional and extensively studied high-level computer vision task. With the continuous improvement of multi-modality datasets, its ability to reflect semantic information in the evaluation of multi-modality image fusion techniques has become increasingly prominent. In this subsection, we focus on discussing the impact of image fusion on object detection.

\noindent\textbf{Implementation details}~We conducted experiments on the latest M$^3$FD dataset using the state-of-the-art detector YOLOv5\footnote{https://github.com/ultralytics/yolov5}. To ensure a fair comparison, we used the YOLOv5s model and directly input the fusion results of various methods into the detector for retraining. We then tested the retrained models on the same randomly partitioned test set. All settings of the detector follow its original configuration, and the quantitative results are directly output by its test code.

\noindent\textbf{Quantitative Comparisons}~The quantitative results obtained from the experiments are shown in Table~\ref{tab:M3FD_detect}. mAP@.5 represents the mAP (mean average precision) value when mIoU is 0.5, mAP@.5:.95 represents the average of all mAP values at different IoU thresholds (from 0.5 to 0.95 in steps of 0.05), and other values represent the AP (average precision) of the corresponding classes.

We conducted a horizontal comparison with the same benchmark methods as in the previous parts. In terms of the most concerned metric mAP@.5, the proposed method did not achieve the top two results but still performed favorably (ranked third). TarDAL, optimized for object detection, achieved the best result, while DIDFuse, capable of generating well-fused images, ranked second. Regarding the more comprehensive metric mAP@.5:.95, which reflects the performance across different IoU, CoCoNet ranked second, demonstrating its detection superiority among common fusion methods. In addition, different methods showed preferential performance for different labels in their respective class APs.

\noindent\textbf{Qualitative Comparisons}~
To illustrate the advantages of our proposed method in facilitating downstream detection tasks, we provide two visual examples in Figure~\ref{fig:detection}, both of which highlight the detection results with confidence greater than 0.6. Scene 1 depicts a pedestrian detection scenario in overcast weather. The fusion results of the proposed method highlight the pedestrians, creating a high-contrast visual effect that fits well with the detection network, achieving the best detection performance. Conversely, methods such as FusionGAN and U2Fusion produce blurred outlines of the person, resulting in low detection confidence. Scene 2 shows a driving detection scenario, which better reflects the degree to which fusion methods preserve and utilize visible information. Our fusion result retain rich visible information while achieving the best detection performance, which meets the requirements of this special scenario.

\noindent\textbf{Ablation of Contrastive Learning}~
To further investigate the impact of the adopted contrastive learning on the object detection task, we also conducted corresponding object detection experiments on three ablation variants (w/o CL: no contrastive regularization, w/ $\mathcal{L}_{ir}$: target constrain only, w/ $\mathcal{L}_{vis}$: detail constrain only), and the quantitative results are shown in Table~\ref{tab:M3FD_detect_cl}. 

In terms of mAP, the proposed method has a clear advantage, followed closely by the variant without contrastive learning. Meanwhile, adding constraints for object or detail separately had a negative impact on the overall detection results. In addition, it is worth noting that the variant with the object constraint indeed exhibits stronger sensitivity to salient objects in the scene, resulting in excellent APs for categories such as person and bus. However, the variant with only the detail constraint cannot adapt to subsequent detection tasks. The complete method with target-detail coupling achieves excellent integration and utilization of the two constraints.

\begin{table}[!htb]
	\caption{Quantitative results of object detection on the M$^3$FD datasets with retrained detectors (YOLOv5) for each ablation variant of contrastive learning. The best result is in {\textcolor{red}{\textbf{red}}} whereas the second best one is in {\textcolor{blue}{\textbf{blue}}}.}
	\centering
	\renewcommand\arraystretch{1.1} 
	\setlength{\tabcolsep}{2.3mm}
	{
			\begin{tabular}{|c|cccc|}
		\hline
		Class & w/o CL & w/ $\mathcal{L}_{ir}$ & w/ $\mathcal{L}_{vis}$ & Ours   \\ \hline
		People &\color{blue}65.2 & \color{red}65.8& 63.0& 65.1    \\
		Car & \color{blue}88.3 & \color{blue}88.3& 87.5& \color{red}88.5    \\
		Bus & 94.1& \color{red}94.8& 94.3& \color{blue}94.6  	\\
		Motor & 69.1& \color{blue}70.0& 69.7& \color{red}71.8\\
		Truck & \color{blue}71.7& 67.1& 70.3& \color{red}73.8\\
		Lamp &88.7& 89.2& \color{red}91.5& \color{blue}90.6 \\ \hline
		mAP@.5 & \color{blue}79.5 & 79.2& 79.4& \color{red}80.7 \\ \hline
		mAP@.5:.95 & \color{blue}53.7 & 53.2& 53.5& \color{red}54.2 \\ \hline
	\end{tabular}
	}
	
	\label{tab:M3FD_detect_cl}
\end{table}

\begin{figure*}
	\centering
	\setlength{\tabcolsep}{1pt} 
	
	\includegraphics[width=0.98\textwidth]{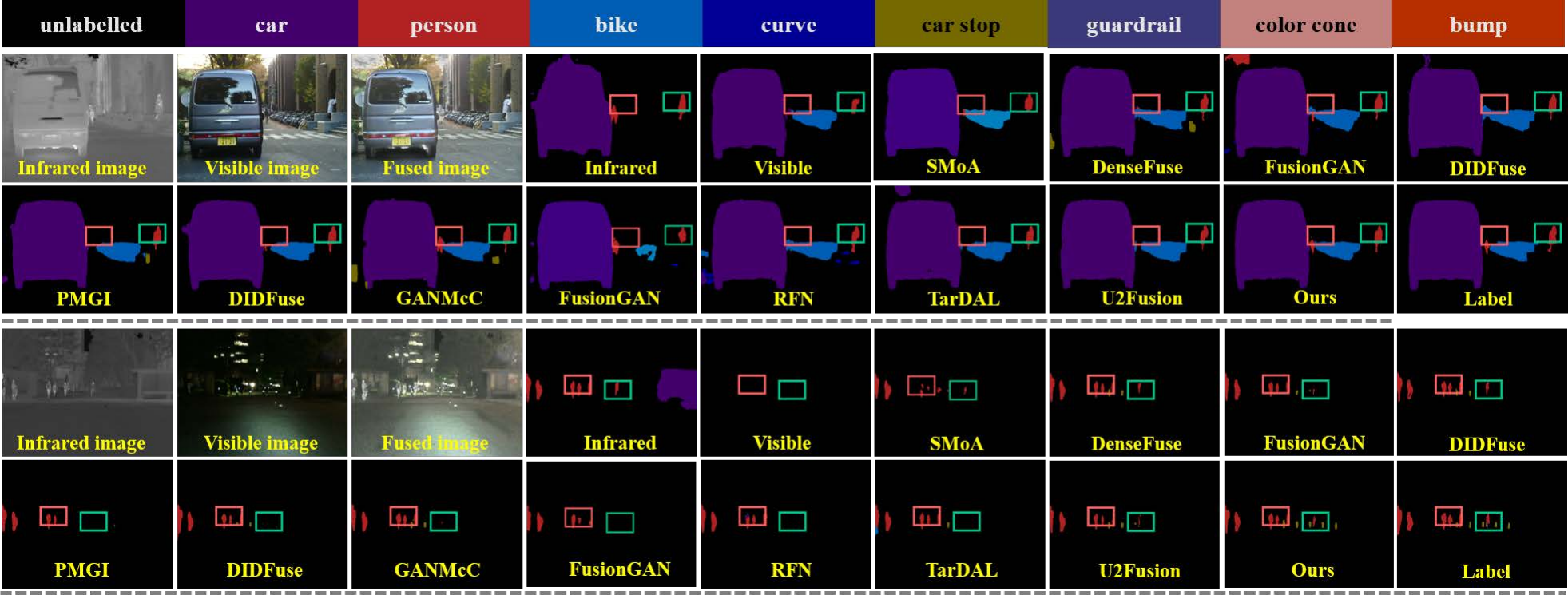}
	
	\caption{Visual comparison of our CoCoNet with state-of-the-art approaches on the {MFNet} dataset.}
	\label{fig:segmentation} 
\end{figure*}
\begin{table*}[thb]
	\caption{Quantitative semantic segmentation results of different methods on the {MFNet} dataset.}
	~\label{tab:MFNet_seg}
	\centering
	\scriptsize
	\renewcommand{\arraystretch}{1.1}
	\setlength{\tabcolsep}{0.8mm}{
		\begin{tabular}{|c|cc|cc|cc|cc|cc|cc|cc|cc|cc|}
			\hline
			\multirow{2}{*}{Methods} &			\multicolumn{2}{c}{Unlabel}&  \multicolumn{2}{c}{Car} & \multicolumn{2}{c}{Person} & \multicolumn{2}{c}{Bike}& \multicolumn{2}{c}{Curve}  & \multicolumn{2}{c}{Car Stop}&
			\multicolumn{2}{c}{Cone}& \multicolumn{2}{c|}{Bump}
			&\multirow{2}{*}{mAcc}&\multirow{2}{*}{mIoU} \\ \hhline{~|*{16}{|-}} 
			&Acc &IoU & Acc &IoU& Acc &IoU& Acc &IoU& Acc &IoU& Acc &IoU& Acc &IoU& Acc &IoU& &\multicolumn{1}{c|}{} \\
			\hline
			SMoA & 98.1 & 97.4 & 93.0 & 81.0 & 85.3 & 63.7 & 76.7 & 58.1 & 53.6 & 28.1 & 69.5 & 18.1 & 85.1 & 39.2 &\color{blue} 85.8 & 29.6 & 73.2 & 46.7 \\
			DenseFuse &\color{blue} 98.2 &\color{blue} 97.7 & 94.1 &\color{blue} 82.4 & 87.3 & 66.0 &\color{red} 86.6 & 57.0 & 65.3 &\color{red} 39.2 & 53.3 & 12.2 & 87.9 & 38.5 & 79.6 & 27.0 & 73.0 & 47.2 \\
			SDNet & 98.1 & 97.6 & 93.6 & 81.4 &\color{blue} 88.7 & 65.4 & 83.1 & 58.3 & 63.3 & 29.2 & 56.3 & 12.3 & 88.5 &\color{blue} 41.5 &\color{red} 89.3 & 42.1 & 73.5 & 47.6 \\
			PMGI &\color{blue} 98.2 & 97.4 & 91.9 & 80.6 & 86.1 & 65.8 & 80.0 & 56.4 & 56.8 & 30.9 & 64.9 & 10.0 &\color{red} 92.5 & 32.6 & 68.5 & 43.1 & 71.9 & 47.0 \\
			DIDFuse & 98.1 & 96.9 & 89.1 & 78.0 & 82.9 & 59.6 & 70.5 & 53.4 & 34.2 & 20.6 & 49.0 &\color{blue} 25.1 & 82.4 & 34.7 & 59.5 & 20.1 & 63.1 & 43.3 \\
			GANMcC & 98.0 & 97.3 & 91.0 & 79.4 & 86.0 & 60.8 & 79.4 & 54.6 & 62.9 & 25.6 & 66.0 & 18.0 & 66.7 & 38.7 & 75.6 & 37.5 & 69.5 & 45.8 \\
			MFEIF & 98.0 & 97.5 & 94.9 & 80.6 & 87.9 & 63.2 & 81.6 & 57.2 & 63.5 & 30.9 & 60.1 & 19.0 & 88.7 & 35.9 & 66.3 & 37.5 & 71.2 & 46.9 \\
			FusionGAN & 98.0 & 97.2 & 87.4 & 76.5 & 86.6 & 63.9 & 82.1 & 52.7 & 67.2 & 32.1 & 64.3 & 14.0 & 80.9 & 37.0 & 83.6 & 38.7 & 72.3 & 45.8 \\
			SwinFusion & 97.9 & 97.6 &\color{red} 96.6 & 80.0 &\color{red} 89.1 &\color{blue} 66.2 & 83.7 & 57.0 &\color{blue} 70.2 & 36.5 &\color{blue} 74.2 & 17.8 & 88.8 & 37.3 & 62.3 & 11.4 & 74.1 & 45.2 \\
			RFN &\color{blue} 98.2 & 97.3 & 91.4 & 77.5 & 86.5 & 63.9 & 71.0 & 51.1 & 61.0 & 38.4 & 51.5 & 17.4 & 76.6 & 36.8 & 63.9 &\color{blue} 47.2 & 67.3 &\color{blue} 48.3 \\
			TarDAL &\color{blue} 98.2 & 97.6 & 92.4 & 80.8 & 87.9 & 65.5 & 79.0 &\color{blue} 60.1 & 59.0 & 29.8 & 69.2 &\color{red} 26.8 & 89.3 & 36.9 & 72.7 & 32.5 & 72.2 & 48.0 \\
			U2Fusion & 98.0 & 97.6 &\color{blue} 95.7 & 81.6 & 85.5 & 64.3 & 79.1 &\color{red} 61.0 &\color{red} 78.8 & 12.8 & 70.6 & 17.2 & 84.0 &\color{red} 42.5 & 77.3 &\color{red} 47.9 &\color{blue} 74.8 & 47.5 \\
			Ours &\color{red} 98.3 &\color{red} 97.8 & 95.3 &\color{red} 83.6 & 87.5 &\color{red} 68.4 &\color{blue} 84.1 & 59.7 & 67.3 &\color{blue} 38.6 &\color{red} 75.0 & 23.5 &\color{blue} 91.8 & 38.1 & 84.3 & 25.6 &\color{red} 76.1 &\color{red} 48.5 \\
			
			\hline
			
	\end{tabular} }
	
\end{table*}
\subsubsection{Evaluation on semantic segmentation}
During the evaluation of image fusion techniques, semantic segmentation is able to more accurately assess its ability to reflect different semantic categories. Its pixel-level classification method places greater emphasis on the richness and accuracy of semantic information. In this section, we focus on discussing the impact of image fusion on semantic segmentation.

\noindent\textbf{Implementation details}~ We conducted experiments on the MFNet dataset using the state-of-the-art semantic segmentation model SegFormer~\cite{xie2021segformer}. For fair comparison, we loaded pre-trained weight~\emph{mb1} and fine-tuned it for the same iterations on all fusion results. The training/validation/testing set split for MFNet followed the conventions of the source dataset.

\noindent\textbf{Quantitative Comparisons}~ We used pixel intersection-over-union (IoU) and accuracy (Acc) to reflect the segmentation expression. Table~\ref{tab:MFNet_seg} reports the specific evaluation results. As can be seen, the proposed method achieved the highest IoU in major object categories, i.e. Car and Person, and ranked first in mIoU \& mAcc. We attribute this advantage to two points. On the one hand, our fusion network removes redundant information between different modalities under the contrastive constraint, retains useful complementary information that supports a better understanding of the overall scene. On the other hand, the proposed MAM effectively integrates semantic features from the high-level network into the fusion process
, which makes our fused images contain rich semantic information.

\noindent\textbf{Qualitative Comparisons}~ We provide visualized segmentation results for daytime and nighttime scenes in Figure~\ref{fig:segmentation}. Visible images can better describe prominent large targets during the day, but overlook distant pedestrians with similar background colors. In contrast, infrared images effectively distinguish pedestrians at night and support excellent result. The segmentation model generates more accurate results on the fused images produced by the proposed method, for both main targets and background objects, such as the pedestrian within the red box in the daytime scene, and the car stops within the green box in the nighttime scene.

\noindent\textbf{Ablation of Contrastive Learning}~Similarly, we also conducted semantic segmentation experiments on three variants (w/o CL, w/ $\mathcal{L}_{ir}$, and w/ $\mathcal{L}_{vis}$), and the quantitative results of IoU are shown in Table~\ref{tab:MFNet_seg_cl}. 

It can be seen that the overall performance of the variants without contrastive constraint is poor. Among the two variants with single-sided constraint, the target of w/ $\mathcal{L}_{ir}$ improves the overall segmentation quality more than the detail of w/ $\mathcal{L}_{vis}$, because it is more beneficial for the extraction, learning, and approximation of infrared salient content when dealing with more nighttime scenes in the used dataset. The proposed complete contrastive constraint achieves the best performance on multiple categories and two average indicators, demonstrating the outstanding performance of using both constraint in parallel.

\begin{table}[!htb]
	\caption{Quantitative results of semantic segmentation on the MFNet datasets with retrained Segformer for each ablation variant of contrastive learning. The best result is in {\textcolor{red}{\textbf{red}}} whereas the second best one is in {\textcolor{blue}{\textbf{blue}}}.}
	\centering
	\renewcommand\arraystretch{1.1} 
	\setlength{\tabcolsep}{2.5mm}
	{
		\begin{tabular}{|c|cccc|}
			\hline
			Class & w/o CL & w/ $\mathcal{L}_{ir}$ & w/ $\mathcal{L}_{vis}$ & Ours   \\ \hline
			Unlabeled &97.1&\color{blue}97.6&97.3&\color{red}97.8    \\
			Car &76.8&\color{blue}80.9&78.6&\color{red}83.6    \\
			Person &57.2&\color{blue}65.7&62.3&\color{red}68.4	\\
			Bike &57.5&\color{blue}59.5&57.4&\color{red}59.7\\
			Curve&16.6&\color{blue}25.3&16.3&\color{red}38.6\\
			Car Stop &22.3&18.8&\color{red}30.9&\color{blue}23.5 \\ 
			Color Cone &37.2&\color{blue}37.7&37.6&\color{red}38.1 \\ 
			Bump &\color{red}35.4&\color{blue}32.4&6.5&25.6\\ \hline
			mIoU&44.6&\color{blue}46.8&43.2&\color{red}48.5 \\ \hline
			mAcc&\color{blue}73.5&70.4&65.7&\color{red}76.1\\ \hline
		\end{tabular}
	}
	\label{tab:MFNet_seg_cl}
\end{table}

\subsection{Extension to other fusion tasks}
To illustrate the broad applicability of our method, we expand the CoCoNet to tackle various issues related to other multi-modality image fusion. For instance, we apply it for the fusion of green fluorescent protein (GFP) and phase contrast (PC) images, as well as near infrared (NIR) and visible (VIS) images. 

In the fusion of GFP and PC images, GFP represents intensity and color information, similar to the role of PET/SPECT images, while PC represents structure and detail information, similar to the role of MRI images. Similarly, NIR images provide rich background information, while VIS images offer clear foreground content. Therefore, the proposed method is able to be competent for these tasks.

\begin{figure}
	\centering
	\setlength{\tabcolsep}{1pt}
	\begin{tabular}{c}
		\includegraphics[width=0.48\textwidth]{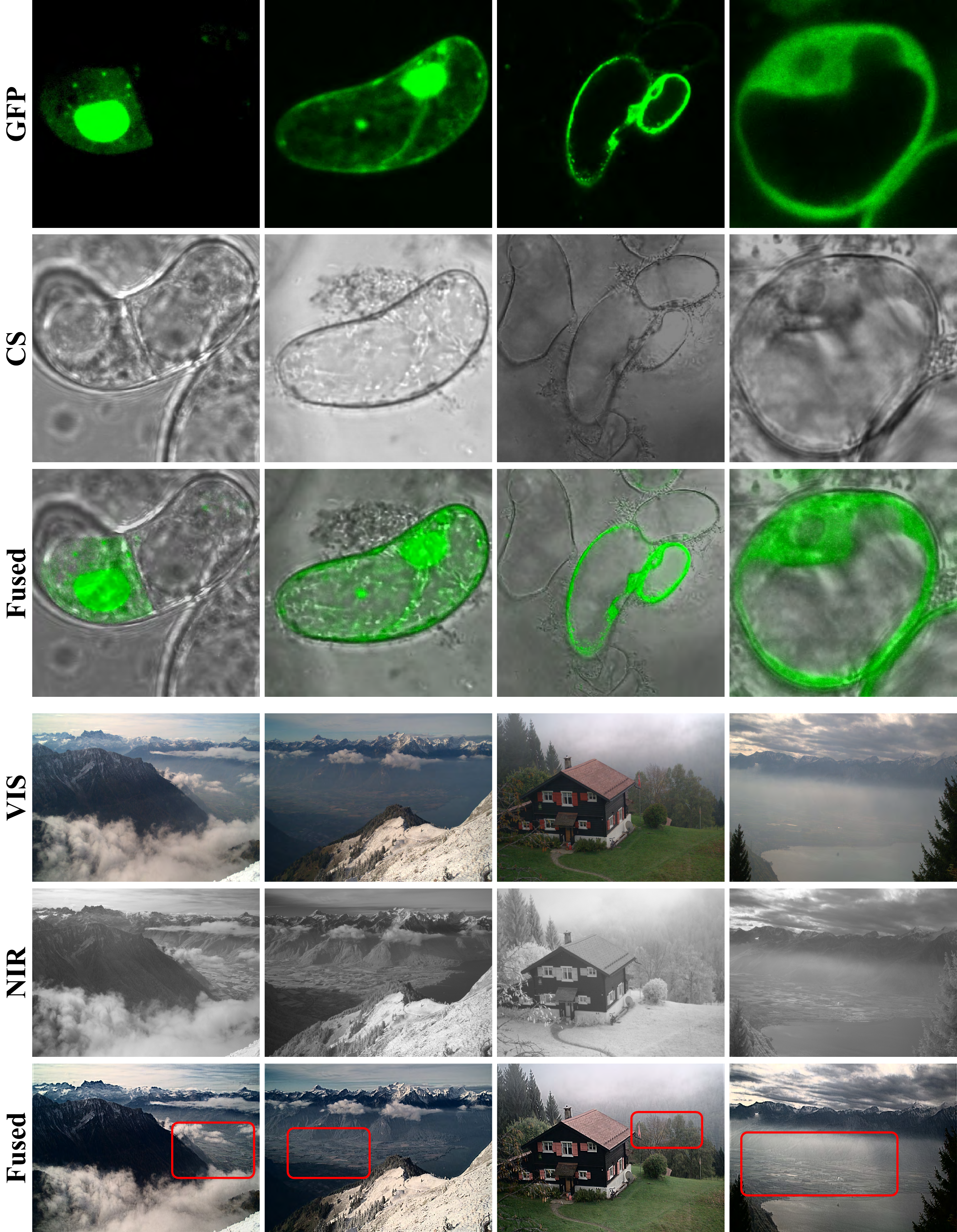}
		\\
	\end{tabular}
	\caption{Visual display of extension to other fusion tasks.}
	\label{fig:extension}
\end{figure}

The qualitative results are shown in Figure~\ref{fig:extension}, which demonstrate that the proposed CoCoNet generates fused images with promising visual results. Specifically, CoCoNet preserve the edge information in the phase contrast images while introducing the color information of green fluorescent protein with low deterioration. In the second task, CoCoNet achieve high contrast while retaining the rich texture information in the NIR images (see the areas in red boxes for details).

\noindent\textbf{Further Discussion on Generality} It is worth noting that we tested all the aforementioned tasks directly on the pre-trained module without fine-tuning, as they are not yet applicable to the proposed coupled contrastive learning strategy. 

As explained above, GFP and PC image fusion tends to be similar to MRI and PET image fusion, but PC images lack the information content that can provide MRI segmentation masks, as they mostly capture gradient details on lines instead of slightly larger areas of pixels. Concurrently, NIR and VIS images are natural images acquired from different but close bands, with approximate overall scene features and texture information. The small magnitudes of their modal differences pose challenges for designing an appropriate mask to conform to the constraints of the coupled contrastive learning.

We believe that in order to properly transfer and apply CCL to other image fusion tasks, it is necessary to consider whether the target multi-modality images are sufficient to generate interpretable and meaningful contrastive learning masks. This constitutes not only a pivotal impediment to its generalizability, but also an essential precondition to ensure its efficacy. We intend to delve deeper into this topic in future work, with the aim of refining the CCL and potentially proposing new method.

\section{CONCLUSION}
This paper proposes a novel contrastive learning network with integrated multi-level features for fusing infrared and visible images.  We develop twin contrastive constraints to preserve typical features and avoid the redundant features during the fusion process. As a result, the twin contrastive constraints can achieve better visual effects in a soft manner, \emph{i.e.,} salient thermal targets and rich faithful details. We also design a multi-level attention in our network to learn rich hierarchical feature representation and ensemble better transmission.  Besides, a self-adaptation weight is designed to overcome the limitation of the manually-craft trade-of weight in the loss function. The qualitative and quantitative results demonstrate that the proposed method achieves the SOTA performance with high efficiency. Moreover, ablation experiments valid the effectiveness of our method. Furthermore, we extend our CoCoNet to apply on medical image fusion, and it can also achieve an superior performance compared with other state-of-the-art approaches.

\section*{Acknowledgment}
\thanks{This work is partially supported by the China Postdoctoral
Science Foundation (2023M730741), and the National Natural
Science Foundation of China (No.62302078,22B2052, 62027826),the National Key R\&D Program of China (No.
2022YFA1004101).}

\section*{Availability of supporting data}

Figures used in this study have been deposited in github

(https://github.com/runjia0124/CoCoNet).

\bibliographystyle{sn-basic}
\bibliography{reference}
\end{document}